# Soft (Gaussian CDE) regression models and loss functions

José Hernández-Orallo           (jorallo@dsic.upv.es)
Departament de Sistemes Informàtics i Computació
Universitat Politècnica de València, Spain

November 7, 2012

**Abstract**

Regression, unlike classification, has lacked a comprehensive and effective approach to deal with cost-sensitive problems by the reuse (and not a re-training) of general regression models. In this paper, a wide variety of cost-sensitive problems in regression (such as bids, asymmetric losses and rejection rules) can be solved effectively by a lightweight but powerful approach, consisting of: (1) the conversion of any traditional one-parameter crisp regression model into a two-parameter soft regression model, seen as a normal conditional density estimator, by the use of newly-introduced enrichment methods; and (2) the reframing of an enriched soft regression model to new contexts by an instance-dependent optimisation of the expected loss derived from the conditional normal distribution.

# Contents

# 1 Introduction

Common day applications of predictive models usually involve a full use of the available contextual information. When the operating context changes, one may fine-tune the by-default (incontextual) prediction or may even abstain from predicting a value (a reject). Consider a common case where a regression model has been built from some training data, and the model has to be deployed to new instances. If the context is the same for the new instances as it was for the training data, then the quality of the predictions will mostly depend on the observed quality of the model for the same context. However, if the context changes, the prediction given by the model may be suboptimal. For instance, if the model has been trained with a symmetric *loss function* but the deployment operating context involves an asymmetric loss function (where, e.g., underestimations have higher loss than overestimations), then predictions will need to be adjusted. In order to do this there are two options: (1) re-train or revise the model by using a possibly modified (e.g., oversampled) training data and the new loss function, or (2) use a *reframing* function which takes the model and the operating context and outputs a new reframed prediction. The first option is not always possible since many regression methods are not cost-sensitive or cannot be (easily) adapted to work with different (possibly complex) loss functions. Also, in the cases where the first option is possible, the training data must be preserved indefinitely and an important computational cost is incurred to retrain the model all over again. This is especially the case whenever the operating context changes recurrently, even for two consecuentive individual predictions.

This kind of general problems has been profusely studied for classification, where the notion of operating context (or condition) is common and well understood. Some of the techniques and notions for addressing these cases are cost matrices, cost-sensitive classification [19], ROC analysis [61, 22, 32], threshold-choice methods [42], calibration [14, 3, 5] and, of course, the notions of soft classifiers (outputting a score or probability) versus the notion of crisp classifiers (just outputting a label). Certainly, there have also been a few efforts to find the parallel of these techniques for regression. However, most of them rely on a crisp view of the regression model, i.e., they work with regression models which *just* output a value. Examples of this are the Regression Error Curves [7], utility-based regression [63, 65], the definition of ranking measures [58] and the use of transformation functions for regression which derive a *global* reframing that must be constantly (or polynomially) applied to the output of the regression model [1, 72]. None of these approaches represents the right mapping between classification and regression. Whenever we consider a scoring classifier (or a ranker) in classification, which can sort their predictions by their reliability (at least in the binary case), we should consider a regression model which can sort their predictions by their reliability. Whenever we consider a probabilistic classifier, which in fact outputs a discrete distribution on the labels (a categorical distribution), we should consider a regression model which outputs a continuous distribution (e.g., a normal distribution), and not a single value. This correspondence is shown in Table 1.

From this correspondence, we see that the natural way of addressing context-sensitive problems in regression is the use of soft regression models (as soft classification models are the natural way of addressing context-sensitive problems). We need regression techniques which not only output the estimated expected value for each instance $x$, i.e. $\hat{\mathbb{E}}(y|x)$ (also referred to as the conditional mean), but also accompany these predictions with an estimated error, reliability or density function. There are many approaches for this. One approach is to obtain the standard error for each prediction as calculated by each specific technique (e.g., linear regression) if the algorithm provides a way to obtain this value for each prediction (which is not always the case). A second approach is to estimate the "reliability of individual regression predictions" [8], through sensitivity analysis, local averaging or other techniques, which can be applied to any regression method, as shown in [9]. A third approach is *conformal prediction* [60, 50, 49, 51], or any other method which derives a confidence interval. Finally, we can, of course, use conditional density (or distribution) estimation methods [57, 44, 40], which can derive the conditional probability density function of the dependent variable $y$, i.e. $\hat{f}(y|x)$, by using kernel or distribution mixtures. It has been recently said that "conditional density estimation has been studied extensively in economics and Bayesian statistics [..., but] it has received only little attention in the machine learning literature" [11]. One reason might be that conditional density estimation is not easy to apply for many regression methods.

|  | Classification | Regression |
| --- | --- | --- |
| Crisp | a class label $\hat{c}(x)$ | a numerical value $\hat{m}(x)$ |
| Soft | a score for each class $\hat{s}_c(x)$ | a numerical value $\hat{m}(x)$ and a reliability measure $\hat{r}(x)$ (e.g., confidence interval) |
| Probabilistic | a categorical distribution (characterised by a conditional probability function $\hat{p}(y|x)$). | a continuous distribution (characterised by a conditional density function $\hat{f}(y|x)$) |

Table 1: Correspondence between different types of classification and regression. Evaluation also depends on the kind of prediction. For instance, crisp prediction implies the comparison of the estimated output with the actual output, while probabilistic prediction implies the comparison of discrete distributions in classification ($p(y|x)$ with $\hat{p}(y|x)$) and the comparison of continuous distributions in regression ($f(y|x)$ with $\hat{f}(y|x)$).

However, given these approaches for soft regression, none of them has been generally applied for context-sensitive problems, because either these proposals are inappropriate, or are much too complex. For instance, standard errors, reliability metrics and confidence intervals are useful to rank the predictions according to their reliability, or to address some tolerance issues, but they cannot be used to get a precise quantifiable magnitude of what the *expected loss* will be for an instance and a specific operating context. On the other hand, conditional density estimation looks like the appropriate setting for this, since we can (theoretically) calculate the expected loss (i.e., the risk) as an integral over all the possible values for the dependent variable, weighted by its density estimation. The problem is that it is not easy to calculate this minimisation since the estimated density function may be non-monotonic, non-convex or not even continuous.

In this paper we propose a simple approach for soft regression. In most cases, it is just sufficient to have a good estimation of a conditional *normal* (i.e., Gaussian) density function. This has several advantages. First, a normal distribution only needs two parameters, the mean (expected value) and the variance. This makes it possible to estimate these two parameters easily. It can be done from the regression methods themselves, from any estimation of the standard error, from a confidence interval and, of course, from any other more complex density function or mixture thereof. Second, the variance can be used to rank predictions in a very straightforward way, as is done with reliabilities, but with a clear interpretation of the magnitudes. Third, and most importantly, we can work analytically with the normal distribution and smoothly derive the exact expression leading to the output that minimises the expected loss for many common loss functions.

In fact, we will see that there are extremely simple methods to estimate this variance which can be applied to any *crisp* regression technique. Some of these methods are just based on comparing the prediction for the training dataset with the actual value, disregarding the input domain. In this sense, these methods are closely related to calibration methods in classification. However, we call them 'enrichment' methods since they preserve the original prediction mean, while only adding a second parameter, the variance, to form a more powerful and flexible soft regression model for context-sensitive applications.

Many common applications of regression where deployment contexts can change are then solved by this setting: cost-sensitive applications where we have asymmetric losses, screening applications where we need rejection rules to determine the examples for which no prediction will be issued, auction and retailing bids where prices (or other continuous variables) are chosen to obtain the maximum expected profit, situations where we want to derive the probability that two or more predictions are in the right order, etc.

In what follows, we analyse and derive the solutions for many of these problem families, using two-parameter regression models and deriving the optimal prediction for the corresponding loss function. For each of these families, we perform a complete set of experiments which show that our general approach is not worse than some specific solutions in the literature for some of these families, and is clearly better for others. As a result, the setting and methodology we introduce in this paper can be effectively (and easily) applied to a wide range of context-sensitive problems.

In brief, the goal of the paper is to show that two-parameter regression (as an estimation of conditional mean and variance or, more easily, as an enrichment of a crisp model by simple conditional variance estimations), followed by a probabilistic reframing assuming a normal distribution, is a simple, general and powerful method which can successfully address many kinds of problems.

The paper is organised as follows. Section 2 introduces some notation about regression models and loss functions, as well as relevant previous work which triggers the introduction of the general notion of reframing. We briefly define different types of reframing and the optimal prediction expressions for probabilistic reframing. From here, the objective of the paper is re-stated in a more precise way and the experimental methodology is settled for the rest of the paper. Section 3 analyses and compares several conditional density estimation methods and other methods for soft regression, as well as methods which can be used for the enrichment of a crisp model by the use of univariate conditional variance estimation methods. Since there are many possible approaches, this analysis is necessary for choosing only a few good, simple methods for the following sections. Section 4 formally derives the optimal decision rules for the two loss function representing bids in auctions, sales or other trading scenarios, assuming a conditional normal distribution. We perform a complete set of experiments using probabilistic reframing with different enrichment methods and compare them against a global reframing method based on a constant shift over the training set. Section 5 performs a similar procedure for asymmetric losses: absolute (linear) and squared (quadratic). We also compare to two global reframing methods, one based on a constant shift and another based on a polynomial shift. Section 6 also introduces loss functions representing the situation where we have rejection rules in regression. Similarly, we derive and reuse the expressions for optimal reframing and compare several approaches. Section 7 makes a comprehensive analysis of results, suggests many other applications which could be modelled as loss functions and closes the paper with a summary of the contributions and the work ahead. Several appendices (which can be skipped on a first reading) complete the paper with some additional information for the datasets and metrics used in the experiments, more detailed results for some techniques that have been dismissed and some proofs.

## 2 Background

We start with some basic definitions and notation, followed by some related work. Then we introduce the key notion of reframing (and the distinction between global and local reframing).

### 2.1 Regression and conditional densities

Let us consider a multivariate input (or predictor) domain $\mathbb{X} \subset \mathbb{R}^d$ and a univariate output (or response) domain $\mathbb{Y} \subset \mathbb{R}$. The domain space $\mathbb{D}$ is then $\mathbb{X} \times \mathbb{Y}$. Labelled examples or instances are just pairs $\langle x, y \rangle \in \mathbb{D}$, and datasets are subsets of $\mathbb{D}$. Unlabelled examples are elements $x \in X$, sometimes represented as $\langle x, ? \rangle$. We denote by $D_X$ and $D_Y$ the projection of $D$ for the input domain and output domain respectively. A *crisp* regression model $\hat{m}$ is a function $\hat{m} : \mathbb{X} \to \mathbb{Y}$. A *soft regression model* accompanies each prediction with a reliability, confidence or, more generally, a conditional probability density function $\hat{f}(y|x)$ with $y \in \mathbb{Y}$ and $x \in \mathbb{X}$. The corresponding cumulative distribution function is $\hat{F}(y|x) = \int_{-\infty}^{y} \hat{f}(t|x)dt = \hat{p}(Y \leq y|x)$, i.e.

the probability of the output being lower or equal than $y$ for an input $x$. The estimated expected value (conditional mean) is denoted by $\hat{\mu}_{\hat{f}}(x) \triangleq \mathbb{E}_{\hat{f}}(y|x) = \int_{-\infty}^{\infty} y\hat{f}(y|x)dy$. We denote its (conditional) standard deviation by $\hat{\sigma}_{\hat{f}}(x)$. We will drop the subindices when clear from the context. Note that the mean and the standard deviation are conditional, i.e., defined for one single example; these are not the mean and standard deviation of a distribution of examples (or a whole dataset).

We can derive a crisp regression model from a conditional density function $\hat{f}(y|x)$, as $\hat{m}(x) = \hat{\mu}_{\hat{f}}(x)$. The target function will be represented by a true density function $f(y|x)$. If the target function is deterministic, it can be represented with a Dirac delta function (all the density mass falls over the true single value) or more simply, as a deterministic function $m : \mathbb{X} \to \mathbb{Y}$.

The normal distribution will be represented as usual $\mathcal{N}(\mu, \sigma^2)$, with probability density function $\phi_{\mu,\sigma^2}(\cdot)$ and cumulative distribution function $\Phi_{\mu,\sigma^2}(\cdot)$. For the standard probability density function and the standard cumulative distribution function we will drop the subindices, and we will write $\phi(\cdot)$ and $\Phi(\cdot)$ respectively.

## 2.2 Cost-sensitive problems and loss functions

In context-sensitive learning, there are several features which describe a context, such as the data distribution, the costs of using some input variables and the loss of the errors over the output variables. In this paper, we focus on loss functions over the output. As we will see, by properly defining the loss function and its parameters we can analyse and address many problem families. Let us start with the definition of loss function:

**Definition 1.** *A loss function is any function $\ell : \mathbb{Y} \times \mathbb{Y} \to \mathbb{R}$ which compares elements in the output domain. For convenience, the first argument will be the estimated value, and the second argument the actual value, so its application is usually denoted by $\ell(\hat{y}, y)$.*

Typical examples of loss functions are the absolute error ($\ell^A$) and the squared error ($\ell^S$), with $\ell^A(\hat{y}, y) = |\hat{y} - y|$ and $\ell^S(\hat{y}, y) = (\hat{y} - y)^2$. These two loss functions are *symmetric*, i.e. for every $y$ and $r$ we have that $\ell(y+r, y) = \ell(y-r, y)$. The are also commutative, i.e., for every $y_1$ and $y_2$ we have that $\ell(y_1, y_2) = \ell(y_2, y_1)$.

While many methods use these generic loss functions (such as $\ell^S$), most applications do have different loss functions. For instance, the bounded absolute error ($\ell_{BA,\beta}$) is defined as $\ell_{BA,\beta}(\hat{y}, y) = min(|\hat{y} - y|, \beta)$, which is also symmetric and commutative. Another example is the bid loss function $\ell_\beta^B(\hat{y}, y) = -\hat{y} + \beta$ if $\hat{y} \leq y$ and 0 otherwise, which is clearly asymmetric. In practice, there can be specialised loss functions for virtually any application domain.

While some previous works in the literature of regression techniques have focussed on re-designing the learning technique to account for specific loss functions during training ([15, 46]), only a few have considered the problem as a post-hoc process, once the model has been learnt. A post-hoc process can be performed in cases where re-training with the new loss functions is not possible (because of the regression technique is not cost-sensitive or because the training data is no longer available). It also has several advantages, such as model reuse and the possibility of applying the same methods to virtually any regression technique. This post-hoc process can be traced back to the seminal work by Granger [37], showing that the optimal predictor for some asymmetric losses can be expressed as the conditional mean plus a constant bias term [38]. However, it is recognised that solving this term is not always easy (or even possible in closed form) for many loss functions and density functions. Specific results have been studied for some particular loss functions, such as *Lin-Exp* (approximately linear on one side and exponential on the other side) and *Quad-Exp* (approximately quadratic on one side and exponential on the other side), which have general solutions with mild conditions [71]. Conversely, general closed-form solutions for *Lin-Lin* (asymmetric linear) and *Quad-Quad* (asymmetric quadratic) do not exist in general [12] [13]. In fact, even general non-closed-form expressions are not always possible unless some constraints are imposed, such as continuous loss functions, finite expected loss and particular properties on the moments of the density function [20].

In general, much of this work is restricted to continuous loss functions in time series or system reliability applications [2] [62], but provide sufficient evidence that working with complex density functions is very problematic for general (and possibly discontinuous or non-convex) loss functions. This has motivated the appearance of other approaches which do not use any density estimation, such as the calculation of a global function which is applied to the outputs [1, 72]. In this case, the restrictions come on the side of the loss function, which must be convex, and the requirement that the training set (technically, only the true values *y*) must be preserved from the training to the deployment stage.

However, some other problems are not usually considered, not even as generalised loss functions [38]. Examples of these context-sensitive problems are *rejection rules*, where we want that the model abstains from outputting a prediction for the most unreliable cases. None of the previous approaches has addressed this problem. In fact, *rejection rules* are common in classification [28, 52], but rarely seen as cost-sensitive problems in regression. However, as we will see, these problems can be modelled with a loss function which sets a cost of rejection. Also, many regression problems used for product prescription, sale predictions and auction bids look at finding the bid price, i.e., the appropriate quantity (or other negotiable feature) which has the maximum expected benefit [4]. Many of these problems can also be modelled by a loss function and, yet again, the predictions of the model can be fine-tuned for them.

As a result of this variety and diversity of problem families which can be modelled with loss functions or other kind of context information, we can integrate and generalise some of the existing (and new) *model adaptation* procedures into a more general term that we call *reframing*.

## 2.3 Reframing and optimal predictions

Given a loss function representing a particular context, the objective is to get predictions with low loss rather than predictions with low error. In order to do this, we do not train a model using this loss function (as risk minimisation approaches could do), because the loss function may not be known at the training stage or the regression technique may not be able to process loss information. Even if possible, re-training a model whenever the context changes is not a very efficient approach in terms of resources. It may also be inefficient in terms of reliability if the application requires stable, validated models. As an alternative, we propose the use of *reframing* functions, which adapt the predictions of the original model to the context, represented by a loss function.

**Definition 2.** *A reframing function is any method which produces a predicted output value given the input value x, the loss $\ell$ and the model $\hat{f}$.*

$$r(x, \ell, \hat{f}) \rightarrow \dot{y} \qquad (1)$$

*where $\dot{y}$ represents the reframed output.*

Figure 1 shows the process of reframing graphically. Note that we do not impose any restriction on how the training data is obtained from the training context. Also, we do not assume that this data generation process has to be similar to the process generating the unlabelled data from the deployment context. In fact, the distributions of predictor *x* and response *y* will usually differ between contexts, as we will reflect in the experimental setting.

For those crisp regression methods not using a density, we can assume a delta Dirac function $\hat{f}_m$, whose mean is clearly the prediction point given by the model, i.e., $\hat{m}(x) = \hat{\mu}_{\hat{f}}(x)$, or we can define reframing methods which do not need a density, by expressing them as a function which only depends on the loss function and the expected mean for each example, as follows.

$$r(x, \ell, \hat{f}) = R(\ell, \hat{\mu}_{\hat{f}}(x))$$

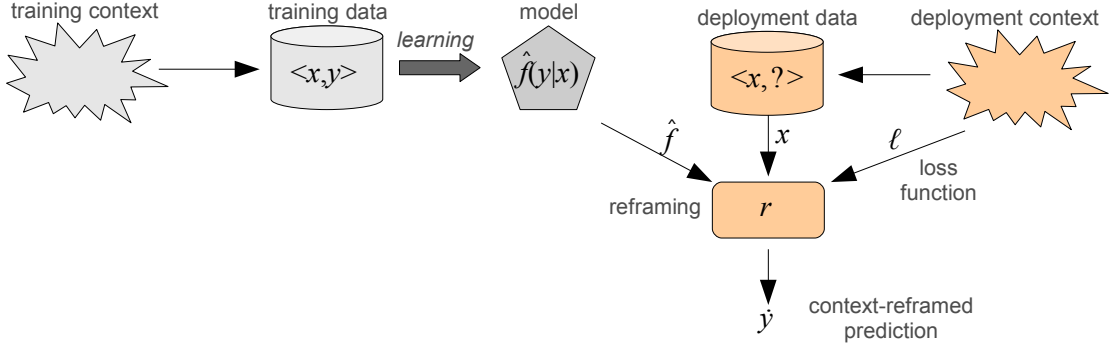

Figure 1: Reframing process adapting predictions from one context to a different context.

**Example 1.** *For instance, given the bid loss function $\ell_\beta^B$ with $\beta = 0$, we might consider the following global reframing function:*

$$R_1(\ell_\beta^B, \hat{y}) = 0.8 \times \hat{y}$$

*which just systematically reduces predictions by a 20%. The rationale here is given by the fact that overestimations imply a 0 loss (no deal) and underestimations always imply a benefit (there is a deal, and we assume that y always represent positive prices, so giving a negative loss). So, a 20% reduction creates some margin which can produce higher overall benefit.*

*Alternatively, a local (probabilistic) reframing could be done using $\hat{f}$, if available. For instance, using the same bid loss function $\ell_\beta^B$ with $\beta = 0$, a probabilistic reframing might be:*

$$r_2(x, \ell_\beta^B, \hat{f}) = \hat{F}^{-1}(0.25)$$

*where $\hat{F}^{-1}$ is the quantile function for $\hat{f}$, i.e., inverse of the cumulative function $\hat{F}$. This means that we predict the value such that 25% of the expectancy for y is below that value.*

Figure 2 shows the use of $R_1$ and $r_2$ above for two different instances.
In general, we can distinguish four kinds of reframing:

- Constant global reframing: all predictions are modified in the same way independently of $\hat{y}$, e.g. adding a constant $s$ (i.e., $\dot{y} \leftarrow \hat{y} + s$). This constant $s$ is called the *shift*.

- Non-constant global reframing: predictions are modified using a (e.g. polynomial) function of $\hat{y}$. While the shift is different for each example, it only depends on the prediction, and it can be considered a 'global' method, since it applies a global function.

- Non-probabilistic local reframing: predictions are modified by a transformation of $\hat{y}$ using some reliability or confidence parameters. For instance, we could define a reframing method which only modifies (or rejects) the instances that are below a given reliability threshold or above a percentage of the confidence width.

- Probabilistic local reframing: the outputs are adapted according to a transformation over the conditional density function. If $\hat{f}$ is a parametric distribution, we can just use the parameters as arguments for the transformation. For instance, if $\hat{f}$ is a normal distribution, then we can just define the reframing transformation in terms of the mean $\hat{y} = \hat{\mu}_{\hat{f}}(x)$ and the conditional standard deviation $\hat{\sigma}_{\hat{f}}(x)$.

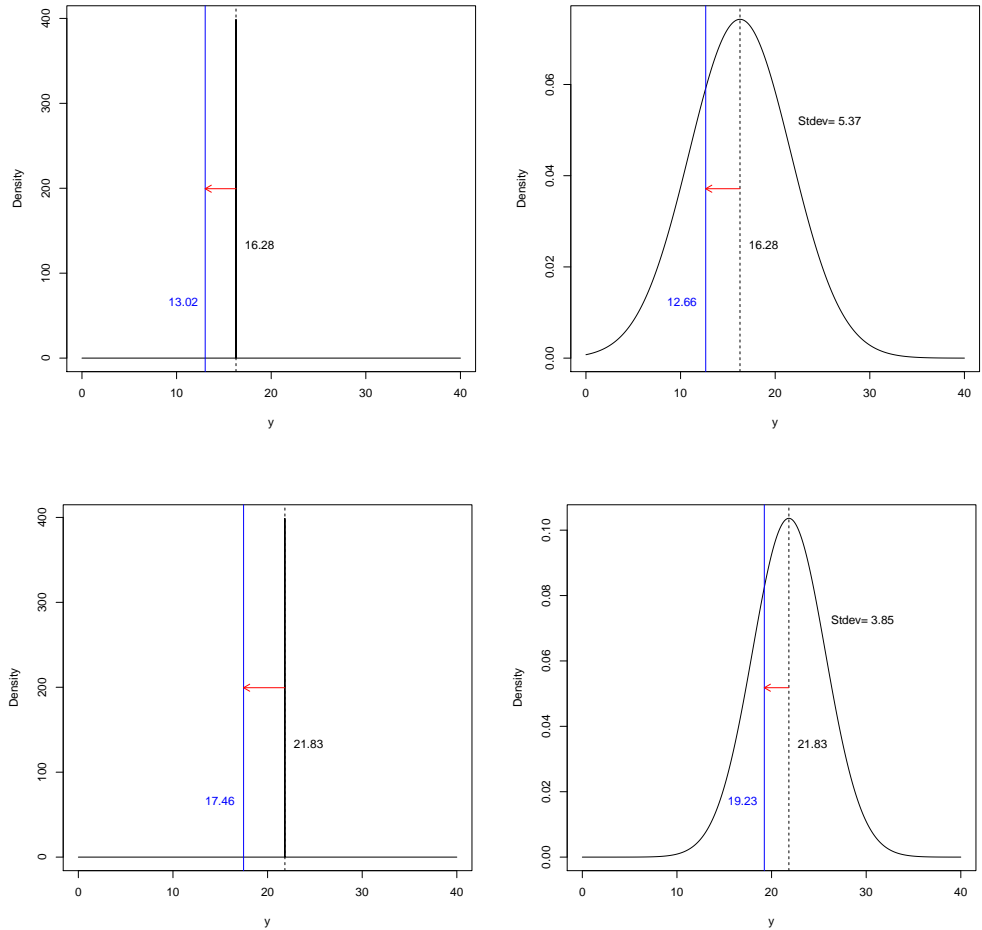

Figure 2: Representation of how predictions are shifted according to different reframing methods. The top row corresponds to an instance whose expected value is 16.28. Without reframing, this would be the output value. The top-left plot shows how the prediction moves to 13.02 (applying the global reframing $R_1$, which multiplies the prediction by 0.8). The top-right plot shows a soft regression model with a normal conditional density function $\hat{f}$ which gives a mean at 16.28 and standard deviation of 5.37 for this instance. The prediction moves to 12.66 (applying the local reframing $r_2$, using $\hat{F}^{-1}(0.25)$). The bottom row shows a similar picture for a different instance where the expected value is 21.83. The soft regression model gives a standard deviation of 3.85 for this example. The reframing $R_1$ leads to 17.46 (bottom left) while the reframing $r_2$ leads to 19.23. As we see, the shift for the first instance (top) is greater for the local method (right), while the shift for the second instance (bottom) is greater for the global method (left).

The notion of shift (and reframing in general) for regression is closely related to similar procedures which are usually performed in classification. For instance, when we have a crisp classifier, we can (randomly) tweak some of the predictions, in order to get a more balanced set of predictions or to favour one class against others according to a cost matrix. This view would correspond to the global reframing methods above. For soft classifiers, the choice of an appropriate threshold to convert scores into predictions would correspond to local reframing. In particular, when we work with calibrated classifiers and we try to find the optimal thresholds (as in [48]) using a probabilistic setting, we have a scenario that is parallel to probabilistic local reframing.

In regression, approaches to global reframing rely on the calculation of a constant (or function) from the training set. For instance, one simple method for global constant reframing is the calculation of the best constant shift for the training set given a loss function. One problem of these methods is that we do not always have the training (or a validation) dataset. Or even if we can have the dataset, it might be costly to keep it. Also, in many applications the loss function parameters may be different for each instance and the shift needs to be recalculated by exploring the whole training dataset all over again. Finally, this reframing may be problematic when the output distribution differs between the training dataset and the deployment dataset, because the global function is optimised for the training dataset.

On the contrary, local reframing does not have the above-mentioned problems, but requires, in the probabilistic case, an *accurate* conditional density estimation and a method to derive the *optimal* reframing in an analytical or numerical way. In what follows, we will focus on the notion of 'optimal reframing' in the probabilistic (local) case, since the notion of optimality for the non-probabilistic (global) case is more elusive (since reliability or confidence measures cannot be used, in general, to quantify the expected loss).

## 2.4 Optimal probabilistic reframing

It seems reasonable to think that better decisions can be made if we take the conditional density function $\hat{f}(y|x)$ into account, rather than just the expected value $\mathbb{E}_{\hat{f}}(y|x)$, provided this density function is well-estimated.

The maximum density reframing is given by $r^{max}(x, \ell, \hat{f}) = \arg\max_y \hat{f}(y|x)$, which ignores the loss function, and just gives the point with maximum density. The *mean* reframing is given by $r^{mean}(x, \ell, \hat{f}) = \mathbb{E}_{\hat{f}}(y|x) = \int_{-\infty}^{\infty} y\hat{f}(y|x)dy = \hat{\mu}_{\hat{f}}(x)$ which also ignores the loss function. For some density functions, e.g., a normal distribution, we have $r^{max} = r^{mean}$.

In general, we want to take the loss function $\ell$ into account. For an unlabelled instance $\langle x, ?\rangle$, the expected loss (risk function) for prediction $t$ is given by:

$$\mathscr{L}(x, t, \hat{f}, \ell) = \int_{-\infty}^{\infty} \ell(t, y)\hat{f}(y|x)dy \qquad (2)$$

Then, the prediction with minimum expected loss is calculated by the following reframing function:

$$r^*(x, \ell, \hat{f}) \triangleq \arg\min_t \mathscr{L}(x, t, \hat{f}, \ell) = \arg\min_t \int_{-\infty}^{\infty} \ell(t, y)\hat{f}(y|x)dy \qquad (3)$$

This equation says that the optimal prediction (ignoring the uncertainty of the estimation of $\hat{f}$) for each example $x$ depends on its estimated distribution and the loss function. Interestingly, the previous equation for $r^*$ is independent of the data (marginal) distribution $f(x)$, which means that it can be applied to each individual instance without considering the rest. This is important, since the loss function $\ell$ may even vary for different instances.

The question, now, is how to *solve* eq. (3). In some cases, if $\ell$ and $\hat{f}$ follow some properties, we can easily solve the equation. For instance, the following proposition gives the result for the easiest (well-known) case (the proof is in appendix H):

**Proposition 1.** *If $\hat{f}$ is symmetric[1] and $\ell$ is symmetric and commutative then*

$$r^*(x, \ell, \hat{f}) = \hat{\mu}_{\hat{f}}(x) = \int_{-\infty}^{\infty} y \hat{f}(y|x) dy = r^{mean}(x, \ell, \hat{f})$$

*which states that the optimal prediction is given by the mean of the conditional density function.*

But in many applications, $\ell$ is not symmetric. Also, for many density estimation mehtods, $\hat{f}$ is not symmetric either. Only if $\hat{f}$ is chosen to be a simple distribution (e.g., a normal distribution), the equation can be solved analytically for specific, asymmetric loss functions, as we will see in the following sections. In the general case, however, the best prediction cannot be calculated in an analytical way and needs to be obtained by a numerical method, such as a Monte Carlo method, or any other method (e.g., hill-climbing) which can exploit the properties of particular cases for $\ell$ and $\hat{f}$, such as (partial) monotonicity or convexity.

## 2.5 Goals and experimental design

Once we have the ingredients, terminology and concepts, we can state our research goal more properly. Namely, in the rest of this paper, we will answer several questions. First, are there general and practical conditional density estimation methods which can be used effectively to reframe an existing crisp regression model? In order to answer this question we need to explore techniques which can produce conditional density estimations for any crisp regression technique. We will focus on normal density estimations, because only two parameters are needed, mean and variance, and the former is already given by any regression model. This implies that we will be able to derive the reframing transformation relatively smoothly for most loss functions. Consequently, the following section will be devoted to the experimental analysis of several old and new approaches to normal (Gaussian) conditional density estimations. From this analysis we will be able to select some methods that will be used in subsequent sections.

The second, ultimate, question of this paper is whether probabilistic reframing methods based on these simple estimators are able to solve a broad set of cost-sensitive problem families, including bidding problems, asymmetric loss functions and rejection rules. We will devote a section to each of these families, we will derive the formal expressions for the reframing transformations and we will compare these methods with other previous *specific* methods in the literature addressing each of these problem families. We will see that probabilistic reframing is more general and effective.

Apart from the theoretical derivations of the reframing transformations, an important part of the upcoming sections relies on experimental results. We will briefly describe the general experimental setting now, and we will let some other details for each specific section.

We will use forty datasets, as shown in tables 11 and 12 in appendix A. The first battery of datasets will be used for the experiments in section 3. We will use the other battery for the experiments in sections 4, 5 and 6. The reason for two different batteries is that we select the best conditional density estimation and enrichment methods in section 3 with the first battery, and we use these methods with a fresh and independent battery for the particular applications in the other sections.

In all the experiments we will use 2-fold cross-validation without previously shuffling the datasets (i.e. the order is preserved before splitting them). This configuration tries to mimic a realistic situation where the training and test distributions may differ. Note that increasing the number of folds or shuffling the datasets would yield similar distributions for training and test (in terms for the predictors *x* and response *y*), which is a quite uncommon scenario in practice (although relatively usual in machine learning research experiments). It is important to highlight that it is not our goal to obtain an estimation of how well each method will perform for each dataset under exactly the same distribution (where usual re-sampling methods such as 10-fold cross-validation would be appropriate), but to compare several methods on a realistic setting where the context between training and test can change, including the data distribution. In order to illustrate how training and test distributions differ, two extra columns (TrTeMD, TrTeKS) in the dataset tables 11 and 12 show the train-test relative means difference, calculated as $\frac{|\mu_{Train}-\mu_{Test}|}{\sigma_{Train}}$, and the train-test Kolmogorov-Smirnoff statistic, respectively. Both are averaged for the two folds. The higher these values are the more dissimilar the training and test distributions are.

---

[1] The notion of symmetry for the loss function has been defined above. The notion of symmetry for a distribution is the classical notion of symmetry relative to the mean.

In order to assess the significance of the experimental results we will use a custom procedure, following [45] and [31, ch.12], which in turn is mostly based on [16]. Since we will not have any baseline method, we will use a Friedman test to tell whether the difference between several methods is significant and then we will apply the Nemenyi post-hoc test. We agree with [35] that the Nemenyi test is a "very conservative procedure and many of the obvious differences may not be detected", but we prefer to be conservative given our experimental setting and the use of a 0.95 confidence level. In some result tables we will show the means (even though in many cases they are not commensurate) and in some other tables we will show the average ranks (from which the Friedman and Nemenyi tests are calculated). We will also include the critical difference for the Nemenyi test, so we will be able to simply tell whether the difference between two algorithms is significant if the difference between their average ranks is greater than the critical difference.

## 3 Normal conditional density estimation (*NCDE*): enrichment methods

A theoretically-optimal decision rule for a conditional density function and a loss function will only work if the conditional density function $\hat{f}(y|x)$ is accurate. While there are many techniques for conditional density estimation (CDE, see appendix C), they may be inappropriate for cost-sensitive scenarios. First, as they focus on the *whole* conditional distribution, the estimated conditional *mean* given by these complex estimated conditional density functions is usually worse than the conditional mean output by many crisp regression methods. Second, CDE methods are usually slow. Third, in many problems, the actual conditional density functions are not multi-modal, and even if they are, it is not clear that adjusting many parameters to approximate this multi-modality will finally lead to the choice of a better (or even significantly different) optimal prediction for many loss functions. Finally, some problems are deterministic and what we really want is an estimation of the residual rather than (technically) a conditional density function for the output variable.

Instead of complex (usually non-parametric) CDE methods, one of the simplest, most common, parametric density functions is given by the normal (Gaussian) distribution. Estimating a normal distribution only requires the estimation of two parameters, the mean and the variance. It is important to clarify that the use of a normal conditional density $\hat{f}(y|x) \sim \mathcal{N}$ does not entail —at all— that the output variable is distributed normally ($\hat{f}(y) \sim \mathcal{N}$). Moreover, the use of an estimated normal conditional density $\hat{f}(y|x)$ does not even mean that we *assume* that the true conditional density $f(y|x)$ is normal. In fact, when having an empirical dataset, we do not have information about the true conditional distribution; we just have examples for which its actual distribution can be seen as a Dirac delta function. In other words, the use of a normal conditional density function follows practical considerations and can be seen (at most) as a representation of the model's belief about how its uncertainty is distributed, i.e., a model of the *distribution of residuals*.

Consequently, in this section we will explore and develop normal conditional density estimation methods, or NCDE methods for short. This boils down to a soft regression model that, for every input instance $x$, just outputs two parameters: $\hat{\mu}(x)$ and $\hat{\sigma}(x)$. The estimation of $\hat{\mu}(x)$ is the goal of all (crisp and soft) regression methods. Consequently, we will focus below on the estimation of $\hat{\sigma}(x)$, comparing the results of several methods. The goal of this section is not to find the best estimator for $\hat{\sigma}(x)$ as an isolated problem, but to find simple and general methods that work well when the conditional mean $\hat{\mu}(x)$ is already given by any crisp regression technique.

In order to perform the comparison, we need evaluation metrics for conditional density estimators. We are interested in metrics that can evaluate (1) how good the conditional mean is, (2) how good the conditional variance is, and (3) how good the conditional density is (which is given by the qualities of the mean and the variance). For the conditional mean we will use the mean relative square error (*mrse*), a standardised version of the square error. For the conditional variance we will use the mean standardised variance ratio (*msvr*), which is a standardised metric of the ratio between the estimated variance and the squared residuals. Finally, for the conditional density we will use the mean standardised likelihood (*msll*), a standardised version of the log-likelihood. All these measures are standardised between 0 (best) and 1 (worst). The exact formulations for these metrics can be found in appendix B.

## 3.1 Directly estimating the variance from the regression techniques

The first way of obtaining the mean and variance for each prediction is choosing a base regression technique which directly or indirectly is able to provide the variance (or a measure of standard error). In this paper, we will work with three common base regression techniques:

- Linear regression (*LR*): many implementations of linear regression can calculate the standard errors for each predicted point, $se(x)$. If this is the case, we can just set $\hat{\sigma}(x) = se(x)$. The particular *LR* method we will use is ordinary least squares using the function `lm` of R [55] with default parameters.

- Nearest neighbours (*kNN*): in this case the variance is calculated as the variance of the actual *y* values for the *k*-closest elements. In particular, we use an unweighted *k*-nearest neighbours algorithm using the Euclidean distance (with all the attributes scaled by the function `scale` in R) with $k = 10$.

- Regression trees (*Tree*): in this case, one easy way of calculating the variance is to calculate the variance for the actual *y* values (in the training set) for each leaf of the tree. Then, for each new prediction on a new dataset, the variance will be given by the variance of the leaf where the example falls. We use the CART algorithm [10] implemented by the function `tree` in the package `tree` in R with its default parameters.

We will use these three *base* techniques throughout the rest of the paper. Table 2 shows the result for the three methods above using the three evaluation metrics (*mrse*, *msll*, and *msvr*).

Interestingly, we can see that for some datasets one method is better than the rest for the conditional mean (evaluated by *mrse*), while it can be the worst for the conditional variance (evaluated by *msvr*). In general, we see that *LR* gives the worst estimations for the conditional mean (*mrse*) and variance (*msvr*), which then implies bad results for the density estimation (*msll*).

## 3.2 *NCDE* from conditional density, variance, reliability or confidence estimators

The procedure seen above is based on using the variance derived from the own regression technique. These techniques are crisp, and are not really designed to obtain good conditional variances or densities. Instead, 'soft' regression techniques (conditional density estimation, conditional variance estimation, reliability estimation and confidence estimation using conformal prediction) look more appropriate for deriving a *normal conditional density estimator* (*NCDE*) model. The use of these methods for *NCDE* would generally involve that we attempt a related, but different (and sometimes more complex) problem first, and then use some transformation or derivation from the soft model to the *NCDE* model. There is nothing against this, provided the results are good and the procedure does not become extremely difficult or inefficient. Below we see why these two criteria are not met. A more detailed exploration is given in the appendices C, D, E and F, which also links to the literature.

|      | LR mrse | LR msll | LR msvr | kNN mrse | kNN msll | kNN msvr | Tree mrse | Tree msll | Tree msvr |
|------|------|------|------|------|------|------|------|------|------|
| 1    | 0.28 | 0.81 | 0.65 | 0.35 | 0.80 | 0.50 | 0.41 | 0.85 | 0.65 |
| 2    | 0.52 | 0.80 | 0.63 | 0.39 | 0.75 | 0.45 | 0.19 | 0.66 | 0.52 |
| 3    | 0.02 | 0.48 | 0.65 | 0.16 | 0.61 | 0.62 | 0.21 | 0.62 | 0.50 |
| 4    | 0.12 | 0.69 | 0.51 | 0.32 | 0.81 | 0.52 | 0.25 | 0.80 | 0.60 |
| 5    | 0.11 | 0.84 | 0.85 | 0.11 | 0.63 | 0.51 | 0.11 | 0.63 | 0.51 |
| 6    | 0.55 | 0.76 | 0.68 | 0.20 | 0.65 | 0.57 | 0.16 | 0.63 | 0.59 |
| 7    | 0.46 | 0.75 | 0.53 | 0.45 | 0.76 | 0.61 | 0.50 | 0.76 | 0.54 |
| 8    | 0.14 | 0.76 | 0.75 | 0.13 | 0.61 | 0.53 | 0.12 | 0.58 | 0.53 |
| 9    | 0.07 | 1.00 | 0.97 | 0.29 | 0.99 | 0.91 | 0.26 | 0.99 | 0.94 |
| 10   | 0.31 | 0.76 | 0.61 | 0.31 | 0.59 | 0.45 | 0.23 | 0.55 | 0.48 |
| 11   | 0.19 | 0.77 | 0.63 | 0.15 | 0.74 | 0.63 | 0.16 | 0.68 | 0.65 |
| 12   | 0.89 | 0.88 | 0.73 | 0.47 | 0.77 | 0.49 | 0.45 | 0.71 | 0.50 |
| 13   | 0.02 | 0.58 | 0.64 | 0.45 | 0.89 | 0.61 | 0.44 | 0.89 | 0.65 |
| 14   | 0.45 | 0.85 | 0.80 | 0.32 | 0.72 | 0.52 | 0.22 | 0.65 | 0.50 |
| 15   | 0.50 | 0.75 | 0.73 | 0.55 | 0.73 | 0.59 | 0.56 | 0.72 | 0.57 |
| 16   | 0.41 | 0.78 | 0.62 | 0.26 | 0.72 | 0.56 | 0.22 | 0.69 | 0.59 |
| 17   | 0.97 | 0.99 | 0.91 | 0.39 | 0.78 | 0.54 | 0.37 | 0.76 | 0.61 |
| 18   | 0.30 | 0.76 | 0.70 | 0.34 | 0.75 | 0.57 | 0.35 | 0.74 | 0.54 |
| 19   | 0.33 | 0.72 | 0.64 | 0.34 | 0.70 | 0.52 | 0.48 | 0.69 | 0.55 |
| 20   | 0.58 | 0.82 | 0.70 | 0.20 | 0.68 | 0.52 | 0.26 | 0.69 | 0.58 |
| Mean | 0.36 | 0.78 | 0.70 | 0.31 | 0.73 | 0.56 | 0.30 | 0.71 | 0.58 |

Table 2: Three regression techniques using their own conditional variance estimation methods. Results use the datasets in Table 11, using the experimental methodology in section 2.5 and the metrics in appendix B.

Let us first review the most general approach, the direct estimation of a conditional density function $\hat{f}(y|x)$. Most conditional density estimation methods are designed to issue a complete model of the distribution, which is usually non-parametric. Appendix C describes this approach and shows how it can be adapted to get a normal conditional density. It also includes some experimental results which show that there is no improvement over the base techniques using their own conditional variance estimation methods. Also, general conditional estimation methods are very inefficient and cannot be used as a post-processing step for a crisp regression technique.

A second approach, conditional variance estimation (*CVE*), is much closer to our specific goal, and can be used to complement an existing crisp regression model by deriving a second parameter, the conditional variance, in order to make up a soft regression model. In fact, these methods can be understood as a post-processing step, which is applied to the whole training set, constructing a model of the residuals. Conditional variance estimation methods are explored in appendix D, but, again, results do not portray a clear advantage.

We have also explored some other methods based on reliability or confidence. Appendix E explores one of the best reliability estimation methods, *CNK*, included in a recent survey by Bosnic & Kononenko's [8] about reliability measures in regression. Also, it is meant to output an estimation of the standard deviation. However, the results are poor. Nonetheless, in appendix E we introduce a 'correction', known as *KNC*, which is just based on comparing the estimated mean with the closest $k$ true values in the training (or validation) set. The results for *KNC* are better, which has spurred us to introduce a univariate version *uKNC* that we will see below, as an enrichment method. Finally, we explored conformal prediction in appendix F, which outputs confidence intervals, but the results were not better than the rest.

We will evaluate a selection of these methods at the end of this section.

### 3.3 *NCDE* through enrichment methods

One of the problems of the previous methods is that they depend on the whole training set for estimating the conditional variance. This looks natural, since in order to get $\hat{\sigma}(x)$, we are supposed to need $x$. However, if we have a regression model, we already have $\hat{y}$, which actually carries information about the input value $x$. The basic idea of an 'enrichment' method is to derive $\hat{\sigma}(x)$ from $\hat{y}$ instead of deriving it from $x$. With the original $\hat{y}$ and the newly derived $\hat{\sigma}(x)$ we just have a *NCDE* model. Figure 3 shows this process of converting a crisp model into an enriched soft model. This univariate derivation can be performed in several different ways.

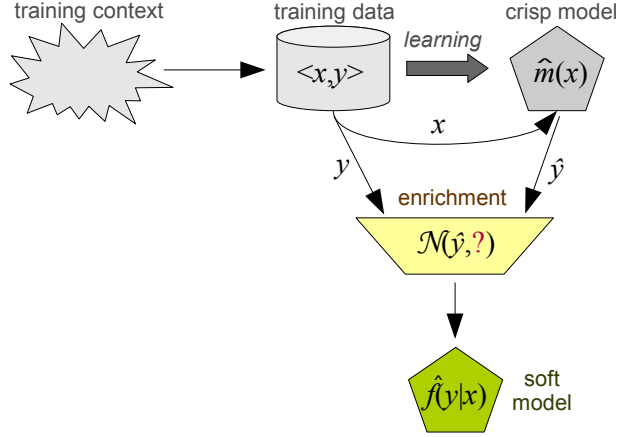

Figure 3: Enrichment methods convert a crisp regression model into a soft regression model by just comparing $y$ with $\hat{y}$. The mean of the resulting conditional density function $\hat{f}(y|x)$ is not altered (the original $\hat{y}$ is kept). Only the second parameter of a normal distribution (the conditional variance) is added.

A first option is to estimate the residual $u = y - \hat{y}$ and derive $\hat{\sigma}(x)$ from it. This procedure, which does a univariate regression on the residuals given the outputs, is called *residual-based enrichment, RBE*. We detail the *RBE* procedure below:

**Definition 3.** *Given an existing regression model ($m_y$), a training or validation set $T$, and a (test) instance $x$, the* residual-based enrichment method (*RBE*) *is defined as follows:*

1. *Obtain $\hat{y}_i = m_y(x_i)$ for each example $\langle x_i, y_i \rangle \in T$.*
2. *Calculate the residuals: $u_i \leftarrow (y_i - \hat{y}_i)$.*
3. *Apply a transformation function $\theta$ to the residuals: $v_i \leftarrow \theta(u_i)$.*
4. *Train a regression model $m_v$ for the dataset $V = \{\langle \hat{y}_i, v_i \rangle\}$.*
5. *Obtain $\hat{y} = m_y(x)$ and $\hat{v} = m_v(\hat{y})$ for the example $x$ to be predicted (in the test set).*

*So, for each example $x$ in the test set, the estimated conditional mean for that example is $\hat{\mu}(x) = \hat{y}$ and the estimated conditional standard deviation is $\hat{\sigma}(x) = \theta^{-1}(\hat{v})$. Note that steps 1 to 4 can be omitted if we just train and keep $m_v$.*

The procedure is similar to the conditional variance estimation methods shown in appendix D, but we remove the dependency on $x$ for the residual model. This procedure also resembles some calibration methods in classification. Platt's method [53] applies a univariate function (a sigmoid) to the outputs, in order to calibrate them. Finally, it also slightly resembles some the idea of mimetic models [30, 21]

In order to apply the *RBE* method, we only need to choose an appropriate transformation function $\theta$ for step 3 and a regression technique for step 4. The transformation function $\theta$ can be used at convenience to ensure that $\hat{\sigma}(x)$ is always positive or to make an estimation of absolute or squared residuals. Several possibilities exist, but a natural choice is $\theta(t) = t^2$, if seen as a variance estimation method [69, 67].

|  | LR ENR kNN msll | LR ENR kNN msvr | LR ENR Tree msll | LR ENR Tree msvr | kNN ENR kNN msll | kNN ENR kNN msvr | kNN ENR Tree msll | kNN ENR Tree msvr | Tree ENR kNN msll | Tree ENR kNN msvr | Tree ENR Tree msll | Tree ENR Tree msvr |
|---|---|---|---|---|---|---|---|---|---|---|---|---|
| 1 | 0.78 | 0.49 | 0.78 | 0.51 | 0.80 | 0.48 | 0.80 | 0.50 | 0.84 | 0.62 | 0.85 | 0.68 |
| 2 | 0.78 | 0.55 | 0.78 | 0.54 | 0.77 | 0.54 | 0.76 | 0.54 | 0.69 | 0.55 | 0.66 | 0.52 |
| 3 | 0.53 | 0.69 | 0.55 | 0.71 | 0.61 | 0.63 | 0.62 | 0.63 | 0.69 | 0.63 | 0.60 | 0.46 |
| 4 | 0.71 | 0.57 | 0.72 | 0.58 | 0.83 | 0.60 | 0.81 | 0.47 | 0.77 | 0.50 | 0.81 | 0.61 |
| 5 | 0.63 | 0.54 | 0.63 | 0.53 | 0.63 | 0.52 | 0.64 | 0.53 | 0.70 | 0.64 | 0.63 | 0.51 |
| 6 | 0.71 | 0.62 | 0.70 | 0.61 | 0.65 | 0.57 | 0.66 | 0.59 | 0.68 | 0.67 | 0.62 | 0.57 |
| 7 | 0.76 | 0.68 | 0.76 | 0.63 | 0.77 | 0.62 | 0.76 | 0.60 | 0.76 | 0.57 | 0.76 | 0.56 |
| 8 | 0.64 | 0.53 | 0.64 | 0.53 | 0.62 | 0.54 | 0.61 | 0.53 | 0.64 | 0.62 | 0.58 | 0.54 |
| 9 | 0.95 | 0.91 | 0.94 | 0.90 | 1.00 | 0.94 | 0.99 | 0.90 | 1.00 | 0.96 | 1.00 | 0.95 |
| 10 | 0.69 | 0.39 | 0.70 | 0.43 | 0.61 | 0.52 | 0.67 | 0.65 | 0.60 | 0.59 | 0.57 | 0.56 |
| 11 | 0.72 | 0.51 | 0.74 | 0.57 | 0.76 | 0.68 | 0.76 | 0.69 | 0.77 | 0.76 | 0.68 | 0.65 |
| 12 | 0.86 | 0.56 | 0.86 | 0.57 | 0.76 | 0.49 | 0.76 | 0.53 | 0.83 | 0.74 | 0.72 | 0.56 |
| 13 | 0.57 | 0.64 | 0.58 | 0.66 | 0.90 | 0.63 | 0.90 | 0.63 | 0.92 | 0.80 | 0.90 | 0.67 |
| 14 | 0.74 | 0.46 | 0.74 | 0.47 | 0.71 | 0.50 | 0.71 | 0.50 | 0.70 | 0.63 | 0.65 | 0.51 |
| 15 | 0.75 | 0.64 | 0.74 | 0.61 | 0.75 | 0.61 | 0.73 | 0.57 | 0.82 | 0.71 | 0.72 | 0.56 |
| 16 | 0.76 | 0.55 | 0.75 | 0.52 | 0.71 | 0.54 | 0.72 | 0.56 | 0.68 | 0.57 | 0.68 | 0.58 |
| 17 | 0.98 | 0.88 | 0.97 | 0.87 | 0.80 | 0.59 | 0.83 | 0.73 | 0.77 | 0.67 | 0.76 | 0.63 |
| 18 | 0.74 | 0.54 | 0.74 | 0.56 | 0.75 | 0.55 | 0.75 | 0.56 | 0.76 | 0.60 | 0.74 | 0.54 |
| 19 | 0.69 | 0.52 | 0.69 | 0.55 | 0.70 | 0.51 | 0.70 | 0.51 | 0.71 | 0.58 | 0.69 | 0.55 |
| 20 | 0.79 | 0.59 | 0.79 | 0.61 | 0.68 | 0.53 | 0.69 | 0.55 | 0.72 | 0.62 | 0.69 | 0.58 |
| Mean | 0.74 | 0.59 | 0.74 | 0.60 | 0.74 | 0.58 | 0.74 | 0.59 | 0.75 | 0.65 | 0.72 | 0.59 |

Table 3: Results (using the datasets in Table 11) for several base techniques (*LR*, *kNN* and *Tree*) with the residual-based enrichment (*RBE*) methods using *kNN* and *Tree* as models for the residuals. All the methods use $\theta(t) = t^2$. Results for *mrse* are not shown since they are equal to Table 2.

We explore the *RBE* method for the base techniques (*LR*, *kNN* and *Tree*) and two methods for calculating the residuals (*kNN* and *Tree*). Table 3 shows the results. We see that the results are now similar for all the base techniques, quite differently to the results in Table 2.

Given that *enrichment* only requires a univariate regression technique, we can look for simpler and equally effective approaches, without the need of using a second regression technique, such as *kNN* and *Tree*. A single approach is *binning*, which just uses a sliding window over the estimated value $\hat{y}$. This approach resembles binning calibration in classification [3] More formally, the *BIN* method is defined as follows:

**Definition 4.** *Given an existing regression model ($m_y$), a training or validation set T, and a (test) instance x, the enrichment method BIN is defined as follows:*

1. *Obtain $\hat{y}_i = m_y(x_i)$ for each example $\langle x_i, y_i \rangle \in T$.*
2. *Calculate the residuals: $u_i \leftarrow (y_i - \hat{y}_i)$.*
3. *Apply a transformation function $\theta$ to the residuals: $v_i \leftarrow \theta(u_i)$.*
4. *Construct a dataset $V = \{\langle \hat{y}_i, v_i \rangle\}$.*
5. *Sort V by $\hat{y}_i$.*
6. *Obtain $\hat{y} = m_y(x)$ for the example x to be predicted (in the test set).*
7. *Construct the set W with the $k/2$ values $v_i$ in V immediately above $\hat{y}$ and the $k/2$ values $v_i$ in V immediately below[2].*
8. *Obtain $\hat{v}$ as the mean of W.*

---
[2] If there are not sufficient elements above or below we take as many as we can.

*The estimated conditional mean is $\hat{\mu}(x) = \hat{y}$ and the estimated conditional standard deviation is $\hat{\sigma}(x) = \theta^{-1}(\hat{v})$. Note that steps 1 to 4 can be omitted (and the training set is no longer necessary) if we just keep the dataset V when training the model.*

A third enrichment method can be defined by constructing the bins using distances and then averaging the deviation of the *true* values against the prediction (instead of averaging the residuals). This method is an univariate version of the method *KNC* (see appendix E):

**Definition 5.** *Given an existing regression model ($m_y$), a train or validation set $T$, and a (test) instance $x$, the univariate k-nearest comparison enrichment method uKNC is defined as follows:*

1. *Obtain $\hat{y}_i = m_y(x_i)$ for each example $\langle x_i, y_i \rangle \in T$.*
2. *Construct a dataset $Q = \{\langle \hat{y}_i, y_i \rangle\}$.*
3. *Obtain $\hat{y} = m_y(x)$ for the example x to be predicted (in the test set).*
4. *Let $S = \langle \hat{y}_j, y_j \rangle$ the set of the k nearest neighbours in Q (using the distance $|\hat{y}_i - \hat{y}|$ between each $\hat{y}_i$ in Q and the fixed $\hat{y}$).*
5. *Obtain $\hat{s}^2 = \frac{1}{k} \sum_{\langle \hat{y}_j, y_j \rangle \in S} (\hat{y} - y_j)^2$.*

*The estimated conditional mean is $\hat{\mu}(x) = \hat{y}$ and the estimated conditional variance is $\hat{\sigma}(x)^2 = \hat{s}^2$.*

Note that this method is different froom *RBE* using *kNN*. The method *uKNC* just looks for the closest estimations in the training set to the estimation for example *x* and compares their true values with the estimation for *x*. Note that the rationale behind this method is that we link the variance to the estimations, i.e., given a set of *k* examples with similar estimations, we calculate how far (on average) the true values are to the centre estimation. By using the centre estimation and not the estimation for each of the *k* examples with most similar estimations, this method can be more robust (since an outlier estimation for one of the *k* estimations has no effect on the result).

Again, we apply these two methods (*BIN* and *uKNC*) to the base techniques (*LR*, *kNN* and *Tree*). Table 4 shows the results. As we see, the performance is not degraded at all by these extremely straightforward and efficient methods. Much on the contrary, their results are good, especially for *uKNC*.

## 3.4 Choosing some appropriate *NCDE* methods for cost-sensitive applications

As we have seen, the number of possible methods which can be used to derive a simple (i.e. normal) conditional density function is really large, and some of them could be parameterised and refined. Nonetheless, our goal was to select a small set of simple *NCDE* methods that could produce a reasonably good normal conditional density estimation or, more precisely, a good pair of conditional mean and conditional variance (from which a normal conditional density estimation is built).

In order to make a selection, we have analysed some of the methods seen so far in order to find a small subset of methods with the following criteria: good performance (for any base regression technique), low dependence on the training set and efficiency. Performance results and significance tests are shown in tables 17, 18 and 19 in appendix G. These tables also include the results for some of the methods mentioned in section 3.2 (which are explained in full detail in appendices C, D, E and F). According to these results and the previous criteria, we decide to use the following *NCDE* methods:

- *Own*: uses the *own* variance estimation methods from each base regression technique (section 3.1).

- *uKNC*: uses the *univariate k-nearest comparison* enrichment method (definition 5, section 3.3).

|   | LR ENR uKNC msll | LR ENR uKNC msvr | kNN ENR uKNC msll | kNN ENR uKNC msvr | Tree ENR uKNC msll | Tree ENR uKNC msvr | LR ENR BIN msll | LR ENR BIN msvr | kNN ENR BIN msll | kNN ENR BIN msvr | Tree ENR BIN msll | Tree ENR BIN msvr |
|---|---|---|---|---|---|---|---|---|---|---|---|---|
| 1 | 0.78 | 0.49 | 0.80 | 0.48 | 0.84 | 0.64 | 0.78 | 0.50 | 0.80 | 0.48 | 0.84 | 0.63 |
| 2 | 0.78 | 0.54 | 0.77 | 0.56 | 0.68 | 0.56 | 0.78 | 0.54 | 0.77 | 0.55 | 0.66 | 0.52 |
| 3 | 0.57 | 0.82 | 0.63 | 0.66 | 0.64 | 0.52 | 0.54 | 0.70 | 0.64 | 0.66 | 0.63 | 0.52 |
| 4 | 0.77 | 0.70 | 0.84 | 0.62 | 0.77 | 0.49 | 0.72 | 0.58 | 0.81 | 0.52 | 0.81 | 0.64 |
| 5 | 0.63 | 0.54 | 0.63 | 0.52 | 0.63 | 0.53 | 0.63 | 0.54 | 0.63 | 0.52 | 0.63 | 0.52 |
| 6 | 0.71 | 0.61 | 0.65 | 0.57 | 0.64 | 0.61 | 0.71 | 0.62 | 0.66 | 0.59 | 0.65 | 0.62 |
| 7 | 0.75 | 0.55 | 0.77 | 0.62 | 0.75 | 0.49 | 0.75 | 0.67 | 0.77 | 0.61 | 0.76 | 0.58 |
| 8 | 0.64 | 0.53 | 0.61 | 0.53 | 0.59 | 0.54 | 0.64 | 0.53 | 0.61 | 0.53 | 0.59 | 0.55 |
| 9 | 0.73 | 0.65 | 0.99 | 0.91 | 0.93 | 0.78 | 0.95 | 0.91 | 0.99 | 0.92 | 1.00 | 0.95 |
| 10 | 0.67 | 0.30 | 0.60 | 0.49 | 0.56 | 0.49 | 0.69 | 0.38 | 0.60 | 0.48 | 0.56 | 0.47 |
| 11 | 0.72 | 0.51 | 0.74 | 0.64 | 0.68 | 0.64 | 0.72 | 0.51 | 0.75 | 0.65 | 0.68 | 0.64 |
| 12 | 0.85 | 0.49 | 0.76 | 0.50 | 0.70 | 0.50 | 0.86 | 0.55 | 0.75 | 0.49 | 0.74 | 0.60 |
| 13 | 0.72 | 0.82 | 0.90 | 0.65 | 0.89 | 0.63 | 0.58 | 0.65 | 0.89 | 0.59 | 0.90 | 0.66 |
| 14 | 0.74 | 0.46 | 0.71 | 0.51 | 0.65 | 0.51 | 0.74 | 0.46 | 0.71 | 0.51 | 0.65 | 0.53 |
| 15 | 0.75 | 0.64 | 0.75 | 0.61 | 0.74 | 0.60 | 0.75 | 0.65 | 0.75 | 0.61 | 0.73 | 0.57 |
| 16 | 0.74 | 0.48 | 0.71 | 0.53 | 0.69 | 0.58 | 0.75 | 0.54 | 0.71 | 0.54 | 0.68 | 0.57 |
| 17 | 0.91 | 0.51 | 0.80 | 0.58 | 0.77 | 0.65 | 0.98 | 0.89 | 0.80 | 0.62 | 0.76 | 0.65 |
| 18 | 0.74 | 0.55 | 0.75 | 0.56 | 0.74 | 0.55 | 0.74 | 0.54 | 0.75 | 0.55 | 0.74 | 0.53 |
| 19 | 0.69 | 0.52 | 0.70 | 0.51 | 0.70 | 0.58 | 0.69 | 0.52 | 0.70 | 0.51 | 0.69 | 0.55 |
| 20 | 0.79 | 0.58 | 0.68 | 0.52 | 0.69 | 0.59 | 0.79 | 0.59 | 0.68 | 0.52 | 0.68 | 0.57 |
| Mean | 0.73 | 0.57 | 0.74 | 0.58 | 0.71 | 0.57 | 0.74 | 0.59 | 0.74 | 0.57 | 0.72 | 0.59 |

Table 4: Results (using the datasets in Table 11) for several base methods (*LR*, *kNN* and *Tree*) with the enrichment method *uKNC* and the enrichment method using binning for the residuals (*BIN*). Method *BIN* uses $\theta(t) = t^2$. Results for *mrse* are not shown since they are equal to Table 2.

- *BIN*: uses the residual-based enrichment method using *binning* (definition 4, section 3.3).

Note that we make a selection for the clarity of exposition. If other *NCDE* methods (either directly or by enrichment) are eventually found to perform better or more efficiently (in general or for a particular problem), this will give further support for (and improve) the probabilistic reframing methods that we will explore in the following sections.

## 4 Bid applications

As explained in the introduction, most regression problems require the minimisation of a loss function representing the cost context, rather than an uncontextualised (quadratic) error. Frequently, this loss function is only known at deployment time, so training is usually performed without this information, as shown in Figure 1. Given the selection of *NCDE* methods at the end of previous section, we are ready to apply reframing to several kinds of cost-sensitive decision problems, where particular families of loss functions are used. For instance, in this section we will explore a family of problems which are very common in econometrics, commerce and retailing applications, where we need to estimate the price (or other quantifiable features) for an offer or bid in the context of a sale, deal or auction ([59, 18, 47, 68, 36]). One of the most relevant features of the loss functions in these applications is that they are highly discontinuous, since an offer which is much too expensive changes loss dramatically: from the maximum attainable benefit to no benefit at all (the offer is not accepted). This is formalised by the following bid loss function:

**Definition 6.** *The bid loss $\ell_\beta^B$ is a loss function defined as follows:*

$$\ell_\beta^B(\hat{y}, y) = -\hat{y} + \beta \quad \text{if } \hat{y} \leq y$$
$$= 0 \quad \text{otherwise}$$

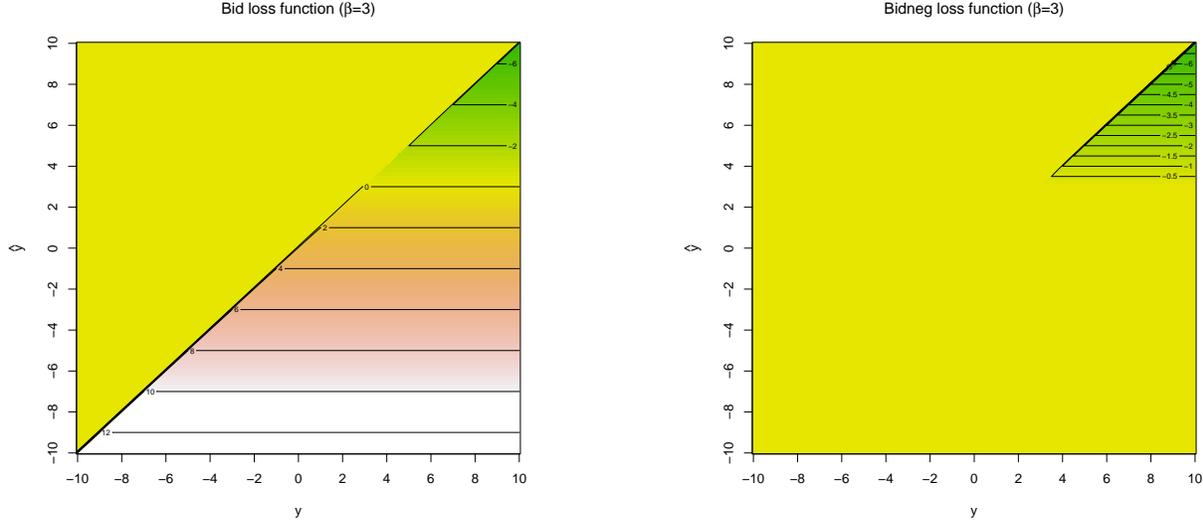

Figure 4: Two loss functions shown as a scatter plot with actual output values on the *x*-axis and estimated output values on the *y*-axis. Costs are shown with contour lines and colours (from benefits to high costs represented with the scale green-yellow-reddish-white). Left: Bid loss function with $\beta = 3$. Right: Bidneg loss function with $\beta = 3$.

*where $\beta$ represents some kind of* base *cost. If $\hat{y} > \beta$ then we have positive profits.*

Figure 4 (left) shows a representation of this loss function with $\beta = 3$. Given the bid loss and a *NCDE* model $\hat{f}(y|x)$, we need to determine the *optimal local reframing*, which get the lowest expected loss (minimum risk). This can be done as follows:

**Proposition 2.** *Given $\ell_\beta^B$,*
$$r^*(x, \ell_\beta^B, \hat{f}) = \arg\min_t \left\{ (\beta - t)(1 - \hat{F}(t|x)) \right\}$$

*Proof.* From eq. (3), we have that $r^*(x, \ell_\beta^B, \hat{f})$ can be written as follows:

$$\begin{aligned}
r^*(x, \ell_\beta^B, \hat{f}) &= \arg\min_t \int_{-\infty}^{\infty} \ell_\beta^B(t, y) \hat{f}(y|x) dy \\
&= \arg\min_t \left\{ \int_{-\infty}^{t} 0 + \int_{t}^{\infty} (-t + \beta) \hat{f}(y|x) dy \right\} \\
&= \arg\min_t \left\{ (\beta - t)(1 - \hat{F}(t|x)) \right\}
\end{aligned}$$

□

The previous equation has no closed form in general (and it does not reduce either for the normal distribution). For most distributions (exceptions are fat-tailed distributions, such as the Cauchy distribution), the value of the estimated cumulative distribution function $\hat{F}(t|x)$ goes to 1 faster than *t* grows to infinity. So this expression is 0 for $t \to \infty$. Hence, in general, it only has a minimum. We can find the minimum of the previous function numerically or we can calculate the derivative $(t - \beta)\hat{f}(t|x) + \hat{F}(t|x) - 1$ and try to find (also numerically) the values which make the expression 0 and see which of them are minima. The first option seems the easiest one, especially for a normal distribution. While the loss function is discontinuous, the expression in proposition 2 is not, and some efficient numerical methods can be used.

Apart from the *local* reframing using *NCDE* methods and proposition 2 seen above, we will compare with a *global* reframing which just uses the expected value (a crisp regression model) and adds a shift which has been optimised for the training set. The methods are then:

- *None*: No reframing. The prediction (conditional estimated mean) is used as it is.

- *Own, uKNC, BIN*: Probabilistic (local) reframing using a numerical approximation for the expression $r^*(x, \ell_\beta^B, \hat{f})$ with a normal distribution, as given by proposition 2. The conditional normal density is obtained by three different methods: as derived by the base technique (*Own*) and the enrichment methods *uKNC* and *BIN*.

- *CoSh*: Global reframing using a *constant shift* $s_0$ for all the predictions: $R^+(x, \ell_\beta^B, \hat{f}) = \mathbb{E}_{\hat{f}}(y|x) + s_0$. In order to calculate a good shift, we look for the best shift for the whole training set[3]. The calculation of $s_0$ can be done numerically. Since $\ell_\beta^B$ is discontinuous, we cannot use many optimisation methods and we need to use a Monte Carlo algorithm or a resolution-bounded covering algorithm, assuming that the solution is inside a (wide) interval.

Now we see how the previous methods perform. We will use several values of $\beta$ using the equation $\beta = (max_y - min_y) \cdot a^2$, where $a$ ranges regularly between 0 and 1, and $max_y$ and $min_y$ are the maximum and minimum values of the output $y$ for the whole dataset. The equation tries to capture a range of reasonable cases for this family of problems. The rationale is that high values of $\beta$ imply that benefits can only be obtained with values of $\hat{y}$ which get close to $max_y$, while low values of $\beta$ imply that we will almost always get benefits. This is the reason why we have squared $a$, in order to make cases with low $\beta$ more frequent, if we just choose $a$ regularly. With this, we explore different reasonable possibilities for $\beta$.

Figure 5 (left) shows the evolution of this loss for different methods and different values of $\beta$ (which is a function of $a$) for *one* dataset as an illustration (the figures may vary significantly for other datasets).

The overall results (Table 5) show that an appropriate cost-sensitive probabilistic (local) reframing outperforms a constant shift method (global reframing). The results are consistent for the three different base techniques (*LR*, *kNN* and *Tree*).

After this first loss function and its results for different methods, we can of course figure out other related loss functions. For instance, a common variant of the bid loss is when the decision rule does not make a bid if we expect no benefit. This is not a rejection rule, which we will see in section 6, but means that there is no offer, no sale and, hence, no profit or loss. In many applications, this is a more realistic loss function, and can be defined as follows.

**Definition 7.** *The non-losing bid loss $\ell_\beta^{\bar{B}}$ is a loss function defined as follows:*

$$\ell_\beta^{\bar{B}}(\hat{y}, y) = -\hat{y} + \beta \quad \text{if } (\hat{y} \leq y) \wedge (\beta \leq \hat{y})$$
$$= 0 \quad \text{otherwise}$$

Figure 4 (right) shows a representation of this loss function with $\beta = 3$. We can get its optimal reframing as we did for $\ell_\beta^B$:

**Proposition 3.** *Given $\ell_\beta^{\bar{B}}$,*

$$r^*(x, \ell_\beta^{\bar{B}}, \hat{f}) = \arg\min_t \left\{ (\beta - t)(1 - \hat{F}(\max(\beta, t)|x)) \right\}$$

---

[3]Note that this method does not require the estimation of a conditional normal distribution (only the expected value is needed), so any crisp regression method can be used directly. However, it requires the complete training set (or at least the actual output values $y$) for every new context (loss function). This might not be possible in many applications. It also assumes the same parameters for the loss function for the whole dataset.

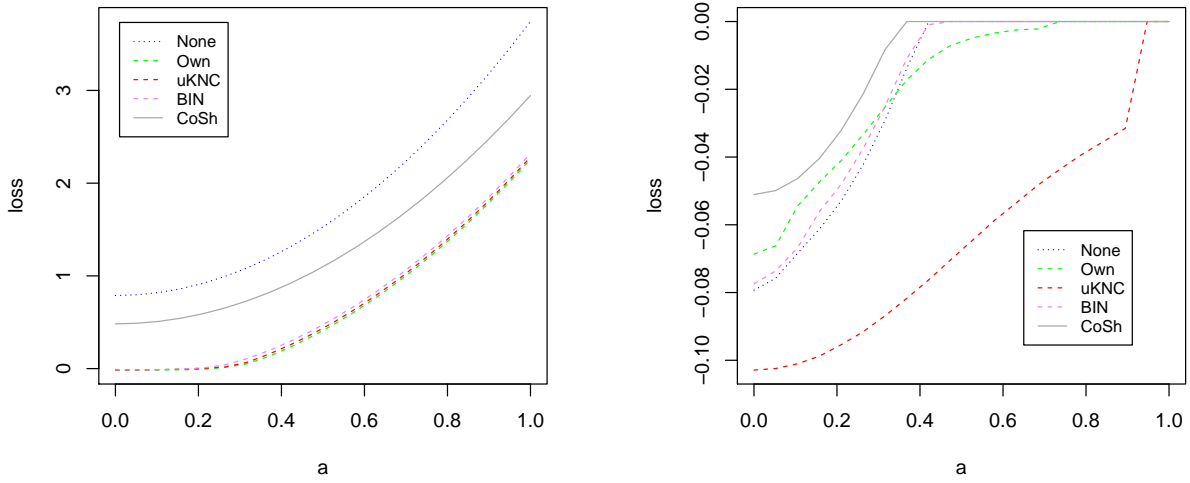

Figure 5: Left: comparing the bid loss using different methods for dataset *rock* with base technique *kNN*. Right: comparing the bidneg loss using different methods for dataset *menarche* with base technique *LR*.

*Proof.* From eq. (3), we have that $r^*(x, \ell_\beta^{\bar{B}}, \hat{f})$ can be written as follows:

$$
\begin{aligned}
r^*(x, \ell_\beta^{\bar{B}}, \hat{f}) &= \arg\min_t \int_{-\infty}^{\infty} \ell_\beta^{\bar{B}}(t,y) \hat{f}(y|x) dy \\
&= \arg\min_t \left\{ \int_{-\infty}^{\max(\beta,t)} 0 + \int_{\max(\beta,t)}^{\infty} (-t+\beta) \hat{f}(y|x) dy \right\} \\
&= \arg\min_t \left\{ (\beta - t)(1 - \hat{F}(\max(\beta,t)|x)) \right\} \quad (4)
\end{aligned}
$$

□  □

Figure 5 (right) shows the evolution of this loss for different values of $\beta$ (which is a function of *a*) for one dataset. The overall results for this variant are shown in Table 6 for several base techniques, with the same configuration as Table 5. The results are similar to those in Table 5, although the differences between the methods are now smaller since there are many more cases where the loss is 0. Again this shows that the use of an appropriate cost-sensitive probabilistic reframing gets better results than a constant shift method (global reframing). The results are again consistent for different base techniques.

While we only show the results for some typical bid functions as for definitions 6 and 7, the same idea can be used in applications where there can be more bids (see, e.g., [4]), or when auctions work in a different way. In that case, a similar result to propositions 2 and 3 could be obtained by applying the maximisation recursively. In the cases where the expressions cannot be simplified into a closed form, as in this case, we can use a numerical method.

As mentioned above, the global reframing method cannot be applied when the loss function parameters may be different for each example. For instance, the $\beta$ parameter of the bid function may depend on the instance, such as cases where this represents the production cost and is different for each product. The probabilistic reframing can still be applied in these cases.

|    | LR None | LR Own | LR uKNC | LR BIN | LR CoSh | kNN None | kNN Own | kNN uKNC | kNN BIN | kNN CoSh | Tree None | Tree Own | Tree uKNC | Tree BIN | Tree CoSh |
|----|---------|--------|---------|--------|---------|----------|---------|----------|---------|----------|-----------|----------|-----------|----------|-----------|
| 1  | 2.45    | -0.19  | -0.75   | -0.77  | -0.05   | 3.97     | -0.33   | -0.37    | -0.38   | 0.88     | 3.78      | -0.39    | 0.01      | -0.30    | 0.72      |
| 2  | 10.85   | 0.26   | -0.25   | -0.23  | 0.97    | 7.25     | -0.29   | -0.25    | -0.34   | 0.12     | 13.82     | 0.09     | 0.11      | 0.16     | 0.87      |
| 3  | 6.99    | -0.42  | -0.56   | -0.52  | -0.39   | 6.08     | -0.64   | -0.65    | -0.64   | -0.40    | 6.81      | -0.52    | -0.57     | -0.46    | 0.32      |
| 4  | 6.64    | 1.89   | 0.51    | 3.68   | 5.61    | 9.14     | 3.52    | 3.61     | 3.75    | 6.74     | 9.47      | 3.87     | 4.17      | 4.31     | 6.88      |
| 5  | 17.93   | -0.23  | -0.98   | 0.37   | 13.17   | 10.14    | 2.46    | 3.26     | 2.35    | 8.40     | 9.93      | 2.52     | 1.54      | 2.61     | 7.17      |
| 6  | 6.50    | -0.16  | -0.42   | -0.41  | 0.06    | 5.70     | -0.37   | -0.42    | -0.36   | -0.20    | 6.22      | -0.24    | -0.38     | -0.21    | 0.14      |
| 7  | 3.42    | 1.97   | -0.20   | -0.24  | 1.96    | 5.54     | -0.27   | -0.30    | -0.31   | 1.64     | 3.42      | -0.09    | 0.13      | 0.08     | 2.24      |
| 8  | 8.25    | 0.10   | -0.32   | 0.34   | 7.34    | 10.90    | 0.63    | 0.65     | 0.64    | 9.98     | 10.58     | 0.65     | 0.55      | 2.88     | 10.07     |
| 9  | 19.96   | 0.58   | -0.09   | 0.58   | 15.45   | 11.94    | -0.23   | -0.16    | -0.07   | 0.16     | 5.82      | -0.23    | -0.18     | -0.14    | 0.43      |
| 10 | 3.75    | 0.01   | -0.54   | -0.18  | 2.40    | 3.58     | -0.51   | -0.48    | -0.48   | -0.16    | 1.87      | -0.56    | -0.60     | -0.51    | -0.17     |
| 11 | 9.03    | 5.58   | 3.31    | 5.58   | 8.47    | 9.93     | 5.58    | 5.58     | 5.58    | 9.30     | 9.92      | 5.58     | 5.58      | 5.58     | 9.58      |
| 12 | 9.90    | -0.37  | -0.78   | -0.79  | 0.02    | 9.29     | -0.77   | -0.75    | -0.76   | -0.14    | 10.81     | -0.67    | -0.67     | -0.67    | 1.13      |
| 13 | 17.18   | 0.48   | -0.41   | -0.41  | -0.06   | 16.55    | -0.38   | -0.37    | -0.38   | -0.25    | 18.88     | -0.34    | -0.36     | -0.37    | 0.25      |
| 14 | 6.83    | 0.95   | -0.24   | -0.26  | 0.67    | 7.56     | -0.23   | -0.32    | -0.32   | 0.26     | 8.64      | 0.00     | -0.02     | 0.10     | 2.71      |
| 15 | 13.24   | -0.16  | -0.10   | -0.14  | 0.42    | 9.02     | -0.30   | -0.22    | -0.33   | 0.34     | 8.44      | -0.17    | -0.25     | -0.13    | 0.82      |
| 16 | 15.98   | 3.32   | -0.37   | -0.37  | -0.15   | 17.37    | -0.41   | -0.37    | -0.39   | -0.27    | 16.60     | -0.37    | -0.36     | -0.33    | -0.21     |
| 17 | 1.14    | 0.06   | -0.29   | 0.18   | 0.91    | 2.89     | 0.60    | 0.68     | 0.65    | 1.89     | 2.80      | 0.84     | 0.60      | 0.93     | 2.22      |
| 18 | 2.17    | 1.48   | -0.29   | -0.36  | 1.07    | 1.95     | -0.32   | -0.31    | -0.34   | 1.04     | 1.63      | 0.33     | -0.36     | 0.58     | 1.31      |
| 19 | 6.83    | -0.50  | -0.37   | -0.51  | 0.51    | 9.46     | -0.51   | -0.62    | -0.54   | -0.07    | 4.02      | -0.51    | -0.55     | -0.49    | 0.26      |
| 20 | 9.07    | -0.58  | -0.61   | -0.56  | 2.62    | 8.05     | -0.59   | -0.53    | -0.61   | 0.19     | 8.64      | -0.50    | -0.51     | -0.48    | 0.90      |
| AR | 5.00    | 2.75   | **1.45**| 2.05   | 3.75    | 5.00     | **1.80**| 2.30     | **1.90**| 4.00     | 5.00      | **1.75** | **1.50**  | 2.75     | 4.00      |

Table 5: Results for the bid loss $\ell_\beta^B$ for the datasets in Table 12, using the experimental methodology in section 2.5. Each row aggregates the folds and ten different values for $\beta$ per fold using the formula $\beta = (max_y − min_y) \cdot a^2$ with $a \in \{0, 0.111, 0.222, \ldots, 1\}$. For visibility all the losses are multiplied by 10. Each section of five columns shows results for different base techniques (*LR*, *kNN* and *Tree*). The average ranks (AR) are calculated for these three groups separately. The Friedman statistics for the three sections are (63.44, 65.12 and 71 respectively), which are greater than the Critical Value (10.92). This means that the null hypothesis is rejected (significance level: 0.05) and the methods do not perform equally. Differences in average ranks higher than the critical difference for the Nemenyi post-hoc test (0.3626) imply that the difference is significant (in bold).

## 5 Asymmetric loss applications

As mentioned in the introduction, many regression problems do not have a symmetric loss. Depending on the application, overestimations might be worse than underestimations (or vice versa). The way in which this asymmetry is modelled has led to the definition of many asymmetric loss functions, such as *Lin-Exp* (approximately linear on one side and exponential on the other side), *Quad-Exp* (approximately quadratic on one side and exponential on the other side), *Lin-Lin* (asymmetric linear) and *Quad-Quad* (asymmetric quadratic). We will focus on the latter two since these are more common and can be seen as generalisations of absolute error and quadratic error respectively.

First, we give a definition for the asymmetric absolute error $\ell_\alpha^A$:

**Definition 8.** *The asymmetric absolute error $\ell_\alpha^A$ is a loss function defined as follows:*

$$\ell_\alpha^A(\hat{y}, y) = \alpha(y - \hat{y}) \quad \text{if } \hat{y} < y$$
$$= (1 - \alpha)(\hat{y} - y) \quad \text{otherwise}$$

with $\alpha$ being the cost proportion (or asymmetry) between 0 and 1, with increasing values meaning higher cost for *low predictions* (underestimation). In other words, when $\alpha = 0$ we mean that predictions below the actual value have no cost. When $\alpha = 1$ we mean that predictions above the actual value have no cost. When $\alpha = 0.5$ we mean that costs above and below are symmetric.

Similarly, we give the definition for the asymmetric squared error:

|    | LR None | LR Own | LR uKNCBIN | LR | LR CoSh | kNN None | kNN Own | kNN uKNCBIN | kNN | kNN CoSh | Tree None | Tree Own | Tree uKNCBIN | Tree | Tree CoSh |
|----|---------|--------|------------|------|---------|----------|---------|-------------|------|----------|-----------|----------|--------------|------|-----------|
| 1  | -0.64 | -0.71 | -0.75 | -0.77 | -0.73 | -0.26 | -0.35 | -0.37 | -0.38 | -0.55 | -0.30 | -0.39 | -0.33 | -0.36 | -0.56 |
| 2  | -0.07 | -0.19 | -0.30 | -0.28 | -0.20 | -0.06 | -0.29 | -0.30 | -0.34 | -0.36 | -0.05 | -0.25 | -0.23 | -0.22 | -0.12 |
| 3  | -0.30 | -0.42 | -0.56 | -0.52 | -0.49 | -0.30 | -0.64 | -0.65 | -0.64 | -0.40 | -0.19 | -0.52 | -0.57 | -0.46 | -0.49 |
| 4  | 0.00 | -0.06 | -0.18 | -0.01 | 0.00 | 0.00 | -0.02 | -0.02 | -0.02 | 0.00 | 0.00 | -0.01 | -0.01 | -0.00 | 0.00 |
| 5  | -0.29 | -0.29 | -0.98 | -0.28 | -0.01 | 0.00 | -0.02 | -0.02 | -0.03 | 0.00 | 0.00 | -0.02 | -0.05 | -0.02 | 0.00 |
| 6  | -0.26 | -0.35 | -0.42 | -0.41 | -0.20 | -0.03 | -0.37 | -0.42 | -0.36 | -0.35 | -0.20 | -0.34 | -0.38 | -0.30 | -0.16 |
| 7  | -0.08 | -0.11 | -0.23 | -0.24 | -0.31 | -0.14 | -0.27 | -0.30 | -0.31 | -0.34 | -0.11 | -0.21 | -0.19 | -0.19 | -0.15 |
| 8  | -0.11 | -0.10 | -0.32 | -0.10 | -0.07 | 0.00 | -0.01 | -0.01 | -0.01 | 0.00 | 0.00 | -0.01 | -0.01 | -0.00 | 0.00 |
| 9  | -0.10 | -0.10 | -0.09 | -0.10 | -0.13 | -0.03 | -0.23 | -0.16 | -0.14 | -0.19 | -0.09 | -0.23 | -0.18 | -0.18 | -0.31 |
| 10 | -0.22 | -0.22 | -0.54 | -0.24 | -0.47 | -0.03 | -0.51 | -0.48 | -0.48 | -0.54 | -0.24 | -0.56 | -0.60 | -0.51 | -0.55 |
| 11 | 0.00 | 0.00 | -0.08 | 0.00 | 0.00 | 0.00 | 0.00 | 0.00 | 0.00 | 0.00 | 0.00 | 0.00 | 0.00 | 0.00 | 0.00 |
| 12 | -0.59 | -0.77 | -0.78 | -0.79 | -0.79 | -0.67 | -0.77 | -0.75 | -0.76 | -0.69 | -0.60 | -0.67 | -0.67 | -0.67 | -0.59 |
| 13 | -0.11 | -0.19 | -0.41 | -0.41 | -0.35 | -0.09 | -0.38 | -0.40 | -0.40 | -0.33 | -0.12 | -0.37 | -0.38 | -0.37 | -0.30 |
| 14 | -0.01 | -0.09 | -0.32 | -0.30 | -0.47 | -0.03 | -0.34 | -0.32 | -0.32 | -0.55 | -0.02 | -0.21 | -0.21 | -0.19 | -0.04 |
| 15 | -0.19 | -0.28 | -0.17 | -0.26 | -0.15 | -0.34 | -0.30 | -0.22 | -0.33 | -0.27 | -0.27 | -0.27 | -0.25 | -0.28 | -0.22 |
| 16 | -0.29 | -0.28 | -0.39 | -0.39 | -0.39 | -0.29 | -0.41 | -0.39 | -0.40 | -0.34 | -0.32 | -0.37 | -0.36 | -0.36 | -0.32 |
| 17 | -0.20 | -0.21 | -0.29 | -0.20 | -0.18 | -0.04 | -0.05 | -0.04 | -0.05 | -0.08 | -0.06 | -0.06 | -0.07 | -0.06 | -0.06 |
| 18 | -0.24 | -0.21 | -0.29 | -0.36 | -0.52 | -0.32 | -0.32 | -0.31 | -0.34 | -0.47 | -0.41 | -0.35 | -0.36 | -0.35 | -0.37 |
| 19 | -0.25 | -0.50 | -0.37 | -0.51 | -0.25 | -0.15 | -0.51 | -0.62 | -0.54 | -0.56 | -0.36 | -0.51 | -0.55 | -0.49 | -0.48 |
| 20 | -0.39 | -0.58 | -0.61 | -0.56 | -0.36 | -0.31 | -0.59 | -0.53 | -0.61 | -0.44 | -0.35 | -0.50 | -0.51 | -0.48 | -0.28 |
| AR | 3.90 | 3.40 | **1.85** | 2.50 | 3.35 | 4.30 | **2.50** | 2.90 | **2.35** | 2.95 | 4.10 | **2.10** | **1.90** | 3.00 | 3.90 |

Table 6: Results for the bidneg loss $\ell_\beta^{\bar{B}}$ for the datasets in Table 12, using the experimental methodology in section 2.5. Each row aggregates the folds and ten different values for $\beta$ per fold using the formula $\beta = (max_y - min_y) \cdot a^2$ with $a \in \{0, 0.111, 0.222, \ldots, 1\}$. For visibility all the losses are multiplied by 10. Each section of five columns shows results for different base techniques (*LR*, *kNN* and *Tree*). The average ranks (AR) are calculated for these three groups separately. The Friedman statistics for the three sections are (21.32, 19 and 32.32 respectively), which are greater than the Critical Value (10.92). This means that the null hypothesis is rejected (significance level: 0.05) and the methods do not perform equally. Differences in average ranks higher than the critical difference for the Nemenyi post-hoc test (0.3626) imply that the difference is significant (in bold).

**Definition 9.** *The asymmetric squared error $\ell_\alpha^S$ is a loss function defined as follows:*

$$\ell_\alpha^S(\hat{y}, y) = \alpha(y - \hat{y})^2 \quad \text{if } \hat{y} < y$$
$$= (1 - \alpha)(\hat{y} - y)^2 \quad \text{otherwise}$$

Figure 6 (left and right) shows a representation of these two loss functions for $\alpha = 0.8$.

Now, we look for the optimal choice in both cases. The case for $\ell_\alpha^A$ is relatively straightforward:

**Proposition 4.** *If $\ell$ is the asymmetric absolute error function $\ell_\alpha^A(\hat{y}, y)$ given by definition 8 and $\hat{f}(y|x)$ is any conditional distribution (whose mean is denoted by $\hat{\mu}(x)$), then the expected loss for a predicted value t is given by:*

$$\mathcal{L}(x, t, \hat{f}, \ell_\alpha^A) = \alpha \hat{\mu}(x) + t\hat{F}(t|x) - \alpha t - \int_{-\infty}^{t} y\hat{f}(y|x)dy$$

*where $\hat{F}$ is the cumulative distribution for $\hat{f}$.*

From now on, we omit all the proofs, which can be found in appendix H.
The previous expression can be used to obtain the optimal prediction easily:

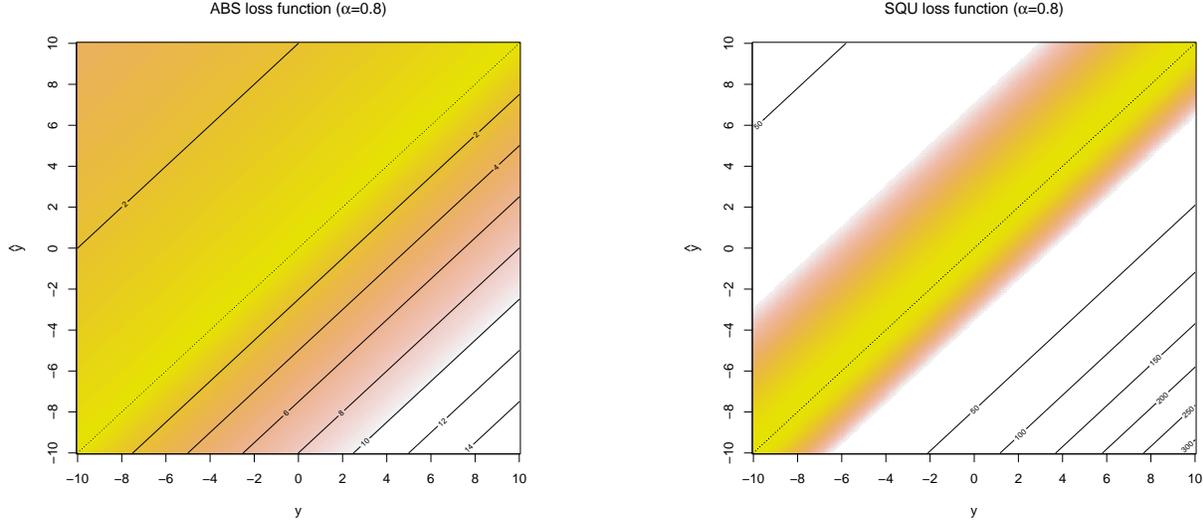

Figure 6: Two loss functions shown as a scatter plot with actual output values on the *x*-axis and estimated output values on the *y*-axis. Costs are shown with contour lines and colours (from benefits to high costs represented with the scale green-yellow-reddish-white). Left: Asymmetric absolute (Lin-Lin) loss function (ABS) with $\alpha = 0.8$. Right: Asymmetric squared (Quad-Quad) loss function (SQU) with $\alpha = 0.8$.

**Proposition 5.** *If $\ell$ is the asymmetric absolute error function $\ell_\alpha^A(\hat{y}, y)$ given by definition 8 and $\hat{f}(y|x)$ is any conditional distribution, then the optimal prediction is given by the value t such that the following equality holds:*

$$\hat{F}(t|x) = \alpha \qquad (5)$$

*where $\hat{F}$ is the cumulative distribution for $\hat{f}$.*

Clearly, the previous result can be instantiated for any distribution, whose cumulative distribution is invertible, and get:

$$\hat{F}^{-1}(\alpha|x) \qquad (6)$$

where $\hat{F}^{-1}$ is the inverse of the cumulative distribution for $\hat{f}$. If $\hat{f}$ is a normal distribution then we can just use the quantile function (or probit function).

The expression in eq. 6 is easy and intuitive. For a normal distribution, if we have $\alpha = 0$ predictions below the actual value have no cost, so the best thing to do is to predict $-\infty$, since the quantile function returns this for $p = 0$. When $\alpha = 1$ predictions above the actual value have no cost, so the best thing to do is to predict $\infty$. If $\alpha = 0.5$ the best result is given by the result of the quantile function for 0.5, i.e., the median, which for a normal distribution is also the mean.

As in the previous section, we compare the method without reframing (*None*), with probabilistic (local) reframing and two methods with global reframing.

- *None*: No reframing. The prediction (conditional estimated mean) is used as it is.

- *Own, uKNC, BIN*: Probabilistic (local) reframing using $r^*(x, \ell_\alpha^A, \hat{f})$, as given by eq. 6. The conditional normal density is obtained by three different methods: as derived from the base technique (*Own*) and enrichment methods (*uKNC* and *BIN*).

- *CoSh*: Global reframing using a constant shift $s_0$ for all the predictions $R^+(x, \ell_\alpha^A, \hat{f}) = \mathbb{E}_{\hat{f}}(y|x) + s_0$. In order to calculate a good shift, we look for the best shift for the whole training set. Interestingly, in

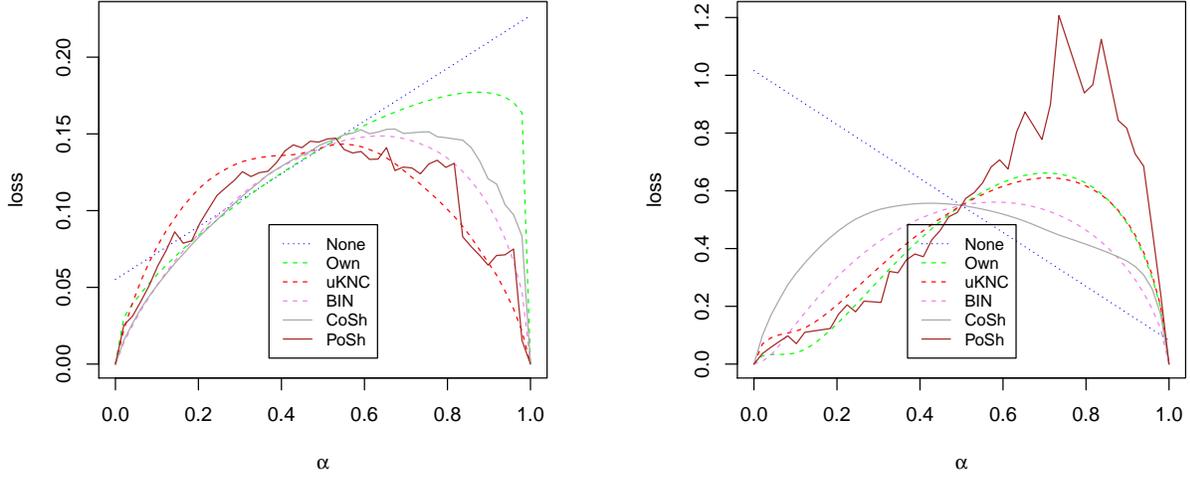

Figure 7: Left: comparing the asymmetric absolute loss using different methods for dataset *iris3* with base technique *LR*. Right: comparing the asymmetric square loss using different methods for dataset *road* with base technique *Tree*.

this case, the calculation of the optimal constant shift for the training data follows a convex function (from a convex loss function) and can be calculated using efficient numerical methods. For instance, [1] use hill climbing to calculate this optimal $s_0$.

- *PoSh*: Global reframing using a polynomial shift $s(x)$ for all the predictions $R^p(x, \ell_\alpha^A, \hat{f}) = s(\mathbb{E}_{\hat{f}}(y|x))$ where $s$ is a polynomial function. Considering that the problem is convex, [72] present a numerical method (also based on hill climbing) to derive this polynomial in a relatively efficient way. We will just show the results for a first-order polynomial because this degree produced the best results.

Using these methods, the evolution of this loss for different values of $\alpha$ for one dataset is shown in Figure 7 (left). The overall results are shown in Table 7 for several base techniques. Here we see that the global reframing methods perform relatively well too, especially for the base technique using trees.

Now we will derive the minimisation expression for the asymmetric squared error (definition 9):

**Proposition 6.** *If $\ell$ is the asymmetric squared error function $\ell_\alpha^S(\hat{y}, y)$ given by definition 9 and $\hat{f}$ is any distribution with mean $\hat{\mu}(x)$ and standard deviation $\hat{\sigma}(x)$, then the expected loss for a predicted value t is given by:*

$$\mathscr{L}(x,t,\hat{f},\ell_\alpha^S) = (1-2\alpha)\left[t^2\hat{F}(t|x) - 2t\int_{-\infty}^t y\hat{f}(y|x)dy + \int_{-\infty}^t y^2\hat{f}(y|x)dy\right] + \alpha\left[t^2 - 2t\hat{\mu}(x) + \hat{\mu}_2(x)\right] \quad (7)$$

*where $\hat{\mu}_2(x)$ is the second raw moment of $\hat{f}(y|x)$.*

**Proposition 7.** *If $\ell$ is the asymmetric squared error function $\ell_\alpha^S(\hat{y}, y)$ given by definition 9 and $\hat{f}$ is any distribution with mean $\hat{\mu}(x)$ and standard deviation $\hat{\sigma}(x)$, then the optimal prediction is given by the value t such that the following equation holds:*

$$(1-2\alpha)\left[2t\hat{F}(t|x) - 2\int_{-\infty}^t y\hat{f}(y|x)dy\right] + 2\alpha t - 2\alpha\hat{\mu}(x) = 0 \quad (8)$$

|    | LR None | LR Own | LR uKNCBIN | LR CoSh | LR PoSh | LR | kNN None | kNN Own | kNN uKNCBIN | kNN CoSh | kNN PoSh | kNN | Tree None | Tree Own | Tree uKNCBIN | Tree CoSh | Tree PoSh | Tree |
|----|------|------|------|------|------|------|------|------|------|------|------|------|------|------|------|------|------|------|
| 1  | 1.40 | 0.97 | 0.95 | 0.94 | 0.90 | 0.89 | 4.23 | 3.06 | 3.08 | 3.08 | 3.02 | 2.84 | 3.32 | 2.48 | 2.42 | 2.49 | 2.45 | 2.35 |
| 2  | 4.47 | 3.02 | 2.80 | 2.87 | 2.90 | 2.92 | 3.31 | 2.21 | 2.27 | 2.27 | 2.18 | 2.53 | 3.67 | 2.48 | 2.49 | 2.54 | 2.61 | 2.52 |
| 3  | 2.55 | 1.73 | 1.69 | 1.66 | 1.70 | 1.81 | 2.86 | 1.82 | 1.84 | 1.83 | 1.82 | 1.91 | 3.25 | 2.09 | 2.07 | 2.12 | 2.12 | 2.09 |
| 4  | 4.18 | 3.28 | 3.22 | 3.20 | 3.26 | 3.34 | 7.87 | 5.82 | 5.62 | 5.63 | 5.81 | 5.72 | 8.38 | 6.37 | 6.40 | 6.29 | 6.19 | 6.04 |
| 5  | 2.62 | 1.92 | 1.65 | 1.90 | 1.88 | 1.72 | 5.53 | 3.94 | 3.75 | 3.88 | 3.64 | 3.52 | 4.44 | 3.27 | 3.15 | 3.30 | 3.12 | 3.01 |
| 6  | 2.18 | 1.57 | 1.69 | 1.58 | 1.46 | 1.64 | 3.13 | 2.30 | 2.18 | 2.28 | 2.32 | 2.60 | 3.78 | 2.46 | 2.39 | 2.44 | 2.44 | 2.35 |
| 7  | 6.01 | 4.59 | 4.26 | 4.27 | 4.22 | 3.91 | 5.20 | 3.60 | 3.60 | 3.59 | 3.52 | 3.38 | 6.06 | 4.32 | 4.37 | 4.35 | 4.39 | 4.10 |
| 8  | 1.72 | 1.31 | 1.18 | 1.33 | 1.34 | 1.33 | 2.89 | 2.19 | 2.19 | 2.19 | 2.23 | 2.22 | 3.30 | 2.30 | 2.27 | 2.33 | 2.25 | 2.00 |
| 9  | 7.38 | 5.38 | 4.34 | 5.78 | 5.76 | 6.02 | 2.85 | 1.92 | 1.92 | 1.96 | 1.99 | 2.13 | 3.96 | 2.49 | 2.56 | 2.55 | 2.52 | 2.07 |
| 10 | 4.89 | 3.45 | 2.72 | 3.59 | 3.55 | 3.47 | 3.54 | 2.36 | 2.38 | 2.39 | 2.33 | 2.87 | 5.55 | 3.25 | 3.20 | 3.45 | 3.52 | 3.25 |
| 11 | 6.58 | 5.15 | 4.90 | 5.08 | 4.98 | 4.83 | 8.20 | 6.44 | 6.25 | 6.30 | 6.28 | 6.10 | 8.20 | 6.46 | 6.46 | 6.47 | 6.46 | 6.43 |
| 12 | 1.27 | 0.95 | 0.80 | 0.81 | 0.80 | 0.87 | 1.28 | 0.86 | 0.84 | 0.84 | 0.83 | 0.88 | 2.00 | 1.31 | 1.31 | 1.30 | 1.30 | 1.27 |
| 13 | 3.84 | 2.80 | 2.47 | 2.47 | 2.43 | 2.44 | 4.00 | 2.60 | 2.58 | 2.58 | 2.53 | 2.49 | 4.20 | 2.71 | 2.70 | 2.70 | 2.68 | 2.61 |
| 14 | 6.00 | 4.37 | 3.93 | 3.98 | 3.99 | 3.88 | 4.85 | 3.09 | 3.11 | 3.11 | 3.12 | 3.09 | 6.49 | 4.34 | 4.38 | 4.38 | 4.50 | 4.22 |
| 15 | 2.28 | 1.52 | 1.53 | 1.48 | 1.51 | 1.42 | 2.16 | 1.44 | 1.53 | 1.48 | 1.41 | 1.56 | 2.77 | 1.84 | 1.76 | 1.82 | 1.86 | 1.71 |
| 16 | 2.47 | 1.86 | 1.56 | 1.57 | 1.56 | 1.65 | 2.27 | 1.46 | 1.48 | 1.46 | 1.47 | 1.56 | 2.33 | 1.49 | 1.51 | 1.51 | 1.52 | 1.57 |
| 17 | 20.18| 15.72| 11.50| 15.93| 15.91| 14.76| 6.03 | 3.77 | 3.78 | 3.75 | 4.23 | 3.35 | 8.48 | 5.13 | 5.24 | 5.54 | 6.66 | 5.21 |
| 18 | 7.27 | 5.56 | 5.24 | 5.20 | 5.28 | 4.55 | 5.45 | 3.94 | 3.84 | 3.78 | 4.08 | 3.88 | 6.15 | 4.74 | 4.69 | 4.74 | 4.72 | 4.51 |
| 19 | 2.74 | 1.74 | 1.72 | 1.72 | 1.68 | 1.80 | 4.21 | 2.71 | 2.70 | 2.74 | 2.69 | 2.65 | 3.73 | 2.45 | 2.51 | 2.53 | 2.53 | 2.42 |
| 20 | 2.97 | 1.91 | 1.91 | 1.95 | 1.98 | 2.01 | 2.89 | 1.90 | 1.93 | 1.91 | 1.96 | 2.13 | 2.94 | 1.92 | 1.88 | 1.90 | 1.93 | 1.99 |
| AR | 6.00 | 3.90 | **2.35** | 3.05 | 2.75 | 2.95 | 6.00 | 2.90 | 3.00 | 3.05 | 2.90 | 3.15 | 6.00 | 3.10 | 2.75 | 3.80 | 3.65 | **1.70** |

Table 7: Results for the absolute loss $\ell_\alpha^A$ for the datasets in Table 12, using the experimental methodology in section 2.5. Each row aggregates the folds and ten different values for $\alpha$ per fold with $\alpha \in \{0, 0.111, 0.222, \ldots, 1\}$. For visibility all the losses are multiplied by 10. Each section of six columns shows results for different base techniques (*LR*, *kNN* and *Tree*). The average ranks (AR) are calculated for these three groups separately. The Friedman statistic for the three sections are (50.29, 43.11 and 59 respectively), which are greater than the Critical Value (12.57). This means that the null hypothesis is rejected (significance level: 0.05) and the methods do not perform equally. Differences in average ranks higher than the critical difference for the Nemenyi post-hoc test (0.5217) imply that the difference is significant (in bold).

The previous result can be simplified for a normal distribution

**Proposition 8.** *If $\ell$ is the asymmetric squared error function $\ell_\alpha^S(\hat{y}, y)$ given by definition 9 and $\hat{f}$ is a normal distribution with mean $\hat{\mu}(x)$ and standard deviation $\hat{\sigma}(x)$, then the optimal prediction $t$ is given by first calculating $t'$ from the following equation:*

$$t'\Phi(t') + \phi(t') + t'\frac{\alpha}{1-2\alpha} = 0 \tag{9}$$

*and then getting $t = \hat{\sigma}(x)t' + \hat{\mu}(x)$. Note the use of the standardised cumulative normal distribution $\Phi$ and the standardised normal density function $\phi$.*

Even though the value of $t'$ cannot be expressed in a closed form, we only need to calculate this value for each $\alpha$ once, since it is calculated for the standard normal distribution. Then, we just use the expression $t = \hat{\sigma}(x)t' + \hat{\mu}(x)$ for each example.

Figure 7 (right) shows the evolution of this loss for different values of $\alpha$ for one dataset, using several reframing methods. The overall results are shown in Table 8 for several base techniques and the same enrichment methods as in Table 7. The results are better for the probabilistic (local) reframing methods. This indicates that the (more common) asymmetric squared loss, which highly penalises wrong big shifts, requires a more detailed (local) reframing.

Apart from $\ell_\alpha^A$ and $\ell_\alpha^S$, there are many other kinds of asymmetric loss. In fact, for instance, we could use a discrete function where loss would be 0 if the error is inside a tolerance band (which can be asymmetric) and, e.g., 1 otherwise. We will discuss this notion of 'tolerance' after the following section.

| | LR None | LR Own | LR uKNC | LR BIN | LR CoSh | LR PoSh | kNN None | kNN Own | kNN uKNC | kNN BIN | kNN CoSh | kNN PoSh | Tree None | Tree Own | Tree uKNC | Tree BIN | Tree CoSh | Tree PoSh |
|---|---|---|---|---|---|---|---|---|---|---|---|---|---|---|---|---|---|---|
| 1 | 0.61 | 0.43 | 0.46 | 0.41 | 0.41 | 0.39 | 5.22 | 3.77 | 3.79 | 3.80 | 3.67 | 3.41 | 3.11 | 2.34 | 2.26 | 2.34 | 2.23 | 2.13 |
| 2 | 7.25 | 5.17 | 4.96 | 5.02 | 5.02 | 5.15 | 4.92 | 3.56 | 3.60 | 3.58 | 3.47 | 4.17 | 6.87 | 4.92 | 4.93 | 4.99 | 4.97 | 4.95 |
| 3 | 2.67 | 1.90 | 1.81 | 1.80 | 1.82 | 1.92 | 2.87 | 1.88 | 1.89 | 1.89 | 1.91 | 2.08 | 3.62 | 2.43 | 2.38 | 2.45 | 2.44 | 2.41 |
| 4 | 6.20 | 4.86 | 4.74 | 4.84 | 4.77 | 4.85 | 14.19 | 10.51 | 10.25 | 10.27 | 10.25 | 9.90 | 15.82 | 11.99 | 12.04 | 11.85 | 11.54 | 11.17 |
| 5 | 2.23 | 1.65 | 1.57 | 1.66 | 1.67 | 1.54 | 8.35 | 6.19 | 6.10 | 6.14 | 6.23 | 6.15 | 6.64 | 5.07 | 4.97 | 5.09 | 5.02 | 5.00 |
| 6 | 2.76 | 2.11 | 2.16 | 2.07 | 2.05 | 2.22 | 4.37 | 3.16 | 3.10 | 3.17 | 3.07 | 3.29 | 5.39 | 3.81 | 3.73 | 3.75 | 3.76 | 3.92 |
| 7 | 8.40 | 6.40 | 5.98 | 6.00 | 5.90 | 5.65 | 6.94 | 4.86 | 4.86 | 4.84 | 4.73 | 4.78 | 8.65 | 6.20 | 6.25 | 6.23 | 6.16 | 5.93 |
| 8 | 3.61 | 2.81 | 2.47 | 2.87 | 2.87 | 2.83 | 6.08 | 4.77 | 4.78 | 4.78 | 5.84 | 6.64 | 6.98 | 5.29 | 5.23 | 5.27 | 5.54 | 6.03 |
| 9 | 185.34 | 129.41 | 110.49 | 147.86 | 147.87 | 147.96 | 4.33 | 3.13 | 3.15 | 3.22 | 3.15 | 4.00 | 5.93 | 4.03 | 4.13 | 4.06 | 4.21 | 4.41 |
| 10 | 8.50 | 6.11 | 5.23 | 6.37 | 6.37 | 6.20 | 4.78 | 3.34 | 3.38 | 3.39 | 3.63 | 7.64 | 7.45 | 4.73 | 4.70 | 4.87 | 4.97 | 4.82 |
| 11 | 9.96 | 7.82 | 7.51 | 7.71 | 7.66 | 7.57 | 14.04 | 11.01 | 10.70 | 10.78 | 11.04 | 10.97 | 14.06 | 11.06 | 11.06 | 11.06 | 11.07 | 10.99 |
| 12 | 0.52 | 0.39 | 0.35 | 0.35 | 0.34 | 0.37 | 0.60 | 0.42 | 0.41 | 0.41 | 0.41 | 0.45 | 1.22 | 0.83 | 0.83 | 0.82 | 0.82 | 0.82 |
| 13 | 5.01 | 3.72 | 3.35 | 3.35 | 3.36 | 3.44 | 5.45 | 3.66 | 3.66 | 3.66 | 3.68 | 3.67 | 5.75 | 3.87 | 3.85 | 3.86 | 3.86 | 3.85 |
| 14 | 11.20 | 8.22 | 7.53 | 7.64 | 7.69 | 7.32 | 7.16 | 4.85 | 4.86 | 4.86 | 4.94 | 5.03 | 12.35 | 8.42 | 8.48 | 8.47 | 8.66 | 8.13 |
| 15 | 2.72 | 1.96 | 1.95 | 1.93 | 1.93 | 2.00 | 2.51 | 1.80 | 1.85 | 1.83 | 1.93 | 2.01 | 2.93 | 2.06 | 1.99 | 2.06 | 2.07 | 2.10 |
| 16 | 2.60 | 2.00 | 1.80 | 1.81 | 1.79 | 1.86 | 2.35 | 1.62 | 1.65 | 1.64 | 1.63 | 1.70 | 2.46 | 1.67 | 1.69 | 1.69 | 1.71 | 1.71 |
| 17 | 163.98 | 127.58 | 99.11 | 129.83 | 129.50 | 127.15 | 10.84 | 7.43 | 7.40 | 7.49 | 8.71 | 14.69 | 20.23 | 12.95 | 13.06 | 13.40 | 14.44 | 13.03 |
| 18 | 13.19 | 10.02 | 9.51 | 9.52 | 9.68 | 9.05 | 7.23 | 5.28 | 5.17 | 5.11 | 5.42 | 4.90 | 9.41 | 7.26 | 7.17 | 7.26 | 7.12 | 6.82 |
| 19 | 2.13 | 1.41 | 1.44 | 1.39 | 1.37 | 1.43 | 5.33 | 3.51 | 3.51 | 3.54 | 3.50 | 3.42 | 4.75 | 3.19 | 3.30 | 3.28 | 3.36 | 3.28 |
| 20 | 2.62 | 1.77 | 1.77 | 1.79 | 1.79 | 1.82 | 2.81 | 1.90 | 1.92 | 1.89 | 1.93 | 2.11 | 2.73 | 1.83 | 1.78 | 1.82 | 1.84 | 1.84 |
| AR | 6.00 | 3.85 | **2.10** | 2.95 | 2.90 | 3.20 | 5.85 | **2.45** | 2.70 | 2.80 | 3.25 | 3.95 | 6.00 | **2.85** | 2.45 | 3.45 | 3.70 | **2.55** |

Table 8: Results for the squared loss $\ell_\alpha^S$ for the datasets in Table 12, using the experimental methodology in section 2.5. Each row aggregates the folds and ten different values for $\alpha$ per fold with $\alpha \in \{0, 0.111, 0.222, \ldots, 1\}$. For visibility all the losses are multiplied by 10. Each section of six columns shows results for different base techniques (*LR*, *kNN* and *Tree*). The average ranks (AR) are calculated for these three groups separately. The Friedman statistic for the three sections are (51.91, 45.83 and 49.83 respectively), which are greater than the Critical Value (12.57). This means that the null hypothesis is rejected (significance level: 0.05) and the methods do not perform equally. Differences in average ranks higher than the critical difference for the Nemenyi post-hoc test (0.5217) imply that the difference is significant (in bold).

# 6 Rejection rule applications

A common situation when working with predictive models appears when there is the possibility of abstention, i.e., to reject the prediction and do nothing (or delegate to an expert or other kind of model). The rationale is to avoid a decision that is likely to have more cost than the abstention itself. In order to do this, we need to know what the cost of an abstention is, which may be constant or may depend on the instance. According to this information, a decision rule which tries to minimise the cost (as an aggregation of the overall prediction loss and the rejection cost) is known as a *rejection rule*. Several works in the literature have been devoted to rejection rules, although most of the work in the area of machine learning is conceived for classification [28, 52, 25].

We can apply a rejection rule on top of any loss function, such as those seen in the previous sections. For instance, for the asymmetric absolute error $\ell_\alpha^A$ we can derive the corresponding loss with rejection option $\ell_{\alpha,\rho}^{AR}$ as follows:

**Definition 10.** *The asymmetric absolute error $\ell_{\alpha,\rho}^{AR}$ with rejection option is a loss function defined as follows:*

$$\ell_{\alpha,\rho}^{AR}(\hat{y}, y) = \quad \rho \quad \text{if REJECT}$$
$$= \ell_\alpha^A(\hat{y}, y) \quad \text{otherwise}$$

A straightforward way of handling this type of loss functions with rejection option is to estimate the expected loss and check whether it is greater than $\rho$. If this is the case we should reject. Otherwise, we should use the decision rule as if there were no rejection. For instance, for $\ell_{\alpha,\rho}^{AR}$, we would just calculate the expected loss using proposition 4, and compare this to $\rho$. Only if it is lower than $\rho$ would we apply the minimisation given by proposition 5.

However, we need to derive a more operative expression for the expected loss from proposition 4:

**Proposition 9.** *Consider the asymmetric absolute error function $\ell_\alpha^A(\hat{y}, y)$ given by definition 8 and a normal conditional distribution $\hat{f}$ with mean $\hat{\mu}(x)$ and standard deviation $\hat{\sigma}(x)$. The expected loss for a prediction value t can be further simplified to:*

$$\mathscr{L}(x, t, \hat{f}, \ell_\alpha^A) = [t'\Phi(t') + \phi(t') - \alpha t']\hat{\sigma}(x) \tag{10}$$

*with $t' = \frac{t - \hat{\mu}(x)}{\hat{\sigma}(x)}$. Note the use of the standardised cumulative normal distribution $\Phi$ and the standardised normal density function $\phi$.*

The expected loss given by proposition 9 is then easy to calculate and can be used to compare it with the cost of rejection. This leads to the decision rule for REJECT:

$$\mathscr{L}(x, t, \hat{f}, \ell_\alpha^A) > \rho \tag{11}$$

Now we are ready to compare the methods based on probabilistic (local) reframing using the above rule with methods which do a global reframing, as we did in the previous section. The methods are:

- *None*: No reframing. The prediction (conditional estimated mean) is used as it is.

- *Own, uKNC, BIN*: We use the rejection rule given by eq. (11) and proposition 9. In case of no reject we apply proposition 5, eq. (6), as used in the previous section.

- *CoSh*: We decide whether we reject or not using the optimal reject *rate* calculated on the training dataset. This rate is calculated assuming that a percentage (rate) of examples is just rejected (the examples are necessarily chosen randomly, since there is no information about reliability). Given the optimal rate and the optimal shift, we reject examples using the rate and, if the example is finally not rejected, we use the CoSh method as in the previous section (using the method from [1]).

- *PoSh*: Similar to CoSh but the polynomial approach in [72] is used instead.

For the experiments, we vary both $\alpha$ and $\rho$. For $\rho$ we apply the following function:

$$\rho = \frac{1}{2}\sigma(D_Y)\frac{r}{1-r}$$

where $\sigma(D_Y)$ is the standard deviation of the output variable for the dataset $D$. We let $r$ range between 0 and 1. The rationale for the previous function can be explained with the extreme cases. If $r = 0$ we get $\rho = 0$ and reject has no cost (so we will always reject). If $r = 0.5$ we get $\rho = \frac{1}{2}\sigma(D_Y)$, which means that with a trivial constant model, residuals will equal the standard deviation (and the expected error), so the expected loss will be $\frac{1}{2}\sigma(D_Y) = (\alpha\sigma(D_Y) + (1-\alpha)\sigma(D_Y))/2$. This means that approximately we will reject half of the times (for a trivial model). And finally, for $r = 1$ we get $\rho = \infty$ and reject has infinite cost (so we will never reject).

Figure 8 (left) shows the evolution of this loss for different values of $\rho$, as derived from $\rho$ (with $\alpha$ fixed to 0.5) for one dataset. The overall results are shown in Table 9. We see that probabilistic reframing takes advantage of a local decision. In the end, the conditional variance is used to make a *ranking*, which is crucial for rejection rules.

For the squared loss with reject ($\ell_{\alpha,\rho}^{SR}$), we work similarly:

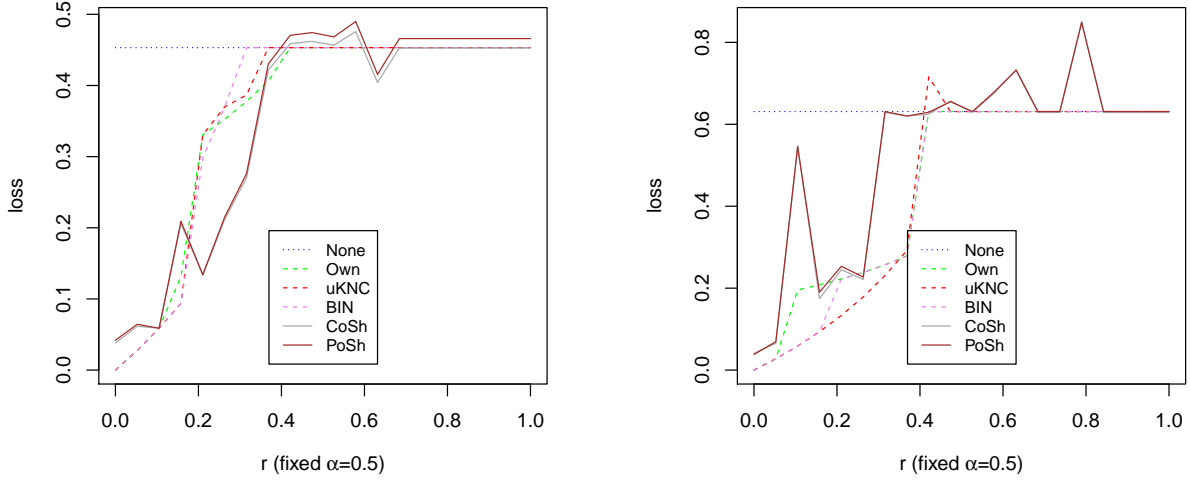

Figure 8: Left: comparing the absolute loss with reject using different methods for dataset *savings* with base technique *Tree*. Right: comparing the squared loss with reject using different methods for dataset *salinity* with base technique *kNN*.

**Proposition 10.** *Consider the asymmetric squared error function $\ell_\alpha^S(\hat{y}, y)$ given by definition 9 and a normal conditional distribution $\hat{f}$ with mean $\hat{\mu}(x)$ and standard deviation $\hat{\sigma}(x)$. The expected loss can be expressed as:*

$$\mathcal{L}(x, t, \hat{f}, \ell_\alpha^S) = \Phi(t')(1 - 2\alpha) \left[ (t'\sigma)^2 + 3t'\sigma^2 q(t') - 2\mu^2 + 4\mu\sigma q(t') - \sigma^2 \right] + \alpha\sigma^2(t' + 1) \quad (12)$$

with $t' = \frac{t - \hat{\mu}(x)}{\hat{\sigma}(x)}$, $q(t') = \frac{\phi(t')}{\Phi(t')}$ and notation $\mu$ for $\hat{\mu}(x)$ and $\sigma$ for $\hat{\sigma}(x)$.

Although the previous expression is long, it can be computed easily with the standard normal distribution. Figure 8 (right) shows the evolution of this loss for a fixed $\alpha = 0.5$ and different values of $r$ (from which $\rho$ is derived) for one dataset. The overall results with the same configuration as the absolute loss with reject are shown in Table 10. The results are even more clear-cut in this case.

# 7 Discussion

After this jaunt through several families of loss functions using different kinds of reframing methods (local or global) we are ready to make a comprehensive analysis of the results, see other (many) applications and close the paper with the overall contributions and some future work.

## 7.1 Overview of results and contributions

As a short recapitulation of results, we can just summarise tables 5, 6, 7, 8, 9 and 10 by counting the number of cases where each reframing is in the group of the (statistically significant) best results. This is 17 (out of 18) for probabilistic local reframing in front of 8 (out of 18) for global reframing. If we focus on particular methods, the probabilistic local reframing based on the enrichment method *uKNC* is in the group of best results 15 times (out of 18) in front of only 6 (out of 18) for the best global reframing (*PoSh*, and *CoSh* for the first two tables). In fact, if we compare *uKNC* against *PoSh* (*CoSh* for the first two tables) using the Nemenyi post-hoc test difference in each case, we have 11 *wins*, 6 *ties* and 1 *lose*. This supports the claim that an appropriate probabilistic reframing with the use of a lightweight normal conditional distribution using appropriate estimations (directly or through enrichment methods) is a good approach for a wide variety of cost-sensitive problems.

|    | LR None | LR Own | LR uKNC | LR BIN | LR CoSh | LR PoSh | kNN None | kNN Own | kNN uKNC | kNN BIN | kNN CoSh | kNN PoSh | Tree None | Tree Own | Tree uKNC | Tree BIN | Tree CoSh | Tree PoSh |
|---|---|---|---|---|---|---|---|---|---|---|---|---|---|---|---|---|---|---|
| 1  | 1.40  | 0.68 | 0.68 | 0.66  | 1.10 | 1.08 | 4.23 | 1.96 | 1.97 | 1.96 | 1.84 | 1.81 | 3.32 | 1.70 | 1.61 | 1.72 | 1.57 | 1.55 |
| 2  | 4.47  | 1.89 | 1.68 | 1.72  | 1.89 | 1.82 | 3.31 | 1.38 | 1.43 | 1.43 | 1.50 | 1.62 | 3.67 | 1.56 | 1.55 | 1.58 | 1.71 | 1.71 |
| 3  | 2.55  | 1.14 | 1.08 | 1.07  | 1.33 | 1.36 | 2.86 | 1.15 | 1.17 | 1.16 | 1.31 | 1.31 | 3.25 | 1.32 | 1.29 | 1.33 | 1.49 | 1.47 |
| 4  | 4.18  | 2.23 | 2.10 | 2.19  | 2.19 | 2.13 | 7.87 | 3.76 | 3.53 | 3.55 | 3.80 | 3.95 | 8.38 | 4.16 | 4.17 | 3.94 | 3.83 | 3.90 |
| 5  | 2.62  | 1.32 | 1.09 | 1.32  | 1.49 | 1.41 | 5.53 | 2.63 | 2.48 | 2.58 | 2.49 | 2.40 | 4.44 | 2.23 | 2.10 | 2.23 | 2.21 | 2.15 |
| 6  | 2.18  | 1.06 | 1.12 | 1.06  | 1.13 | 1.17 | 3.13 | 1.45 | 1.36 | 1.45 | 1.51 | 1.75 | 3.78 | 1.57 | 1.48 | 1.54 | 1.75 | 1.61 |
| 7  | 6.01  | 3.09 | 2.66 | 2.67  | 2.45 | 2.40 | 5.20 | 2.24 | 2.24 | 2.22 | 2.20 | 2.18 | 6.06 | 2.70 | 2.78 | 2.73 | 2.46 | 2.40 |
| 8  | 1.72  | 0.89 | 0.70 | 0.91  | 0.92 | 0.90 | 2.89 | 1.51 | 1.51 | 1.51 | 1.66 | 1.66 | 3.30 | 1.52 | 1.51 | 1.54 | 1.59 | 1.46 |
| 9  | 7.38  | 1.54 | 1.24 | 3.94  | 3.87 | 3.81 | 2.85 | 1.21 | 1.22 | 1.27 | 1.45 | 1.55 | 3.96 | 1.50 | 1.54 | 1.57 | 1.87 | 1.56 |
| 10 | 4.89  | 2.14 | 1.49 | 2.29  | 2.29 | 2.31 | 3.54 | 1.47 | 1.50 | 1.49 | 1.62 | 2.01 | 5.55 | 1.92 | 1.87 | 2.05 | 2.33 | 2.16 |
| 11 | 6.58  | 3.51 | 3.22 | 3.41  | 2.92 | 2.90 | 8.20 | 4.37 | 4.08 | 4.16 | 3.93 | 3.88 | 8.20 | 4.39 | 4.39 | 4.39 | 4.04 | 4.03 |
| 12 | 1.27  | 0.66 | 0.57 | 0.57  | 0.99 | 1.01 | 1.28 | 0.61 | 0.59 | 0.59 | 0.94 | 0.95 | 2.00 | 0.89 | 0.89 | 0.89 | 1.14 | 1.14 |
| 13 | 3.84  | 1.85 | 1.52 | 1.51  | 1.50 | 1.48 | 4.00 | 1.59 | 1.57 | 1.57 | 1.54 | 1.53 | 4.20 | 1.66 | 1.64 | 1.64 | 1.61 | 1.59 |
| 14 | 6.00  | 2.71 | 2.24 | 2.29  | 2.29 | 2.24 | 4.85 | 1.86 | 1.88 | 1.88 | 1.94 | 1.92 | 6.49 | 2.51 | 2.53 | 2.53 | 2.44 | 2.38 |
| 15 | 2.28  | 1.01 | 1.01 | 0.98  | 1.36 | 1.31 | 2.16 | 0.94 | 1.01 | 0.97 | 1.19 | 1.22 | 2.77 | 1.21 | 1.14 | 1.19 | 1.40 | 1.33 |
| 16 | 2.47  | 1.28 | 1.02 | 1.03  | 1.11 | 1.13 | 2.27 | 0.95 | 0.97 | 0.96 | 1.06 | 1.10 | 2.33 | 0.97 | 0.98 | 0.99 | 1.10 | 1.11 |
| 17 | 20.18 | 8.83 | 3.96 | 10.43 | 5.18 | 5.16 | 6.03 | 2.29 | 2.29 | 2.27 | 2.36 | 2.00 | 8.48 | 2.83 | 2.87 | 3.00 | 3.05 | 2.87 |
| 18 | 7.27  | 3.60 | 3.17 | 3.18  | 2.63 | 2.43 | 5.45 | 2.52 | 2.40 | 2.36 | 2.21 | 2.17 | 6.15 | 3.25 | 3.17 | 3.20 | 2.39 | 2.36 |
| 19 | 2.74  | 1.15 | 1.14 | 1.13  | 1.28 | 1.30 | 4.21 | 1.64 | 1.64 | 1.67 | 1.71 | 1.66 | 3.73 | 1.53 | 1.56 | 1.59 | 1.61 | 1.58 |
| 20 | 2.97  | 1.24 | 1.25 | 1.27  | 1.45 | 1.47 | 2.89 | 1.23 | 1.25 | 1.23 | 1.45 | 1.51 | 2.94 | 1.24 | 1.21 | 1.23 | 1.41 | 1.41 |
| AR | 6.00  | 3.50 | **1.80** | 2.75 | 3.70 | 3.25 | 6.00 | **2.50** | 2.85 | 2.60 | 3.70 | 3.35 | 6.00 | 2.85 | **2.30** | 3.45 | 3.70 | **2.70** |

Table 9: Results for the absolute reject loss $\ell^{AR}_{\alpha,\rho}$ for the datasets in Table 12, using the experimental methodology in section 2.5. Each row aggregates the folds and five different values for $\alpha$ with $\alpha \in \{0, 0.25, 0.5, 0.75, 1\}$ and ten different values for $\rho$ with $r \in \{0, 0.111, 0.222, \ldots, 1\}$ and $\rho = \frac{1}{2}\sigma(D_Y)\frac{r}{1-r}$ (totalling 50 variations for fold). For visibility all the losses are multiplied by 10. Each section of six columns shows results for different base techniques (*LR*, *kNN* and *Tree*). The average ranks (AR) are calculated for these three groups separately. The Friedman statistic for the three sections are (56.03, 48.83 and 50.26 respectively), which are greater than the Critical Value (12.57). This means that the null hypothesis is rejected (significance level: 0.05) and the methods do not perform equally. Differences in average ranks higher than the critical difference for the Nemenyi post-hoc test (0.5217) imply that the difference is significant (in bold).

This claim is accompanied by a series of major contributions:

- We push forward an *appropriate mapping between classification and regression*, and develop the right parallelism between crisp and soft models in classification and regression.

- We vindicate *reframing* as a flexible and powerful approach for context-sensitive applications, and characterise the distinction between *global* reframing (typically based on crisp regression models) and *local* reframing (typically based on soft regression models).

- We uphold a lightweight view of soft regression models as *normal conditional density estimators*, only requiring two parameters. This entails several benefits: easier, more robust estimation and simpler optimisation formulae resulting from expected loss.

- We introduce *new metrics* for the evaluation of soft regression models.

- We present *enrichment* as a way to convert any traditional one-parameter crisp regression model into a two-parameter soft regression model, by just working with the actual and predicted values. This has several advantages: enrichment methods are easily applicable to any regression method, they only require the actual output values of a training (or validation) dataset, and the two-parameter normal density estimator does not need to be recalculated whenever the loss function changes.

|    | LR None | LR Own | LR uKNC | LR BIN | LR CoSh | LR PoSh | kNN None | kNN Own | kNN uKNC | kNN BIN | kNN CoSh | kNN PoSh | Tree None | Tree Own | Tree uKNC | Tree BIN | Tree CoSh | Tree PoSh |
|---|---|---|---|---|---|---|---|---|---|---|---|---|---|---|---|---|---|---|
| 1 | 0.61 | 0.31 | 0.40 | 0.30 | 1.33 | 1.33 | 5.22 | 2.55 | 2.64 | 2.65 | 2.27 | 2.09 | 3.11 | 1.62 | 1.59 | 1.62 | 1.62 | 1.59 |
| 2 | 7.25 | 3.26 | 2.90 | 3.01 | 3.02 | 3.16 | 4.92 | 2.23 | 2.41 | 2.32 | 2.48 | 2.69 | 6.87 | 3.09 | 3.11 | 3.22 | 3.06 | 3.01 |
| 3 | 2.67 | 1.32 | 1.14 | 1.12 | 1.45 | 1.49 | 2.87 | 1.14 | 1.18 | 1.16 | 1.54 | 1.58 | 3.62 | 1.51 | 1.41 | 1.57 | 1.79 | 1.79 |
| 4 | 6.20 | 3.54 | 3.45 | 3.36 | 3.85 | 3.81 | 14.19 | 7.03 | 6.47 | 6.52 | 6.74 | 6.61 | 15.82 | 8.26 | 8.27 | 7.82 | 7.37 | 7.39 |
| 5 | 2.23 | 1.15 | 1.12 | 1.15 | 1.74 | 1.66 | 8.35 | 4.19 | 4.06 | 4.16 | 4.33 | 4.25 | 6.64 | 3.49 | 3.42 | 3.50 | 3.69 | 3.70 |
| 6 | 2.76 | 1.56 | 1.62 | 1.47 | 1.55 | 1.64 | 4.37 | 1.96 | 1.88 | 2.07 | 2.05 | 2.19 | 5.39 | 2.52 | 2.38 | 2.44 | 2.75 | 2.87 |
| 7 | 8.40 | 4.41 | 4.08 | 4.11 | 3.74 | 3.65 | 6.94 | 3.24 | 3.21 | 3.19 | 3.28 | 3.28 | 8.65 | 4.26 | 4.33 | 4.31 | 3.76 | 3.74 |
| 8 | 3.61 | 1.92 | 0.93 | 1.94 | 2.12 | 2.10 | 6.08 | 3.25 | 3.25 | 3.25 | 4.01 | 4.43 | 6.98 | 3.49 | 3.40 | 3.42 | 3.97 | 4.07 |
| 9 | 185.34 | 1.03 | 10.11 | 99.93 | 86.15 | 86.25 | 4.33 | 2.03 | 2.09 | 2.17 | 2.21 | 2.43 | 5.93 | 2.43 | 2.61 | 2.42 | 2.73 | 2.89 |
| 10 | 8.50 | 3.75 | 2.02 | 4.42 | 3.71 | 3.63 | 4.78 | 2.10 | 2.20 | 2.17 | 2.72 | 3.89 | 7.45 | 2.60 | 2.51 | 2.90 | 3.03 | 2.92 |
| 11 | 9.96 | 5.32 | 5.21 | 5.28 | 4.94 | 4.91 | 14.04 | 7.48 | 7.28 | 7.32 | 7.22 | 7.23 | 14.06 | 7.52 | 7.52 | 7.52 | 7.18 | 7.12 |
| 12 | 0.52 | 0.27 | 0.26 | 0.25 | 1.33 | 1.33 | 0.60 | 0.31 | 0.30 | 0.30 | 1.20 | 1.21 | 1.22 | 0.60 | 0.60 | 0.60 | 1.52 | 1.52 |
| 13 | 5.01 | 2.58 | 1.97 | 1.93 | 1.96 | 1.97 | 5.45 | 2.10 | 2.14 | 2.13 | 2.07 | 2.07 | 5.75 | 2.29 | 2.22 | 2.24 | 2.14 | 2.12 |
| 14 | 11.20 | 5.31 | 3.83 | 4.14 | 3.92 | 3.86 | 7.16 | 2.93 | 2.93 | 2.93 | 3.06 | 3.14 | 12.35 | 4.67 | 4.75 | 4.75 | 4.24 | 4.15 |
| 15 | 2.72 | 1.35 | 1.29 | 1.32 | 1.81 | 1.82 | 2.51 | 1.20 | 1.29 | 1.25 | 1.44 | 1.49 | 2.93 | 1.36 | 1.27 | 1.37 | 1.63 | 1.62 |
| 16 | 2.60 | 1.37 | 1.14 | 1.17 | 1.38 | 1.39 | 2.35 | 0.93 | 1.06 | 1.03 | 1.36 | 1.38 | 2.46 | 0.94 | 1.00 | 1.00 | 1.42 | 1.42 |
| 17 | 163.98 | 76.23 | 12.29 | 88.02 | 21.47 | 21.23 | 10.84 | 4.21 | 4.16 | 4.27 | 4.55 | 6.00 | 20.23 | 5.32 | 5.47 | 5.67 | 6.04 | 5.99 |
| 18 | 13.19 | 6.86 | 6.07 | 6.28 | 4.38 | 4.30 | 7.23 | 3.69 | 3.49 | 3.41 | 3.05 | 2.94 | 9.41 | 4.98 | 4.94 | 4.98 | 3.58 | 3.54 |
| 19 | 2.13 | 0.98 | 1.04 | 0.96 | 1.45 | 1.44 | 5.33 | 2.08 | 2.11 | 2.12 | 2.19 | 2.15 | 4.75 | 1.93 | 2.16 | 2.14 | 2.08 | 2.01 |
| 20 | 2.62 | 1.22 | 1.23 | 1.25 | 1.62 | 1.63 | 2.81 | 1.28 | 1.32 | 1.27 | 1.58 | 1.61 | 2.73 | 1.24 | 1.16 | 1.21 | 1.70 | 1.72 |
| AR | 5.80 | 3.35 | **1.95** | 2.65 | 3.60 | 3.65 | 5.90 | **2.30** | 2.60 | 2.45 | 3.70 | 4.05 | 5.90 | **2.65** | 2.55 | 3.20 | 3.50 | 3.20 |

Table 10: Results for the squared reject loss $\ell^{SR}_{\alpha,\rho}$ for the datasets in Table 12, using the experimental methodology in section 2.5. Each row aggregates the folds and five different values for $\alpha$ with $\alpha \in \{0, 0.25, 0.5, 0.75, 1\}$ and ten different values for $\rho$ with $r \in \{0, 0.111, 0.222, \ldots, 1\}$ and $\rho = \frac{1}{2}\sigma(D_Y)\frac{r}{1-r}$ (totalling 50 variations per fold). For visibility all the losses are multiplied by 10. Each section of six columns shows results for different base techniques (*LR*, *kNN* and *Tree*). The average ranks (AR) are calculated for these three groups separately. The Friedman statistic for the three sections are (48.4, 54.03 and 43.23 respectively), which are greater than the Critical Value (12.57). This means that the null hypothesis is rejected (significance level: 0.05) and the methods do not perform equally. Differences in average ranks higher than the critical difference for the Nemenyi post-hoc test (0.5217) imply that the difference is significant (in bold).

- We develop new *straightforward enrichment methods which show good performance* as conditional density estimators.

- We show that local reframing has a broad range of applicability over many different problems. We illustrate its *effectiveness on three families of problems* (bid applications, asymmetric loss applications and rejection rule applications) by the theoretical derivation of the expression that leads to optimal reframing in each case and a thorough empirical validation against other approaches, such as global reframing.

These contributions are important for machine learning because regression, unlike classification, has lacked a comprehensive and effective approach to deal with cost-sensitive problems by the reuse (and not a retraining) of general regression models.

## 7.2 Alternative approaches and other applications

The experimental results shown in previous sections could even be better for probabilistic reframing if we are able to get better enrichment methods (or other normal conditional density estimation methods). In fact, we envisage an intensive work in this line, similar to what was done in the last decade for probability estimation in classification, where many classification methods were rethought and redesigned to get good probabilities or good rankings (e.g., probability estimation trees [54, 26, 24]). Similarly, an important progress was made in calibration methods [70, 3]. Other possibilities for soft regression models could be conceived, leading to possibly simpler minimisation solutions (e.g., triangular or uniform distributions). Also, distributions with more parameters (e.g., asymmetric normal, truncated normal or Lévy distributions) could be explored.

The applicability of the enrichment and reframing methods for a diversity of base regression techniques (from non-parametric regression trees and *kNN* to parametric *LR*) with very different mathematical and statistical properties has led us to validate the approaches experimentally. However, the definition of specific combinations of a base technique with a particular enrichment method (e.g., *kNN* with *uKNC*, or *LR* with $ENR-LR$) could be analysed theoretically to derive statistical properties that may characterise their general behaviour better.

One important element in the development and improvement of soft and probabilistic regression models is the use of appropriate evaluation metrics and graphical representations, prior to any specific context-sensitive application, as has been done here in section 3 (before sections 4, 5, 6). We have used *msll* and *msvr* here, but other (new) metrics could be used, inspired by classification metrics[29] and plots such as ROC curves, cost curves [17], Brier curves [41], calibration plots, etc., and their derived measures, such as AUC [33, 27].

This paper has only included some representative applications, by choosing some common loss functions. There are, of course, many other domains and possible loss functions. For instance, tolerance is a concept that has been frequently used to bring ideas from classification to regression, since a tolerance level can be used to classify estimations as 'correct' or 'incorrect'. An example of a general tolerance loss can be defined as follows, by considering asymmetric losses and asymmetric tolerance levels for overestimations and underestimations:

**Definition 11.** *The tolerance loss* $\ell_{T,\alpha,\tau^-,\tau^+}$ *is a loss function defined as follows:*

$$\begin{aligned}
\ell_{T,\alpha,\tau^-,\tau^+}(\hat{y},y) &= \quad \alpha \quad &\text{if } \hat{y}+\tau^- < y \\
&= (1-\alpha) \quad &\text{if } \hat{y}-\tau^+ > y \\
&= \quad 0 \quad &\text{otherwise}
\end{aligned}$$

A related loss could originate from ordinal prediction if we define a loss function in such a way that it is 0 if the prediction is inside the bin of the discretisation (e.g., low (0..3), mid (3..7), high(7..10)), and, say, the number of bins it has to cross to go to the right bin otherwise. It would be interesting to see how probabilistic reframing could work in these two cases.

Apart from the loss functions which relate the true value *y* and the estimated value $\hat{y}$, there are many other kinds of costs and contexts [66]. For instance, loss functions can be instance-dependent, such as those that are a function of the input values, represented as $\ell(\hat{y},y,x)$. It is important to note again that this invalidates global reframing methods (but not local reframing methods). More generally, we can even have a relevance or prior distribution $U(x)$. This can be addressed by giving more relevance to some examples than others, in the integration (or sum) of the overall estimated cost ($\sum U(x)\ell(\hat{y},y)$) or in graphical representations, such as ROCIV (instance-varying ROC curves, [23]). In other cases, this relevance function can be more complex, as in the so-called utility-based regression [64, 56]. One way or another, it is important to realise that the methods in this paper are applicable when there is a change of the prior distribution, since the minimisation of the loss is local to each example (the experiments in this paper used 2-fold cross-validation without shuffling to simulate this situation). Changing the output distribution (or relevance) does not change the methods, since each optimisation is independent from the rest.

Finally, a soft regression model (issuing probabilities, reliabilities or confidence intervals) can be useful for some tasks, such as quantification [34] for regression, in the same way it has been shown beneficial for quantification for classification [6]. Also, screening applications can also take advantage of enrichment methods, since some elements in a rank could be considered a tie if their conditional distributions overlap a certain degree (possibly determining this with a test over the two normals, such as a KS-statistic). This can be applied to preference learning, where we can answer not only whether for two given examples $x_1$ and $x_2$ we have that $\hat{y}_1 > \hat{y}_2$, but we can also calculate the probability $Prob(\hat{y}_1 > \hat{y}_2)$ (if the regression model is probabilistic). This, for instance, suggests an evaluation metric related to the Wilcoxon-Mann-Whitney statistic interpretation of the *AUC* (area under the ROC curve), simply as $Prob(\hat{y}_1 > \hat{y}_2 | y_1 > y_2)$.

### 7.3 Concluding remarks

The goal of the paper was to show that cost-sensitive applications in regression can be successfully handled by a probabilistic reframing using enriched regression models in the form of a two-parameter normal conditional distribution. In order to accomplish this goal we needed to compare enrichment methods to other approaches for conditional density estimation in terms of estimation quality and efficiency. Another important issue that we needed to consider is the simplicity of the expressions leading to the optimal reframing with minimum expected loss (minimum risk). The choice of a normal distribution consummates all this, and consolidates a view of regression as a two-parameter estimation problem: conditional mean and variance. Also, we have seen that we can *enrich* any existing regression technique with reasonable good variance estimations, using some existing techniques and, most especially, some novel *enrichment* methods that are extremely lightweight. Enrichment methods in regression are somewhat similar to calibration methods in classification. However, the key difference is that the original prediction is kept and complemented by a second parameter, the variance.

Other approaches for context-sensitive applications build a model which is specialised for a very specific context, embedding the context in the model. In reframing, we reuse a general model for a wide range of contexts and operating contexts. The philosophy is completely different: models can be reused and validated across different operating contexts, improving robustness and efficiency.

Local reframing uses information about each prediction (reliability, confidence or probability) to adapt each local prediction. When we have probabilities (from the use of a conditional density function, e.g., a normal distribution), we can solve the decision rules analytically and, in cases where a closed form cannot be derived, use simple numerical approximations. In fact, probabilistic reframing only needs to derive the conditional variance once and for all, either from the method itself (e.g., regression trees derive this variance as the variance in each leaf of the tree) or by the use of enrichment methods, which only require the comparison to the output value. Once a regression model is equipped with a good conditional variance estimation we can apply the model to a variety of problems. Moreover, we can even use a different loss function (or different loss parameters) for each individual example.

Global reframing, on the other hand, tries to infer one global function from the training set which is applied to all the examples. This implies an optimisation procedure over the whole training set whenever the loss function (or any of its parameters) changes. Also, except for some convex loss functions where some efficient numerical methods can be used, this procedure may be time-consuming. These differences and the fact that the experimental results are, in general, favourable to probabilistic reframing, suggest that an instance-based (local) approach is the best option for reframing. It also links much better to the areas of risk minimisation in decision theory.

Overall, this paper contains a number of contributions and integrates a wide range of techniques that should trigger further research on conditional variance estimation, enrichment methods, calibration tech-niques for regression, evaluation metrics for regression, and better reframing techniques on these and other context-sensitive applications of regression models.

# Acknowledgements

Some ideas in this paper have benefited from early discussions with Peter Flach, around the notions of generalising operating conditions and reframing, as for the view shown in Table 1 and the relation to conformal prediction. I'm also grateful to Nicolas Lachiche for his very useful comments and suggestions on a preliminary version of this paper. This work was supported by the MEC/MINECO projects CONSOLIDER-INGENIO CSD2007-00022 and TIN 2010-21062-C02-02, and GVA project PROMETEO/2008/051. Finally, part of this work was motivated by the *REFRAME* project (http://www.reframe-d2k.org) granted by the European Coordinated Research on Long-term Challenges in Information and Communication Sciences & Technologies ERA-Net (CHIST-ERA), and funded by the respective national research councils and ministries.

# A  Datasets

Datasets, shown in tables 11 and 12, are obtained from eight packages of the CRAN distribution of *R*-project [55], namely: 'class', 'boot', 'MASS', 'nlme', 'lattice', 'np', 'survival' and 'farway'. Some of them have been processed to eliminate redundant or null attributes. All the datasets and the scripts in *R* (for the methods and tests) are available at http://www.dsic.upv.es/~jorallo/reframe-reg/scripts-data.zip.

|    | name          | size | attr | TrTeMD | TrTeKS |
|----|---------------|------|------|--------|--------|
| 1  | seatbelts     | 192  | 7    | 1.22   | 0.53   |
| 2  | theoph        | 132  | 4    | 0.10   | 0.11   |
| 3  | USjudgeratings| 44   | 12   | 0.24   | 0.25   |
| 4  | cars          | 50   | 2    | 1.72   | 0.74   |
| 5  | faithful      | 272  | 2    | 0.02   | 0.07   |
| 6  | boston        | 506  | 14   | 0.40   | 0.24   |
| 7  | UScrime       | 48   | 16   | 0.32   | 0.19   |
| 8  | gilgais       | 364  | 9    | 0.20   | 0.10   |
| 9  | wtloss        | 52   | 2    | 3.57   | 1.00   |
| 10 | cefamandole   | 84   | 3    | 0.37   | 0.14   |
| 11 | dialyzer      | 140  | 4    | 0.42   | 0.47   |
| 12 | earthquake    | 182  | 5    | 0.31   | 0.18   |
| 13 | gasoline      | 32   | 6    | 2.09   | 0.75   |
| 14 | glucose       | 376  | 4    | 0.07   | 0.10   |
| 15 | IGF           | 236  | 3    | 0.03   | 0.11   |
| 16 | nitrendipene  | 88   | 4    | 0.09   | 0.14   |
| 17 | wheat         | 48   | 4    | 0.74   | 0.42   |
| 18 | environmental | 112  | 4    | 0.28   | 0.20   |
| 19 | wage1         | 526  | 21   | 0.23   | 0.12   |
| 20 | ozone         | 330  | 10   | 0.28   | 0.20   |

Table 11: Dataset battery used in section 3 and related appendices. We show the size, the number of attributes, the relative difference in means (of the output value) between train and test (TrTeMD) and the Kolmogorov-Smirnoff statistic between train and test (TrTeKS).

|    | name       | size | attr | TrTeMD | TrTeKS |
|----|------------|------|------|--------|--------|
| 1  | iris3      | 150  | 4    | 2.48   | 0.70   |
| 2  | savings    | 50   | 5    | 0.22   | 0.20   |
| 3  | USarrests  | 50   | 4    | 0.38   | 0.20   |
| 4  | rock       | 48   | 4    | 7.73   | 0.83   |
| 5  | trees      | 32   | 3    | 3.37   | 0.94   |
| 6  | salinity   | 28   | 4    | 0.16   | 0.43   |
| 7  | birthwt    | 188  | 10   | 0.67   | 0.63   |
| 8  | menarche   | 24   | 3    | 4.97   | 1.00   |
| 9  | road       | 26   | 6    | 0.09   | 0.46   |
| 10 | stormer    | 24   | 3    | 0.32   | 0.25   |
| 11 | bodyweight | 176  | 4    | 11.09  | 1.00   |
| 12 | oxboys     | 234  | 3    | 0.10   | 0.09   |
| 13 | oecdpanel  | 616  | 7    | 0.58   | 0.26   |
| 14 | lungcancer | 168  | 10   | 0.67   | 0.33   |
| 15 | chicago    | 48   | 7    | 0.34   | 0.17   |
| 16 | diabetes   | 402  | 3    | 0.11   | 0.07   |
| 17 | divusa     | 76   | 7    | 0.54   | 0.55   |
| 18 | exa        | 256  | 3    | 0.40   | 0.47   |
| 19 | prostate   | 96   | 9    | 1.33   | 0.50   |
| 20 | seatpos    | 38   | 9    | 0.28   | 0.26   |

Table 12: Dataset battery used in sections 4, 5 and 6. We show the size, the number of attributes, the relative difference in means (of the output value) between train and test (TrTeMD) and the Kolmogorov-Smirnoff statistic between train and test (TrTeKS).

## B  Evaluation metrics for conditional density estimators

We present three metrics for evaluating the conditional mean, the conditional variance and the conditional density of a soft regression model. Since we need to work with any possible base regression model we present general measures instead of technique-specific measures for particular goodness-of-fit, parameter estimation or intrinsic variance estimation. In order to make results more commensurate and easier to compare, for all the measures which are not in the interval $[0,1]$, we will apply the logistic function $\Lambda(t) \triangleq \frac{1}{1+e^{-t}}$. We will use the word 'standardised' to refer to this logistic normalisation. In some cases, we will apply the function $1-t$ or other transformations to always get a decreasing $[0,1]$ scale (0 for very good estimations and 1 for very bad estimations).

The evaluation of the conditional mean or expected value is usually measured by the mean squared error $\frac{1}{|D|}\sum_{\langle x,y \rangle \in D}(y-\hat{m}(x))^2$ over a dataset $D$, although other metrics are also common, such as the mean absolute error, several correlation indices, mean relative squared error, etc. We will use the *mean relative squared error* (*mrse*) to make the measure less dependent on the dataset and easy to compare with the constant (trivial) regression model (the model which always outputs the mean of the training dataset, $\mu(D_y)$):

$$mrse(\hat{f},D) \triangleq 2\Lambda\left(\frac{\sum_{\langle x,y \rangle \in D}(y-\hat{m}(x))^2}{\sum_{\langle x,y \rangle \in D}(y-\mu(D_y))^2}\ln 3\right) - 1 \qquad (13)$$

The factor $\ln 3$ and the linear transformation makes that we get 0.5 if the error is the same as the constant (trivial) model, 0 for a perfect regression model and close to 1 for very bad estimations.

A more complex issue is to evaluate the quality of conditional density estimators. One possibility is the mean squared error for the distributions, i.e. $\int (f(y|x) - \hat{f}(y|x))^2$, but cannot be properly seen as an evaluation metric, since $f(y|x)$ is usually unknown or is pointwise infinite if we calculate this for a given dataset. This also happens for other distribution divergences, such as the KL-divergence. Consequently, one common measure is the mean negative log-likelihood (*nll*), which is defined as $\frac{1}{|D|}\sum_{\langle x,y \rangle \in D} -\ln(\hat{f}(y|x))$. We will again use the logistic function, applied to the log-likelihood, i.e.: $v \triangleq \Lambda(\ln(\hat{f}(y|x))) = \frac{1}{1+e^{-\ln(\hat{f}(y|x))}} =$ $\frac{1}{1+\frac{1}{\hat{f}(y|x)}}$. From here, we just switch to $1-v$ to get 0 for very good estimations and 1 for very bad estimations, and derive the mean standardised likelihood (*msll*), as follows.

$$msll(\hat{f}, D) \triangleq \frac{1}{|D|} \sum_{\langle x,y \rangle \in D} 1 - \frac{1}{1+\frac{1}{\hat{f}(y|x)}} = \frac{1}{|D|} \sum_{\langle x,y \rangle \in D} \frac{1}{1+\hat{f}(y|x)} \quad (14)$$

The log-likelihood (or its logistic variant) evaluates, at the same time, the quality of the mean and the quality of the variance. If we want a measure of the latter only, a possibility might be the squared error between the residual and the standard deviation. However, this measure also depends on how well the means are estimated, since when mean estimations are accurate (respectively inaccurate) the residuals are low (respectively high), and variances would tend to be low (respectively high) as well. An alternative is to calculate the variance ratio versus the squared residual. If we denote the residual as $res_{\hat{f}}(x,y) \triangleq (\hat{\mu}_{\hat{f}}(x) - y)$, we can define the variance ratio as $vr_{\hat{f}}(x,y) \triangleq \frac{\hat{\sigma}_{\hat{f}}(x)^2}{res_{\hat{f}}(x,y)^2}$. The numerator is the estimated variance and the denominator is the squared residual. This ratio will be close to 1 if both quantities are similar. If both the numerator and the denominator are 0, $vr_{\hat{f}}(x,y) = 1$ by definition. From here, the mean standardised variance ratio is given by the logistic function of the log ratio:

$$msvr(\hat{f}, D) \triangleq \frac{1}{|D|} \sum_{\langle x,y \rangle \in D} \left| 1 - 2\Lambda(\ln(vr_{\hat{f}}(x,y))) \right| = \frac{1}{|D|} \sum_{\langle x,y \rangle \in D} \left| 1 - 2\frac{1}{1+vr_{\hat{f}}(x,y)} \right| \quad (15)$$

This measure is always between 0 and 1, with 0 being a perfect variance estimation (variance is always equal to the squared residual) and 1 being the worst variance estimation (variance being much higher or much lower than the squared residual).

## C  Conditional density estimation methods

It can be argued that if we want to obtain a conditional density estimation, we should use conditional density estimation techniques, instead of crisp regression methods. Conditional density estimation techniques [44] are methods which directly[4] obtain $\hat{f}(y|x)$. While this is the most general and informative way for the regression problem (since conditional means, variances, confidence intervals and other measures can be obtained from it), the techniques are usually slower and suffer from a number of restrictions. A general way to tackle this estimation is through non-parametric methods (see, e.g., [43]). For instance, many approaches are restricted to only one (or two input variables, such as R's `hdrcde` package [44]), or just calculate multivariate densities, which have to be normalised for each input value $x$ to get a univariate density.

It is not the goal of this paper to evaluate several of the approaches for density estimation methods, but it is important to see whether these methods are better, in general, than 'augmented' or 'enriched' methods for which just a conditional mean and variance are obtained (from which a simple Gaussian density function is estimated). In addition, we are interested in the result of 'reducing' a local and detailed density function into a Gaussian. In order to do all this, we illustrate this approach with a kernel-based (non parametric) conditional density method. We use the function `npcdens` in the R's `np` (non parametric) package. This function computes estimates for the density function (i.e., $\hat{f}(y|x)$), for a bandwidth specification using the method in [39]. From the density we calculate the conditional mean and the conditional variance as a pointwise average which approximates the integral of the expected value and the moment respectively. We used the median point and the four points at $\pm\sigma$ and $\pm 2\sigma$, given by a Gaussian.

---
[4]Some other methods calculate the joint distribution $\hat{f}(y,x)$ or the likelihood $\hat{f}(x|y)$, from which the conditional density is just derived by dividing by $\hat{f}(x)$ or applying Bayes theorem respectively. However, this is usually more complex than the original problem.

|   | CDE gaus mrse | CDE gaus msll | CDE gaus msvr | CDE orig msll | LR CDE msll | LR CDE msvr | kNN CDE msll | kNN CDE msvr | Tree CDE msll | Tree CDE msvr |
|---|---|---|---|---|---|---|---|---|---|---|
| 1 | 0.47 | 0.83 | 0.53 | 0.64 | 0.83 | 0.49 | 0.83 | 0.49 | 0.83 | 0.51 |
| 2 | 0.29 | 0.72 | 0.41 | 0.66 | 0.72 | 0.53 | 0.72 | 0.50 | 0.72 | 0.53 |
| 3 | 0.05 | 0.51 | 0.57 | 0.51 | 0.50 | 0.61 | 0.51 | 0.66 | 0.51 | 0.67 |
| 4 | 0.52 | 0.88 | 0.71 | 0.89 | 0.88 | 0.55 | 0.88 | 0.56 | 0.88 | 0.52 |
| 5 | 0.17 | 0.68 | 0.49 | 0.63 | 0.68 | 0.54 | 0.68 | 0.51 | 0.68 | 0.52 |
| 6 | 0.14 | 0.63 | 0.55 | 0.60 | 0.64 | 0.63 | 0.63 | 0.61 | 0.63 | 0.58 |
| 7 | 0.46 | 0.76 | 0.52 | 0.73 | 0.75 | 0.63 | 0.76 | 0.60 | 0.74 | 0.57 |
| 8 | 0.17 | 0.62 | 0.53 | 0.60 | 0.63 | 0.53 | 0.61 | 0.55 | 0.62 | 0.58 |
| 9 | 0.50 | 0.97 | 0.78 | 0.99 | 0.97 | 0.48 | 0.97 | 0.56 | 0.97 | 0.55 |
| 10 | 0.37 | 0.59 | 0.53 | 0.54 | 0.59 | 0.61 | 0.59 | 0.67 | 0.58 | 0.62 |
| 11 | 0.30 | 0.72 | 0.55 | 0.81 | 0.72 | 0.66 | 0.72 | 0.60 | 0.72 | 0.68 |
| 12 | 0.58 | 0.76 | 0.57 | 0.73 | 0.76 | 0.75 | 0.76 | 0.58 | 0.76 | 0.47 |
| 13 | 0.28 | 0.77 | 0.56 | 0.75 | 0.77 | 0.62 | 0.76 | 0.73 | 0.77 | 0.71 |
| 14 | 0.40 | 0.74 | 0.46 | 0.66 | 0.74 | 0.49 | 0.74 | 0.52 | 0.74 | 0.58 |
| 15 | 0.84 | 0.80 | 0.46 | 0.72 | 0.80 | 0.69 | 0.80 | 0.65 | 0.80 | 0.64 |
| 16 | 0.64 | 0.81 | 0.63 | 0.77 | 0.81 | 0.56 | 0.81 | 0.56 | 0.81 | 0.54 |
| 17 | 0.40 | 0.77 | 0.43 | 0.94 | 0.78 | 0.81 | 0.77 | 0.55 | 0.77 | 0.58 |
| 18 | 0.63 | 0.82 | 0.45 | 0.74 | 0.82 | 0.55 | 0.82 | 0.57 | 0.82 | 0.57 |
| 19 | 0.60 | 0.76 | 0.41 | 0.67 | 0.76 | 0.59 | 0.76 | 0.55 | 0.76 | 0.60 |
| 20 | 0.36 | 0.73 | 0.55 | 0.71 | 0.73 | 0.63 | 0.73 | 0.53 | 0.73 | 0.55 |
| Mean | 0.41 | 0.74 | 0.53 | 0.72 | 0.74 | 0.60 | 0.74 | 0.58 | 0.74 | 0.58 |

Table 13: Results (using the datasets in Table 11) for the kernel-based (non parametric) conditional density method given by the function `npcdens` in the `R`'s `np` (non parametric) package. The first three columns show the results for the parametric density estimation (using the Gaussian approximation). The column "CDE orig msll" shows the *msll* result by using the original non-parametric density function. The six rightmost columns show the results for the mean given by the base method with the variance estimation given by the conditional density method (using the Gaussian approximation). Results for *mrse* are not shown for the six most right columns since they are equal to Table 2.

Table 13 shows the results of this method. Non-parametric conditional density estimation methods can get good estimations for large datasets with complex densities (e.g., bimodal) but here we see that the results are, in general, worse than those of simple regression methods such as a *kNN* or *Tree* for the conditional mean, as shown in Table 2. In fact, while the conditional variances seem better (0.53 in front of 0.70, 0.56 and 0.58 in Table 2.), the conditional densities are not better (except for *LR*) and the squared error (*mrse*) is also worse. A possible idea is then to combine the good conditional means from the base techniques with the conditional variance from the conditional density estimation method. This is what the six last columns show. However, as expected, this does not increase the quality of the conditional densities, because conditional variance estimates must be *linked* to a mean estimation. Consequently, neither as a standalone method nor combined with the base classifier can we get better performance. Also, this method is about two orders of magnitude slower than the direct methods in subsection 3.1 (and many other methods that we see in the rest of section 3).

## D  Conditional variance estimation methods

Instead of deriving a full conditional density, we can just (re-)use the conditional mean of a classical (crisp) regression method and derive a conditional variance. A usual, but not generally well-known, way of estimating the conditional variance is as follows (steps 1 and 2 can be omitted if we already have a regression model):

**Definition 12.** *Given a training or validation set T, and a (test) instance x, the* two-step conditional variance estimation *method 2SCVE is defined as follows:*

1. Train a regression model $m_y$ using $T$.
2. Obtain $\hat{y}_i \leftarrow m_y(x_i)$ for each example $\{\langle x_i, y_i \rangle\} \in T$.
3. Calculate the residuals: $u_i \leftarrow (y_i - \hat{y}_i)$.
4. Apply a transformation function $\theta$: $v_i \leftarrow \theta(u_i)$.
5. Train a regression model $m_v$ for the dataset $H = \{\langle x_i, v_i \rangle\}$.
6. Obtain $\hat{y} = m_y(x)$ and $\hat{v} = m_v(x)$ for the example $x$ to be predicted (in the test set).

*This estimates the conditional mean as $\hat{\mu}(x) = \hat{y}$ and the conditional standard deviation as $\hat{\sigma}(x_j) = \theta^{-1}(\hat{v})$.*

Usual choices for $\theta(t)$ are $\theta(t) = t^2$ and $\theta(t) = \ln(t^2)$, i.e., we model the (logarithm of) the squared residuals [69, 67]. The square is usually included since these methods are aimed at estimating the variance and also because otherwise we would need to remove the sign.

The estimation will depend on the quality of the regression model $m_y$ and most especially on the second regression model $m_v$. In fact, the previous algorithm (from steps 1 to 5) is usually iterated by retraining the regression model $m_y(x)$ using the heteroscedasticity information about the recently estimated conditional variance. This information can only be used by some regression techniques, e.g., a weighted least squares with inverse variance weights. This is usually called *iteratively re-weighted least squares*.

At this point it is important to notice that we really estimate the variance of the residuals of our model conditional to $x$, not the variance of $y$ conditional to $x$. Only if $m_y$ is a perfect regression model, these two variances will be equal.

It is usual to apply a non-parametric model in step 5. For instance, if we use nearest neighbours, the previous algorithm boils down to estimating $\hat{\sigma}^2(x_i)$ as the mean of the squared residuals of the $k$-closest examples, which is similar to what we did for *kNN* in section 3.1.

Here we will explore the *kNN* and *Tree* techniques for the residual model $m_u(x)$, jointly with the three base techniques *LR*, *kNN* and *Tree* as usual. Table 14 shows some of these methods for $\theta(t) = t^2$ (we ran the same experiments with other configurations of $\theta$ with equal or worse results). The results only show an improvement for *LR* compared to the results in Table 2. For *kNN* or *Tree*, the results are worse than the results in Table 2.

# E  Conditional variance estimation based on reliability

A reliability measure for regression is any numerical value which is directly related to the degree of certainty about an accurate prediction being produced or, more precisely, inversely related to the expected (absolute) residual. However, the magnitude of this value can follow any scale. Bosnic & Kononenko [8] compare several reliability estimators for regression. Among them, *CNK* is a simple method which shows good performance (as a reliability estimator). This method works as follows. For each example $\langle x, y \rangle$, this method just calculates de $k$-closest elements in the training set to $x$ and calculates the mean of their output values, denoted by $C$ or, in other words, calculates the *kNN* prediction for $x$. Then it calculates the absolute difference between $C$ and the prediction (of presumably a regression technique which is not *kNN*). This is the estimated standard deviation.

The previous approach makes an average of the true values, and then compares this to a single estimation. While this is good as a reliability measure, the magnitude of the estimation will typically be low (for a standard deviation), since it just compares the prediction of two methods. Consequently, we suggest a correction, which goes as follows:

|   | LR cve kNN msll | LR cve kNN msvr | LR cve Tree msll | LR cve Tree msvr | kNN cve kNN msll | kNN cve kNN msvr | kNN cve Tree msll | kNN cve Tree msvr | Tree cve kNN msll | Tree cve kNN msvr | Tree cve Tree msll | Tree cve Tree msvr |
|---|---|---|---|---|---|---|---|---|---|---|---|---|
| 1 | 0.78 | 0.50 | 0.78 | 0.53 | 0.80 | 0.50 | 0.80 | 0.52 | 0.83 | 0.62 | 0.83 | 0.62 |
| 2 | 0.76 | 0.47 | 0.77 | 0.49 | 0.75 | 0.46 | 0.75 | 0.46 | 0.67 | 0.53 | 0.68 | 0.53 |
| 3 | 0.53 | 0.67 | 0.55 | 0.70 | 0.58 | 0.59 | 0.58 | 0.58 | 0.66 | 0.61 | 0.62 | 0.52 |
| 4 | 0.71 | 0.57 | 0.72 | 0.58 | 0.81 | 0.50 | 0.81 | 0.50 | 0.78 | 0.52 | 0.81 | 0.61 |
| 5 | 0.63 | 0.54 | 0.63 | 0.53 | 0.63 | 0.52 | 0.64 | 0.52 | 0.63 | 0.53 | 0.63 | 0.53 |
| 6 | 0.69 | 0.59 | 0.69 | 0.59 | 0.65 | 0.59 | 0.65 | 0.57 | 0.61 | 0.56 | 0.62 | 0.57 |
| 7 | 0.77 | 0.71 | 0.80 | 0.74 | 0.76 | 0.60 | 0.76 | 0.64 | 0.76 | 0.59 | 0.77 | 0.58 |
| 8 | 0.63 | 0.53 | 0.64 | 0.55 | 0.60 | 0.52 | 0.61 | 0.53 | 0.58 | 0.53 | 0.59 | 0.56 |
| 9 | 0.95 | 0.91 | 0.94 | 0.90 | 1.00 | 0.94 | 0.99 | 0.90 | 1.00 | 0.95 | 1.00 | 0.95 |
| 10 | 0.70 | 0.43 | 0.70 | 0.42 | 0.63 | 0.55 | 0.66 | 0.57 | 0.56 | 0.51 | 0.57 | 0.56 |
| 11 | 0.69 | 0.44 | 0.73 | 0.52 | 0.74 | 0.64 | 0.76 | 0.68 | 0.67 | 0.64 | 0.67 | 0.64 |
| 12 | 0.88 | 0.72 | 0.89 | 0.75 | 0.76 | 0.46 | 0.81 | 0.61 | 0.72 | 0.52 | 0.72 | 0.55 |
| 13 | 0.57 | 0.64 | 0.56 | 0.57 | 0.89 | 0.63 | 0.91 | 0.67 | 0.90 | 0.69 | 0.91 | 0.69 |
| 14 | 0.74 | 0.46 | 0.74 | 0.44 | 0.72 | 0.51 | 0.71 | 0.52 | 0.67 | 0.56 | 0.66 | 0.53 |
| 15 | 0.73 | 0.61 | 0.73 | 0.60 | 0.73 | 0.59 | 0.73 | 0.59 | 0.73 | 0.58 | 0.72 | 0.55 |
| 16 | 0.76 | 0.57 | 0.76 | 0.55 | 0.71 | 0.55 | 0.71 | 0.55 | 0.68 | 0.58 | 0.68 | 0.58 |
| 17 | 0.97 | 0.87 | 0.98 | 0.87 | 0.79 | 0.60 | 0.78 | 0.59 | 0.75 | 0.62 | 0.76 | 0.63 |
| 18 | 0.74 | 0.54 | 0.74 | 0.56 | 0.75 | 0.56 | 0.75 | 0.55 | 0.74 | 0.54 | 0.75 | 0.55 |
| 19 | 0.69 | 0.55 | 0.69 | 0.55 | 0.70 | 0.51 | 0.71 | 0.54 | 0.70 | 0.57 | 0.70 | 0.57 |
| 20 | 0.80 | 0.64 | 0.79 | 0.62 | 0.68 | 0.54 | 0.69 | 0.56 | 0.69 | 0.56 | 0.69 | 0.58 |
| Mean | 0.74 | 0.60 | 0.74 | 0.60 | 0.74 | 0.57 | 0.74 | 0.58 | 0.72 | 0.59 | 0.72 | 0.60 |

Table 14: Results (using the datasets in Table 11) for several base methods (*LR*, *kNN* and *Tree*) with conditional variance estimation using *kNN* and *Tree* as models for the residuals. All the methods use $\theta(t) = t^2$. The results for *mrse* are not shown since they are equal to Table 2.

1. Given an example $\langle x, y \rangle$, we estimate $\hat{y}$ by any base regression technique.

2. Let $S = \langle x_i, y_i \rangle$ the set of the *k* nearest neighbours of *x* in a training or validation dataset.

3. Calculate $\sum_{\langle x_i, y_i \rangle \in S} (\hat{y} - y_i)^2$ as the output estimated variance.

Since it can be seen as a symmetric version to method CNK, we call it KNC. The results are shown in Table 15. As we can see, *CNK* is not a good variance estimation method in general. In addition, it cannot work, by definition, for *kNN*, since this method uses *kNN* as the true value, and the estimated 'residuals' will be 0. This can be clearly seen on the two columns "kNN CNK". On the contrary, *KNC* works well. In fact, it improves the results for the *LR* base technique shown in Table 2.

## F  Conditional variance estimation using conformal prediction

Confidence intervals are an alternative (and statistically convenient) way of measuring the reliability of a prediction. Conformal prediction [60, 50] is a general technique for deriving confidence intervals. It can be applied to any predictive task, such as classification and regression. While originally introduced for a transductive scenario, it has also been extended to a more classical inductive setting [49]. Conformal prediction works as follows. Given an error probability $\varepsilon$ and any regression method that makes a prediction $\hat{y}$, it produces a region $\Gamma^\varepsilon$ such that it contains the true value *y* in at least a proportion $1 - \varepsilon$ of the cases (the confidence level). Logically, by making the region infinitely large, we can get any confidence level. The key issue is that the tightness and therefore usefulness of the prediction region depends on the *nonconformity measure* used. A nonconformity measure is any measure that evaluates how unusual an example is (with respect to the others). In regression, for instance, a typical nonconformity measure is the absolute difference: $|y - \hat{y}|$.

|   | LR CNK msll | LR CNK msvr | kNN CNK msll | kNN CNK msvr | Tree CNK msll | Tree CNK msvr | LR KNC msll | LR KNC msvr | kNN KNC msll | kNN KNC msvr | Tree KNC msll | Tree KNC msvr |
|---|---|---|---|---|---|---|---|---|---|---|---|---|
| 1 | 0.86 | 0.75 | 0.99 | 0.99 | 0.87 | 0.72 | 0.78 | 0.49 | 0.80 | 0.51 | 0.82 | 0.49 |
| 2 | 0.83 | 0.72 | 1.00 | 1.00 | 0.75 | 0.65 | 0.77 | 0.47 | 0.74 | 0.45 | 0.74 | 0.62 |
| 3 | 0.51 | 0.73 | 1.00 | 1.00 | 0.62 | 0.46 | 0.60 | 0.85 | 0.60 | 0.61 | 0.66 | 0.59 |
| 4 | 0.75 | 0.66 | 1.00 | 1.00 | 0.91 | 0.80 | 0.77 | 0.70 | 0.81 | 0.53 | 0.77 | 0.49 |
| 5 | 0.73 | 0.68 | 0.99 | 0.99 | 0.79 | 0.77 | 0.63 | 0.54 | 0.63 | 0.51 | 0.64 | 0.53 |
| 6 | 0.72 | 0.53 | 1.00 | 1.00 | 0.66 | 0.62 | 0.69 | 0.49 | 0.65 | 0.57 | 0.64 | 0.60 |
| 7 | 0.75 | 0.53 | 1.00 | 1.00 | 0.76 | 0.55 | 0.78 | 0.60 | 0.76 | 0.61 | 0.78 | 0.52 |
| 8 | 0.70 | 0.62 | 1.00 | 1.00 | 0.66 | 0.64 | 0.65 | 0.55 | 0.61 | 0.53 | 0.61 | 0.59 |
| 9 | 0.72 | 0.64 | 1.00 | 1.00 | 1.00 | 0.97 | 0.73 | 0.65 | 0.99 | 0.92 | 0.96 | 0.86 |
| 10 | 0.76 | 0.59 | 1.00 | 1.00 | 0.66 | 0.62 | 0.69 | 0.39 | 0.59 | 0.44 | 0.62 | 0.56 |
| 11 | 0.79 | 0.63 | 1.00 | 1.00 | 0.74 | 0.74 | 0.73 | 0.52 | 0.74 | 0.63 | 0.76 | 0.75 |
| 12 | 0.85 | 0.41 | 1.00 | 1.00 | 0.81 | 0.71 | 0.87 | 0.42 | 0.76 | 0.48 | 0.77 | 0.54 |
| 13 | 0.74 | 0.83 | 1.00 | 1.00 | 0.95 | 0.82 | 0.77 | 0.91 | 0.89 | 0.63 | 0.87 | 0.53 |
| 14 | 0.80 | 0.67 | 1.00 | 1.00 | 0.74 | 0.65 | 0.75 | 0.47 | 0.72 | 0.51 | 0.72 | 0.64 |
| 15 | 0.74 | 0.66 | 1.00 | 1.00 | 0.80 | 0.72 | 0.73 | 0.62 | 0.73 | 0.59 | 0.73 | 0.58 |
| 16 | 0.78 | 0.55 | 1.00 | 1.00 | 0.77 | 0.67 | 0.74 | 0.47 | 0.72 | 0.56 | 0.72 | 0.62 |
| 17 | 0.91 | 0.46 | 1.00 | 1.00 | 0.77 | 0.62 | 0.89 | 0.36 | 0.78 | 0.55 | 0.77 | 0.54 |
| 18 | 0.80 | 0.75 | 1.00 | 1.00 | 0.77 | 0.62 | 0.75 | 0.55 | 0.75 | 0.56 | 0.76 | 0.57 |
| 19 | 0.74 | 0.66 | 1.00 | 1.00 | 0.75 | 0.65 | 0.72 | 0.59 | 0.70 | 0.51 | 0.72 | 0.58 |
| 20 | 0.80 | 0.52 | 0.99 | 0.99 | 0.73 | 0.62 | 0.77 | 0.45 | 0.68 | 0.52 | 0.70 | 0.55 |
| Mean | 0.76 | 0.63 | 1.00 | 1.00 | 0.78 | 0.68 | 0.74 | 0.55 | 0.73 | 0.56 | 0.74 | 0.59 |

Table 15: Results (using the datasets in Table 11) for several base techniques (*LR*, *kNN* and *Tree*) with Bosnic & Kononenko's *CNK* [8] and a more accurate variation that we dub *KNC*.

The idea of outputting a confidence region is richer and more informative than only a prediction point (which in regression is just the conditional mean). One advantage of confidence regions is that there is no assumption about the conditional distribution. However, this is one of its drawbacks for cost-sensitive learning, because we cannot quantify the probability of error in the prediction[5].

One way of deriving a density function from an interval is by assuming a distribution. Since we advocate for the normal distribution for context-sensitive applications, we can devise a simple method by assuming this distribution. In particular, for a normal distribution (with cumulative distribution $\Phi_{\mu,\sigma^2}$) we know that a proportion $p$ of the values inside $\mu \pm a\sigma$ is given by $p = \Phi_{\mu,\sigma^2}(\mu + a\sigma) - \Phi_{\mu,\sigma^2}(\mu - a\sigma)$. For a conformal region $\Gamma^\varepsilon$ we know that a proportion $1 - \varepsilon$ of the values fall inside the region. By taking different values for $a$ we can get different points where we can derive the correspondence. For instance, for $a = 1$, we get $p = 0.6827$. Setting $\varepsilon = 1 - 0.6827$ we then calculate the conformal region $\Gamma^{0.3173}$. The width of this region, denoted by $width(\Gamma^{0.3173})$, has to be $2a\sigma$. Since we chose $a = 1$, we have that:

$$\sigma = \frac{1}{2} width(\Gamma^{0.3173}) \quad (16)$$

We have implemented the inductive conformal regression presented in [51]. It presents seven nonconformity measures. The first one, that we denote by *A*, is just $|y - \hat{y}|$. The other six are just modifications which take some metrics of the *k*-nearest neighbours into account, such as the mean (or median) (input domain) distance in relation to the average distance of the training dataset, or the mean (or median) deviation (of the output domain), corresponding to formulas (24), (25), (29), (30), (31), (32) in [51]. Some of them have parameters ($\gamma$ and $\rho$), which we set to 0.5 (as in [51]).

---

[5]For instance, if we have two regions $\Gamma_1^\varepsilon = [3.2, 5.4]$ and $\Gamma_2^\varepsilon = [5.3, 15.9]$ for two different examples, we see that the first interval is much tighter. However, we cannot directly see whether an actual value of 5.2 has higher probability for the first example than a value of 14.2 for the second, because we cannot directly derive probabilities. A possible way of answering this specific question is by adjusting the error probability $\varepsilon$. If we increase our tolerance, we may get tighter intervals and we may see that some of the values fall out of the interval. However, we cannot derive the probability for each point either. In order to do this we need a conditional probability density function.

|   | LRc mrse | LRc msll | LRc msvr | kNNc mrse | kNNc msll | kNNc msvr | Treec mrse | Treec msll | Treec msvr | Conf mrse | Conf msll | Conf msvr |
|---|---|---|---|---|---|---|---|---|---|---|---|---|
| 1 | 0.28 | 0.79 | 0.60 | 0.35 | 0.81 | 0.59 | 0.41 | 0.83 | 0.60 | 0.41 | 0.83 | 0.56 |
| 2 | 0.52 | 0.78 | 0.53 | 0.39 | 0.77 | 0.53 | 0.19 | 0.68 | 0.53 | 0.40 | 0.76 | 0.55 |
| 3 | 0.02 | 0.76 | 0.82 | 0.16 | 0.64 | 0.67 | 0.21 | 0.67 | 0.66 | 0.22 | 0.65 | 0.61 |
| 4 | 0.12 | 0.77 | 0.67 | 0.32 | 0.84 | 0.59 | 0.25 | 0.79 | 0.50 | 0.37 | 0.83 | 0.48 |
| 5 | 0.11 | 0.65 | 0.56 | 0.11 | 0.65 | 0.53 | 0.11 | 0.64 | 0.52 | 0.11 | 0.64 | 0.52 |
| 6 | 0.55 | 0.70 | 0.61 | 0.20 | 0.66 | 0.63 | 0.16 | 0.62 | 0.58 | 0.20 | 0.66 | 0.60 |
| 7 | 0.46 | 0.75 | 0.63 | 0.45 | 0.77 | 0.60 | 0.50 | 0.76 | 0.57 | 0.41 | 0.74 | 0.56 |
| 8 | 0.14 | 0.64 | 0.54 | 0.13 | 0.62 | 0.56 | 0.12 | 0.59 | 0.58 | 0.13 | 0.62 | 0.57 |
| 9 | 0.07 | 0.92 | 0.87 | 0.29 | 0.94 | 0.79 | 0.26 | 0.97 | 0.89 | 0.33 | 1.00 | 0.94 |
| 10 | 0.31 | 0.70 | 0.46 | 0.31 | 0.71 | 0.74 | 0.23 | 0.61 | 0.61 | 0.34 | 0.76 | 0.79 |
| 11 | 0.19 | 0.70 | 0.45 | 0.15 | 0.75 | 0.61 | 0.16 | 0.67 | 0.64 | 0.16 | 0.74 | 0.61 |
| 12 | 0.89 | 0.88 | 0.68 | 0.47 | 0.84 | 0.72 | 0.45 | 0.77 | 0.70 | 0.52 | 0.79 | 0.66 |
| 13 | 0.02 | 0.69 | 0.71 | 0.45 | 0.86 | 0.42 | 0.44 | 0.85 | 0.40 | 0.47 | 0.86 | 0.42 |
| 14 | 0.45 | 0.78 | 0.55 | 0.32 | 0.73 | 0.56 | 0.22 | 0.70 | 0.61 | 0.32 | 0.74 | 0.58 |
| 15 | 0.50 | 0.70 | 0.55 | 0.55 | 0.73 | 0.57 | 0.56 | 0.71 | 0.56 | 0.57 | 0.71 | 0.57 |
| 16 | 0.41 | 0.77 | 0.57 | 0.26 | 0.71 | 0.54 | 0.22 | 0.67 | 0.55 | 0.24 | 0.69 | 0.56 |
| 17 | 0.97 | 0.92 | 0.69 | 0.39 | 0.79 | 0.56 | 0.37 | 0.76 | 0.58 | 0.44 | 0.80 | 0.57 |
| 18 | 0.30 | 0.75 | 0.56 | 0.34 | 0.76 | 0.58 | 0.35 | 0.74 | 0.54 | 0.34 | 0.73 | 0.53 |
| 19 | 0.33 | 0.70 | 0.55 | 0.34 | 0.73 | 0.56 | 0.48 | 0.70 | 0.57 | 0.35 | 0.70 | 0.54 |
| 20 | 0.58 | 0.80 | 0.63 | 0.20 | 0.68 | 0.55 | 0.26 | 0.68 | 0.56 | 0.19 | 0.68 | 0.54 |
| Mean | 0.36 | 0.76 | 0.61 | 0.31 | 0.75 | 0.60 | 0.30 | 0.72 | 0.59 | 0.33 | 0.75 | 0.59 |

Table 16: Results (using the datasets in Table 11) for three regression methods for which the variance estimation has been replaced by the variance given by conformal prediction using the nonconformity measure $A$, denoted by *LRc*, *kNNc* and *Treec*. The method *Conf* keeps the mean which is estimated by conformal prediction.

We analysed the results for all the nonconformity measures, but we just show the results for the nonconformity measure $A$ in Table 16, since this measure gives the best results (although the results are relatively similar for all of them). Comparing to the results in Table 2, it seems that there is no improvement in the variance estimation, except for linear regression.

## G  Comparison between *NCDE* methods

At the end of section 3 we perform a selection of some of the NCDE methods seen in the section. In this appendix, we include the results for a selection of the most relevant methods: the own estimation from the base techniques (section 3.1), conformal prediction (appendix F), a conditional density estimation (*CDE*) method (appendix C), a conditional variance estimation (*CVE*) method using Tree for residual regression (appendix D), and three enrichment methods described in section 3.3: *RBE* (using Tree as residual regression), *uKNC* and *BIN*. For all these methods, Tables 17, 18 and 19 show the comparison for the base techniques *LR*, *KNN* and *Tree* respectively. We only include the results for the metric *msll*, which considers both the quality of the conditional mean and the conditional variance estimation. Note that only for Table 17 the results are significant, so other criteria (such as simplicity) are used to finally select the methods in section 3.

## H  Proofs

Here we include the proofs for several results in the paper.

|   | LR msll | LRc msll | LR CDE msll | LR CVE Tree msll | LR ENR Tree msll | LR ENR uKN msll | LR ENR CBIN msll |
|---|---|---|---|---|---|---|---|
| 1 | 0.81 | 0.79 | 0.83 | 0.78 | 0.78 | 0.78 | 0.78 |
| 2 | 0.80 | 0.78 | 0.72 | 0.77 | 0.78 | 0.78 | 0.78 |
| 3 | 0.48 | 0.76 | 0.50 | 0.55 | 0.55 | 0.57 | 0.54 |
| 4 | 0.69 | 0.77 | 0.88 | 0.72 | 0.72 | 0.77 | 0.72 |
| 5 | 0.84 | 0.65 | 0.68 | 0.63 | 0.63 | 0.63 | 0.63 |
| 6 | 0.76 | 0.70 | 0.64 | 0.69 | 0.70 | 0.71 | 0.71 |
| 7 | 0.75 | 0.75 | 0.75 | 0.80 | 0.76 | 0.75 | 0.75 |
| 8 | 0.76 | 0.64 | 0.63 | 0.64 | 0.64 | 0.64 | 0.64 |
| 9 | 1.00 | 0.92 | 0.97 | 0.94 | 0.94 | 0.73 | 0.95 |
| 10 | 0.76 | 0.70 | 0.59 | 0.70 | 0.70 | 0.67 | 0.69 |
| 11 | 0.77 | 0.70 | 0.72 | 0.73 | 0.74 | 0.72 | 0.72 |
| 12 | 0.88 | 0.88 | 0.76 | 0.89 | 0.86 | 0.85 | 0.86 |
| 13 | 0.58 | 0.69 | 0.77 | 0.56 | 0.58 | 0.72 | 0.58 |
| 14 | 0.85 | 0.78 | 0.74 | 0.74 | 0.74 | 0.74 | 0.74 |
| 15 | 0.75 | 0.70 | 0.80 | 0.73 | 0.74 | 0.75 | 0.75 |
| 16 | 0.78 | 0.77 | 0.81 | 0.76 | 0.75 | 0.74 | 0.75 |
| 17 | 0.99 | 0.92 | 0.78 | 0.98 | 0.97 | 0.91 | 0.98 |
| 18 | 0.76 | 0.75 | 0.82 | 0.74 | 0.74 | 0.74 | 0.74 |
| 19 | 0.72 | 0.70 | 0.76 | 0.69 | 0.69 | 0.69 | 0.69 |
| 20 | 0.82 | 0.80 | 0.73 | 0.79 | 0.79 | 0.79 | 0.79 |
| Mean | 0.78 | 0.76 | 0.74 | 0.74 | 0.74 | 0.73 | 0.74 |
| AR | 5.45 | 4.40 | 4.00 | 3.50 | 4.00 | 3.30 | 3.35 |

Table 17: Results (using the datasets in Table 11) for base technique *LR* using a selection of the methods seen in section 3 as described in appendix G. Methods do not perform equally since the Friedman statistic (14.68) is greater than the Critical Value (14.16), so the null hypothesis is rejected (significance level: 0.05). Critical difference for the Nemenyi post-hoc test: 0.6922.

.

*Proof.* (for proposition 1) We have that $r^*(x, \ell, \hat{f})$ can be written as follows:

$$
\begin{aligned}
r^*(x, \ell, \hat{f}) &= \arg\min_t \int_{-\infty}^{\infty} \ell(t, y) \hat{f}(y|x) dy \\
&= \arg\min_t \int_{-\infty}^{\infty} \ell(t, \hat{\mu}_{\hat{f}}(x) + s) \hat{f}((\hat{\mu}_{\hat{f}}(x) + s)|x) ds \\
&= \arg\min_t \left\{ \int_{-\infty}^{0} \ell(t, \hat{\mu}_{\hat{f}}(x) + s) \hat{f}((\hat{\mu}_{\hat{f}}(x) + s)|x) ds + \int_{0}^{\infty} \ell(t, \hat{\mu}(x) + s) \hat{f}((\hat{\mu}_{\hat{f}}(x) + s)|x) ds \right\} \\
&= \arg\min_t \left\{ \int_{-\infty}^{0} \ell(t, \hat{\mu}_{\hat{f}}(x) + s) \hat{f}((\hat{\mu}_{\hat{f}}(x) + s)|x) ds - \int_{-\infty}^{0} \ell(t, \hat{\mu}(x) - s) \hat{f}((\hat{\mu}_{\hat{f}}(x) - s)|x) ds \right\} \\
&= \arg\min_t \int_{-\infty}^{0} \left\{ \ell(t, \hat{\mu}_{\hat{f}}(x) + s) \hat{f}((\hat{\mu}_{\hat{f}}(x) + s)|x) - \ell(t, \hat{\mu}_{\hat{f}}(x) - s) \hat{f}((\hat{\mu}_{\hat{f}}(x) - s)|x) \right\} ds
\end{aligned}
$$

Since $\hat{f}$ is symmetric relative to the mean:

$$
r^*(x, \ell, \hat{f}) = \arg\min_t \int_{-\infty}^{0} \left\{ \ell(t, \hat{\mu}_{\hat{f}}(x) + s) - \ell(t, \hat{\mu}_{\hat{f}}(x) - s) \right\} \hat{f}((\hat{\mu}_{\hat{f}}(x) + s)|x) ds
$$

But $\ell$ is symmetric, so we have that for every $y$ and $r$ we have that $\ell(y+r, y) = \ell(y-r, y)$ which, jointly with its commutativity, implies $\ell(y, y-r) = \ell(y, y+r)$, so a minimum of the above expression can be found when $t = \hat{\mu}_{\hat{f}}(x)$, leading to the expression $\ell(\hat{\mu}_{\hat{f}}(x), \hat{\mu}_{\hat{f}}(x) + s) - \ell(\hat{\mu}_{\hat{f}}(x), \hat{\mu}_{\hat{f}}(x) - s) = 0$. So, $r^*(x, \ell, \hat{f}) = \hat{\mu}_{\hat{f}}(x)$. □ □

|   | kNN msll | kNNckNN msll | kNN CDE msll Tree msll | kNN CVE msll Tree msll | kNN ENR msll msll | kNN ENR msll | kNN ENR uKNCBIN msll |
|---|---|---|---|---|---|---|---|
| 1 | 0.80 | 0.81 | 0.83 | 0.80 | 0.80 | 0.80 | 0.80 |
| 2 | 0.75 | 0.77 | 0.72 | 0.75 | 0.76 | 0.77 | 0.77 |
| 3 | 0.61 | 0.64 | 0.51 | 0.58 | 0.62 | 0.63 | 0.64 |
| 4 | 0.81 | 0.84 | 0.88 | 0.81 | 0.81 | 0.84 | 0.81 |
| 5 | 0.63 | 0.65 | 0.68 | 0.64 | 0.64 | 0.63 | 0.63 |
| 6 | 0.65 | 0.66 | 0.63 | 0.65 | 0.66 | 0.65 | 0.66 |
| 7 | 0.76 | 0.77 | 0.76 | 0.76 | 0.76 | 0.77 | 0.77 |
| 8 | 0.61 | 0.62 | 0.61 | 0.61 | 0.61 | 0.61 | 0.61 |
| 9 | 0.99 | 0.94 | 0.97 | 0.99 | 0.99 | 0.99 | 0.99 |
| 10 | 0.59 | 0.71 | 0.59 | 0.66 | 0.67 | 0.60 | 0.60 |
| 11 | 0.74 | 0.75 | 0.72 | 0.76 | 0.76 | 0.74 | 0.75 |
| 12 | 0.77 | 0.84 | 0.76 | 0.81 | 0.76 | 0.76 | 0.75 |
| 13 | 0.89 | 0.86 | 0.76 | 0.91 | 0.90 | 0.90 | 0.89 |
| 14 | 0.72 | 0.73 | 0.74 | 0.71 | 0.71 | 0.71 | 0.71 |
| 15 | 0.73 | 0.73 | 0.80 | 0.73 | 0.73 | 0.75 | 0.75 |
| 16 | 0.72 | 0.71 | 0.81 | 0.71 | 0.72 | 0.71 | 0.71 |
| 17 | 0.78 | 0.79 | 0.77 | 0.78 | 0.83 | 0.80 | 0.80 |
| 18 | 0.75 | 0.76 | 0.82 | 0.75 | 0.75 | 0.75 | 0.75 |
| 19 | 0.70 | 0.73 | 0.76 | 0.71 | 0.70 | 0.70 | 0.70 |
| 20 | 0.68 | 0.68 | 0.73 | 0.69 | 0.69 | 0.68 | 0.68 |
| Mean | 0.73 | 0.75 | 0.74 | 0.74 | 0.74 | 0.74 | 0.74 |
| AR | 3.40 | 4.90 | 3.90 | 4.00 | 3.85 | 3.95 | 4.00 |

Table 18: Results (using the datasets in Table 11) for base technique *kNN* using a selection of the methods seen in section 3 as described in appendix G. Methods may perform equally since the Friedman statistic (5.164) is lower than the Critical Value (14.16), so the null hypothesis cannot be rejected (significance level: 0.05). Critical difference for the Nemenyi post-hoc test: 0.6922.

.

*Proof.* (for proposition 4) We use the expression for $\ell_\alpha^A(\hat{y}, y)$ and decompose it depending on whether $t < y$ or not.

$$
\begin{aligned}
\mathscr{L}(x,t,\hat{f},\ell_\alpha^A) &= \int_{-\infty}^{\infty} \ell_\alpha^A(t,y)\hat{f}(y|x)dy \\
&= \int_{-\infty}^{t} (1-\alpha)(t-y)\hat{f}(y|x)dy + \int_{t}^{\infty} \alpha(y-t)\hat{f}(y|x)dy \\
&= \int_{-\infty}^{t} (1-\alpha)t\hat{f}(y|x)dy - \int_{-\infty}^{t} (1-\alpha)y\hat{f}(y|x)dy + \int_{t}^{\infty} \alpha(y)\hat{f}(y|x)dy - \alpha\int_{t}^{\infty} t\hat{f}(y|x)dy \\
&= (1-\alpha)t\hat{F}(t|x) - \int_{-\infty}^{t} (1-\alpha)y\hat{f}(y|x)dy + \int_{t}^{\infty} \alpha y\hat{f}(y|x)dy - \alpha t(1-\hat{F}(t|x)) \\
&= (1-\alpha)t\hat{F}(t|x) - \int_{-\infty}^{t} y\hat{f}(y|x)dy + \int_{-\infty}^{\infty} \alpha y\hat{f}(y|x)dy - \alpha t(1-\hat{F}(t|x)) \\
&= (1-\alpha)t\hat{F}(t|x) - \int_{-\infty}^{t} y\hat{f}(y|x)dy + \alpha\hat{\mu}(x) - \alpha t(1-\hat{F}(t|x)) \\
&= \alpha\hat{\mu}(x) + t\hat{F}(t|x) - \alpha t - \int_{-\infty}^{t} y\hat{f}(y|x)dy
\end{aligned}
$$

☐ ☐

*Proof.* (for proposition 5) From proposition 4 we just derive the expression for minimising the expected

|    | Tree msll | Treec msll | Tree CDE msll | Tree CVE Tree msll | Tree ENR Tree msll | Tree ENR uKN msll | Tree ENR BIN msll |
|----|------|-------|------|------|------|------|------|
| 1  | 0.85 | 0.83 | 0.83 | 0.83 | 0.85 | 0.84 | 0.84 |
| 2  | 0.66 | 0.68 | 0.72 | 0.68 | 0.66 | 0.68 | 0.66 |
| 3  | 0.62 | 0.67 | 0.51 | 0.62 | 0.60 | 0.64 | 0.63 |
| 4  | 0.80 | 0.79 | 0.88 | 0.81 | 0.81 | 0.77 | 0.81 |
| 5  | 0.63 | 0.64 | 0.68 | 0.63 | 0.63 | 0.63 | 0.63 |
| 6  | 0.63 | 0.62 | 0.63 | 0.62 | 0.62 | 0.64 | 0.65 |
| 7  | 0.76 | 0.76 | 0.74 | 0.77 | 0.76 | 0.75 | 0.76 |
| 8  | 0.58 | 0.59 | 0.62 | 0.59 | 0.58 | 0.59 | 0.59 |
| 9  | 0.99 | 0.97 | 0.97 | 1.00 | 1.00 | 0.93 | 1.00 |
| 10 | 0.55 | 0.61 | 0.58 | 0.57 | 0.57 | 0.56 | 0.56 |
| 11 | 0.68 | 0.67 | 0.72 | 0.67 | 0.68 | 0.68 | 0.68 |
| 12 | 0.71 | 0.77 | 0.76 | 0.72 | 0.72 | 0.70 | 0.74 |
| 13 | 0.89 | 0.85 | 0.77 | 0.91 | 0.90 | 0.89 | 0.90 |
| 14 | 0.65 | 0.70 | 0.74 | 0.66 | 0.65 | 0.65 | 0.65 |
| 15 | 0.72 | 0.71 | 0.80 | 0.72 | 0.72 | 0.74 | 0.73 |
| 16 | 0.69 | 0.67 | 0.81 | 0.68 | 0.68 | 0.69 | 0.68 |
| 17 | 0.76 | 0.76 | 0.77 | 0.76 | 0.76 | 0.77 | 0.76 |
| 18 | 0.74 | 0.74 | 0.82 | 0.75 | 0.74 | 0.74 | 0.74 |
| 19 | 0.69 | 0.70 | 0.76 | 0.70 | 0.69 | 0.70 | 0.69 |
| 20 | 0.69 | 0.68 | 0.73 | 0.69 | 0.69 | 0.69 | 0.68 |
| Mean | 0.71 | 0.72 | 0.74 | 0.72 | 0.72 | 0.71 | 0.72 |
| AR | 3.45 | 3.40 | 5.30 | 4.20 | 3.70 | 3.95 | 4.00 |

Table 19: Results (using the datasets in Table 11) for base technique *Tree* using a selection of the methods seen in section 3 as described in appendix G. Methods may perform equally since the Friedman statistic (10.65) is lower than the Critical Value (14.16), so the null hypothesis cannot be rejected (significance level: 0.05). Critical difference for the Nemenyi post-hoc test: 0.6922.

.

loss:

$$r^*(x, \ell_{A,\alpha}, \hat{f}) = \arg\min_t \left\{ \alpha \hat{\mu}(x) + t\hat{F}(t|x) - \alpha t - \int_{-\infty}^{t} y\hat{f}(y|x)dy \right\}$$

In order to find the minimum, we calculate the first derivative and equal it to 0:

$$\hat{F}(t|x) + t\hat{f}(t|x) - \alpha - t\hat{f}(t|x) = 0$$
$$\hat{F}(t|x) = \alpha$$

Since the second derivate is positive this is a minimum. □ □

*Proof.* (for proposition 6) We follow the same initial steps as in the absolute case. We derive the expected loss (eq. 2) and decompose the expression for $\ell^S_\alpha(\hat{y},y)$ depending on whether $t<y$ or not.

$$
\begin{aligned}
\mathscr{L}(x,t,\hat{f},\ell^S_\alpha) &= \int_{-\infty}^{\infty} \ell^S_\alpha(t,y)\hat{f}(y|x)dy \\
&= \int_{-\infty}^{t} (1-\alpha)(t-y)^2 \hat{f}(y|x)dy + \int_{t}^{\infty} \alpha(y-t)^2 \hat{f}(y|x)dy \\
&= \int_{-\infty}^{t} (1-\alpha)t^2 \hat{f}(y|x)dy + \int_{-\infty}^{t} (1-\alpha)(-2ty+y^2)\hat{f}(y|x)dy \\
&\quad + \int_{t}^{\infty} \alpha t^2 \hat{f}(y|x)dy + \int_{t}^{\infty} \alpha(-2ty+y^2)\hat{f}(y|x)dy \\
&= (1-\alpha)t^2 \hat{F}(t|x) + \alpha t^2 (1-\hat{F}(t|x)) \\
&\quad + \int_{-\infty}^{t} (1-\alpha)(-2ty+y^2)\hat{f}(y|x)dy + \int_{t}^{\infty} \alpha(-2ty+y^2)\hat{f}(y|x)dy \\
&= (1-2\alpha)t^2 \hat{F}(t|x) + \alpha t^2 + \int_{-\infty}^{t}(1-2\alpha)(-2ty+y^2)\hat{f}(y|x)dy - 2\alpha t \hat{\mu}(x) + \alpha \hat{\mu}_2(x) \\
&= (1-2\alpha)\left[t^2 \hat{F}(t|x) - 2t\int_{-\infty}^{t} y\hat{f}(y|x)dy + \int_{-\infty}^{t} y^2 \hat{f}(y|x)dy\right] + \alpha \left[t^2 - 2t\hat{\mu}(x) + \hat{\mu}_2(x)\right]
\end{aligned}
$$

where $\hat{\mu}_2(x)$ is the second raw moment of $\hat{f}(y|x)$. □ □

*Proof.* (for proposition 7) From proposition 6, we have:

$$
\begin{aligned}
r^*(x, \ell_{S,\alpha}, \hat{f}) &= \arg\min_{t} \mathscr{L}(x,t,\hat{f},\ell^S_\alpha) \\
&= \arg\min_{t} \left\{ (1-2\alpha)\left[t^2 \hat{F}(t|x) - 2t\int_{-\infty}^{t} y\hat{f}(y|x)dy + \int_{-\infty}^{t} y^2 \hat{f}(y|x)dy\right] + \alpha \left[t^2 - 2t\hat{\mu}(x) + \hat{\mu}_2(x)\right] \right\}
\end{aligned}
$$

Again, in order to find the minimum, we calculate the first derivative and equal it to 0:

$$
(1-2\alpha)\left[t^2 \hat{f}(t|x) + 2t\hat{F}(t|x) - 2\int_{-\infty}^{t} y\hat{f}(y|x)dy - 2t\cdot t\hat{f}(t|x) + t^2\hat{f}(t|x)\right] + 2\alpha t - 2\alpha\hat{\mu}(x) + 0 = 0
$$

$$
(1-2\alpha)\left[2t\hat{F}(t|x) - 2\int_{-\infty}^{t} y\hat{f}(y|x)dy\right] + 2\alpha t - 2\alpha\hat{\mu}(x) = 0
$$

The second derivative is:

$$
(1-2\alpha)\left[2\hat{F}(y|x) + 2t\hat{f}(y|x) - 2t\hat{f}(y|x)\right] + 2\alpha = (1-2\alpha)2\hat{F}(y|x) + 2\alpha
$$

which is always positive since both $\hat{F}(y|x)$ and $\alpha$ are between 0 and 1. Consequently, we have a minimum. □ □

*Proof.* (for proposition 8) Assuming $\hat{f}(t|x)$ is a normal distribution, we can standardise $\hat{f}(t|x)$ as $\phi(t')$ with $t' = \frac{t'-\hat{\mu}(x)}{\hat{\sigma}(x)}$ Then, proposition 7 reduces to:

$$
(1-2\alpha)\left[2t'\Phi(t') - 2\int_{-\infty}^{t'} y\phi(y)dy\right] + 2\alpha t' - 0 = 0
$$

The partial (from $-\infty$ to $t$) first moment of the standard normal distribution is just $-\phi(t)$. This can also be seen as a truncated standard normal distribution whose expected value is: $\mathbb{E}(u|u \leq t) = -\frac{\phi(t)}{\Phi(t)}$. Since the truncated standard normal distribution is normalised by $\Phi(t)$ we get $-\phi(t)$. This can also be obtained by just solving the integral. From here,

$$(1-2\alpha)\left[2t'\Phi(t')+2\phi(t')\right]+2\alpha t' = 0$$
$$t'\Phi(t')+\phi(t')+t'\frac{\alpha}{1-2\alpha} = 0$$

where $t$ is obtained using $t = \hat{\sigma}(x)t' + \hat{\mu}(x)$. □ □

*Proof.* (for proposition 9) We start from the expression of the expected loss (proposition 4):

$$\mathscr{L}(x,t,\hat{f},\ell_\alpha^A) = -\int_{-\infty}^t y\hat{f}(y|x)dy + \alpha\hat{\mu}(x) + t\hat{F}(t|x) - \alpha t$$

In order to reduce $\int_{-\infty}^t y\hat{f}(y|x)dy$, we see that it is a partial moment of the normal distribution. This is equal to an unnormalised version of the expected value of a truncated distribution, which is $\mathbb{E}(X|X \leq T) = \mu - \sigma\frac{\phi(\tau)}{\Phi(\tau)}$ with $\tau = \frac{T-\mu}{\sigma}$. Consequently, this term reduces to $\Phi(t')\hat{\mu}(x) - \hat{\sigma}(x)\phi(t')$ with $t' = \frac{t-\hat{\mu}(x)}{\hat{\sigma}(x)}$ ($t = \hat{\sigma}(x)t' + \hat{\mu}(x)$).
So, we have:

$$\begin{aligned}\mathscr{L}(x,t,\hat{f},\ell_\alpha^A) &= \alpha\hat{\mu}(x) + t\Phi(t') - \alpha t - \Phi(t')\hat{\mu}(x) + \hat{\sigma}(x)\phi(t')\\
&= (t-\hat{\mu}(x))\Phi(t') + \hat{\sigma}(x)\phi(t') - \alpha(t-\hat{\mu}(x))\\
&= \left[\frac{t-\hat{\mu}(x)}{\hat{\sigma}(x)}\Phi(t') + \frac{\hat{\sigma}(x)}{\hat{\sigma}(x)}\phi(t') - \alpha\frac{t-\hat{\mu}(x)}{\hat{\sigma}(x)}\right]\hat{\sigma}(x)\\
&= [t'\Phi(t') + \phi(t') - \alpha t']\hat{\sigma}(x)\end{aligned}$$

□ □

*Proof.* (for proposition 10) We start from the expression of the expected loss (proposition 6):

$$\mathscr{L}(x,t,\hat{f},\ell_\alpha^S) = (1-2\alpha)\left[t^2\hat{F}(t|x) - 2t\int_{-\infty}^t y\hat{f}(y|x)dy - \int_{-\infty}^t y^2\hat{f}(y|x)dy\right]$$
$$+\alpha[t^2 - 2t\hat{\mu}(x) + \hat{\mu}_2(x)] \tag{17}$$

We reduce $\int_{-\infty}^t y\hat{f}(y|x)dy$ as we did in the proof of proposition 9, as a partial moment of the normal distribution, to $\Phi(t')\hat{\mu}(x) - \hat{\sigma}(x)\phi(t')$ with $t' = \frac{t-\hat{\mu}(x)}{\hat{\sigma}(x)}$ ($t = \hat{\sigma}(x)t' + \hat{\mu}(x)$).

We reduce $\int_{-\infty}^t y^2\hat{f}(y|x)dy$ as a partial second order moment of the normal distribution (or a full second order moment of the truncated normal distribution), to $\Phi(t')(\hat{\mu}(x)^2 - 2\hat{\mu}(x)\hat{\sigma}(x)\frac{\phi(t')}{\Phi(t')} + \hat{\sigma}(x)^2(1 - t'\frac{\phi(t')}{\Phi(t')}))$. The last term $\hat{\mu}_2(x)$ is the second order moment of the normal distribution, which is just $\hat{\mu}(x)^2 + \hat{\sigma}(x)^2$. Plugging all this into (17), and using the short notation $\mu$ for $\hat{\mu}(x)$ and $\sigma$ for $\hat{\sigma}(x)$ we have:

$$
\begin{aligned}
\mathscr{L}(x,t,\hat{f},\ell_\alpha^S) &= (1-2\alpha)\left[t^2\Phi(t') - 2t(\Phi(t')\mu - \sigma\phi(t')) - \Phi(t')(\mu^2 - 2\mu\sigma\frac{\phi(t')}{\Phi(t')} + \sigma^2(1 - t'\frac{\phi(t')}{\Phi(t')}))\right] \\
&\quad + \alpha[t^2 - 2t\mu + (\mu^2 + \sigma^2)] \\
&= (1-2\alpha)\Phi(t')\left[t^2 + -2t(\mu - \sigma\frac{\phi(t')}{\Phi(t')}) - (\mu^2 - 2\mu\sigma\frac{\phi(t')}{\Phi(t')} + \sigma^2(1 - t'\frac{\phi(t')}{\Phi(t')}))\right] \\
&\quad + \alpha((t-\mu)^2 + \sigma^2) \\
&= (1-2\alpha)\Phi(t')\left[t^2 + -2t(\mu - \sigma q(t')) - (\mu^2 - 2\mu\sigma q(t') + \sigma^2(1 - t' q(t')))\right] \\
&\quad + \alpha((t-\mu)^2 + \sigma^2) \\
&= \Phi(t')(1-2\alpha)\left[t^2 - 2t(\mu - \sigma q(t')) - (\mu^2 - 2\mu\sigma q(t') + \sigma^2(1 - t'q(t')))\right] + \alpha((t-\mu)^2 + \sigma^2) \\
&= \Phi(t')(1-2\alpha)\left[(t'\sigma + \mu)^2 - 2(t'\sigma + \mu)(\mu - \sigma q(t')) - (\mu^2 - 2\mu\sigma q(t') + \sigma^2(1 - t'q(t')))\right] \\
&\quad + \alpha((t'\sigma)^2 + \sigma^2) \\
&= \Phi(t')(1-2\alpha)\left[(t'\sigma)^2 + \mu^2 + 2t'\sigma^2 q(t') - 2\mu^2 + 2\mu\sigma q(t') - \mu^2 + 2\mu\sigma q(t') - \sigma^2 + \sigma^2 t' q(t')\right] \\
&\quad + \alpha\sigma^2(t' + 1) \\
&= \Phi(t')(1-2\alpha)\left[(t'\sigma)^2 + 3t'\sigma^2 q(t') - 2\mu^2 + 4\mu\sigma q(t') - \sigma^2\right] + \alpha\sigma^2(t' + 1)
\end{aligned}
$$

with $q(t') = \frac{\phi(t')}{\Phi(t')}$. $\square$ $\square$